\pdfoutput=1

\documentclass[]{openmoss}
\usepackage{helvet}

\usepackage[toc,page,header]{appendix}
\usepackage{titletoc}
\usepackage{minitoc}

\usepackage{algorithm}
\usepackage{algpseudocode}

\usepackage{wasysym}
\usepackage{fontawesome}

\usepackage{enumitem}

\usepackage{listings}

\usepackage{pgfplots}
\pgfplotsset{compat=1.18}
\usetikzlibrary{arrows.meta, positioning, calc, fit, backgrounds, shapes.geometric, decorations.pathmorphing}

\usepackage{arydshln}
\usepackage{adjustbox}
\usepackage{colortbl}
\usepackage{booktabs}
\usepackage{makecell}
\usepackage{multirow}
\usepackage{multicol}

\usepackage{xspace}
\usepackage{nicefrac}
\usepackage{float}
\usepackage{comment}
\usepackage{wrapfig}
\usepackage{pifont}
\usepackage{dsfont}

\usepackage{amsmath, amssymb}
\usepackage{amsthm}

\usepackage[most]{tcolorbox}
\tcbuselibrary{breakable}
\usepackage{tabularx}
\usepackage{xcolor}
\definecolor{case_red}{RGB}{190,0,0}
\definecolor{case_blue}{RGB}{0,102,204}
\definecolor{case_green}{RGB}{0,153,0}

\title{Mechanics of Long-Context Hybrid Models\\ Part 1.1: From Hybrid Attention to Hybrid Position}

\author{
    Xiaoran Liu\textsuperscript{1,2,3},\ 
    Ziwei He\textsuperscript{2,3,$\dagger$},\ 
    Xipeng Qiu\textsuperscript{1,2,3,$\dagger$}
}
\affiliation[1]{\mbox{Fudan University}}
\affiliation[2]{\mbox{Shanghai Innovation Institute}}
\affiliation[3]{\mbox{OpenMOSS Team}}

\abstract{
The architectural design of Large Language Models (LLMs) is shifting from traditional full-attention-only models to hybrid models, which combine different attention modules to improve long-context efficiency and performance in length extrapolation and context extension. To explain why hybrid models work and how to design them better, we propose \textbf{Mechanics of Long-Context Hybrid Models}. As Part 1.1 of this series, we begin with hybrids of full attention and either sliding-window attention (SWA) or gated variants of linear attention (LA), represented by GLA and GDN. We first observe a \textbf{Seesaw Effect in Context Extension}: LA hybrids benefit more from long-context continual pretraining, whereas SWA hybrids perform better under length extrapolation. We attribute this behavior to differences in the positional inductive biases induced by these attention mechanisms. We find that SWA hybrids suffer from a \textbf{Short-Context Learning Trap}, Short-Window Weariness, and Long-Window Laziness, and require extended windows to enhance performance in continual long-context pretraining. For LA hybrids, we summarize the \textbf{Matthew Effect of Hybrid Position Extrapolation} and propose \textbf{Sliding-Window Linear Attention}, achieving 16$\times$ training-free length extrapolation while maintaining 100\% accuracy on NIAH-SK1 in 64k context length.
}

\contact{
    \email{xrliu24@m.fudan.edu.cn},\ \ 
    \email{ziwei.he@sii.edu.cn},\ \ 
    \email{xpqiu@fudan.edu.cn}
}
\coderepository{\url{https://github.com/OpenMOSS/Hybrid-Mechanics}}

\begin{document}
\maketitle

\begin{tcolorbox}[colback=blue!5!white,colframe=blue!75!black,title=\textbf{Takeaway List}]
\begin{itemize}[leftmargin=*]
    \itemsep0.3em
    \item[\textbf{T1}] \textbf{From Hybrid Attention to Hybrid Position}. Hybrid models improve long-context performance and length extrapolation through hybrid position, combining NoPE with other position biases. 
    \item[\textbf{T2}] \textbf{Seesaw Effect in Context Extension of SWA Hybrid}. The advantage of SWA-NoPE hybrids over LA-NoPE hybrids in long-context performance is reversed after long-context pretraining, especially for layer-wise hybrids, due to their \textbf{Short-Context Learning Trap in Context Extension}.
    \item[\textbf{T3}] \textbf{No-Free-Lunch Effect in Length Extrapolation of LA Hybrid}. LA-NoPE hybrids are weaker than SWA-NoPE hybrids in direct extrapolation, though stronger within the training context.
    \item[\textbf{T4}] \textbf{Tidal Effect in Collaboration of Hybrid Position}. Effective collaboration of hybrid position relies on a minority of high-hit-rate NoPE attention with coarse aggregation and a majority of low-entropy position-biased attention (i.e., RoPE attention and gated linear attention) for noise reduction.
    \item[\textbf{T5}] \textbf{Short-Window Weariness and Long-Window Laziness of SWA Hybrid}. In length extrapolation, SWA hybrids with short window size perform better, but in context extension, SWA hybrids with sliding window extended to a larger size overcome the short-context trap better.
    \item[\textbf{T6}] \textbf{Matthew Effect of Hybrid Position Extrapolation}. Applying a sliding window in position-biased attention and enhanced global aggregation in NoPE attention leads to strong length extrapolation.
\end{itemize}
\end{tcolorbox}

\section{Introduction}

The architectural design of mainstream open-source Large Language Models (LLMs)~\citep{brown2020language,sun2024moss,cai2024internlm2,agiteam2025nexn1,dubey2024llama,guo2025deepseekr1,yang2025qwen3} is shifting from traditional full-softmax-attention-only models~\citep{vaswani2017attention,ainslie2023gqa,liu2024deepseek,liu2025deepseek} to hybrid models~\citep{team2024jamba,agarwal2025gpt,team2026gemma,li2025minimax,team2026kimi,qwen35blog,mimo2026v25pro,zeng2026glm,xu2026deepseek}, which combine different attention modules across layers or heads and introduce different position embeddings to improve long-context efficiency and performance~\citep{wang2025systematic,team2025kimi,merrill2026olmo}. To this end, we propose \textbf{Mechanics of Long-Context Hybrid Models}. Through systematic training, evaluation, mechanism analysis, and closed-loop validation, we analyze how different attention modules inside a hybrid model collaborate and how these collaborations affect length extrapolation and long-context performance, aiming to answer "\textbf{why hybrid models are effective}" and "\textbf{how to design them more effectively}" for long-context LLMs.
\begin{itemize}
    \item \textbf{Why Hybrid?} Hybrid models have become an important paradigm in LLM design~\citep{agarwal2025gpt,team2026kimi,qwen35blog}. Compared with traditional RoPE-based full-attention-only models~\citep {su2024roformer,touvron2023llama}, hybrid models not only offer higher computational and memory efficiency~\citep{liu2024deepseek,liu2025deepseek,xu2026deepseek}, but also provide better performance in retrieval, state tracking, language modeling, and length extrapolation~\citep{puvvada2025swan,hu2026hardware,hu2026every,team2025kimi,merrill2026olmo,yang2026rope}. 
    \item \textbf{Why Long-Context?} LLMs need to support longer context~\citep{liu2025thus,liu2025comprehensive,zeng2024scaling,li2025minimax,xu2026deepseek}, and long context is one of the most important application scenarios for current hybrid models~\citep{puvvada2025swan,team2025kimi,merrill2026olmo}. At the same time, million-token-level contexts in the agentic era impose strong constraints on model architectures~\citep{li2025minimax,team2026kimi,xu2026deepseek}. Relying solely on full attention is prohibitively expensive, while any single efficient attention is unreliable~\citep{yang2024gated,yang2025gated,team2025kimi}. This makes hybrid models worth studying in long-context LLMs~\citep{wang2025systematic,merrill2026olmo,qiao2026rethinking,tan2026hydrahead}.
    \item \textbf{Why Mechanics?} There has already been substantial work discussing hybrid models from the perspectives of downstream task evaluation, scaling curves, or system efficiency~\citep{puvvada2025swan,wang2025systematic,merrill2026olmo,qiao2026rethinking,tan2026hydrahead}, but these results often remain at the level of comparing the phenomenon of "which model is better"~\citep{merrill2026olmo,qiao2026rethinking}. We instead focus on how different attention modules inside a hybrid model interact, collaborate, and affect length extrapolation~\citep{presstrain} during inference and context extension during training. This is why we use Mechanics to summarize this series of works.
\end{itemize}

On account of this long journey involving comparisons across multiple hybrid modules, hybrid paradigms, and training scenarios, we proceed step by step. This paper is Part 1.1 of the whole series.
\begin{itemize}
    \item \textbf{In Part 1}, we first discuss the hybridization of memory-constant sliding-window attention (SWA) or linear attention (LA) with full attention, which is also one of the most widely used classes of hybrid models at present~\citep{agarwal2025gpt,team2026gemma,mimo2026v25pro,merrill2026olmo,qwen35blog,team2026kimi}; in subsequent work, we will continue to explore hybrid models such as sparse-attention hybrids~\citep{liu2025deepseek,lai2026minimax,gao2026hysparse,hu2026every,hu2026hierarchical} and compressed-attention hybrids~\citep{xu2026deepseek,chu2026kwai}. 
    \item \textbf{In Part 1.1}, we start from the perspective of hybrid position, focusing on the length extrapolation (i.e. training-free generalization beyond the pretraining context length) and context extension (i.e. long-context continual pretraining that expands context length) performance of RoPE-based full-attention models, RoPE-NoPE hybrid-position full-attention models, SWA hybrid models, and LA hybrid models represented by GLA/GDN~\citep{yang2024gated,yang2025gated,qwen35blog}. In this paper, we focus on the long-context language modeling~\citep{rae2019compressive,fang2024wrong}, retrieval~\citep{niah,hsieh2024ruler}, and state-tracking abilities~\citep{kuratov2024babilong} in pretraining; in subsequent work, we will continue to explore more advanced abilities in post-training and multi-modal training.
\end{itemize}

Regarding the results obtained in Part 1.1, we give a complete Takeaway List on the first page of the paper. Specifically, the contributions of this paper can be summarized as follows:
\begin{itemize}
    \item We compare RoPE-based full attention, RoPE-NoPE hybrid, SWA hybrid, and GLA/GDN hybrid, both layer-wise and head-wise, across model scales of 376M, 776M, and 1B through long-context retrieval, state-tracking, and language-modeling performance under length extrapolation in short-context pretraining and context extension after long-context continual pretraining.
    \item \textbf{On hybrid position}, we find that hybrid models improve long-context performance and length extrapolation by combining NoPE with other positional biases, i.e., RoPE attention and gated linear attention (\textbf{Takeaway 1}). They occupy distinct functional regions in the entropy-hit-rate scatter diagram, where the boundaries shift with the hybrid ratio (\textbf{Takeaway 4}, \textbf{Tidal Effect}).
    \item \textbf{On SWA hybrids}, we demonstrate that although SWA hybrids exhibit clear advantages in length extrapolation, they suffer from the short-context learning trap in context extension (\textbf{Takeaway 2, Seesaw Effect}). To address this, we improve SWA hybrids by enlarging the window size and applying the LongCE loss, and further summarize the underlying short-window weariness (\textbf{Takeaway 5}).
    \item \textbf{On LA hybrids}, we find that LA hybrids perform better in long-context pretraining, but are weaker in length extrapolation (\textbf{Takeaway 3}, \textbf{No-Free-Lunch Effect}). Through our extended attention-entropy analysis, we observe that LA and RoPE attention exhibit similar patterns in the entropy-hit-rate diagram, and both can be viewed as position-biased attention. Based on this, we propose \textbf{Sliding-Window Linear Attention}, achieving 16$\times$ length extrapolation with 100\% accuracy on NIAH-SK1, and summarize the underlying \textbf{Matthew Effect} of Hybrid Position Extrapolation (\textbf{Takeaway 6}).
\end{itemize}

\section{Related Work}

\subsection{The Rise of Hybrid Models}

LLM architecture design has shifted from traditional full-attention-only models to hybrid models~\citep{team2024jamba,agarwal2025gpt,team2026gemma,li2025minimax,team2026kimi,qwen35blog,mimo2026v25pro,zeng2026glm,xu2026deepseek}, combining attention modules with different forms, complexities, and position embeddings to mix information along the sequence dimension~\citep{wang2025systematic,team2025kimi}. For example, Qwen-3.5~\citep{qwen35blog,qwen3.8flashnext}, Kimi-K3~\citep{team2025kimi,team2026kimi}, and OLMo-Hybrid~\citep{merrill2026olmo} mix full softmax attention (e.g., GQA~\citep{ainslie2023gqa} and MLA~\citep{liu2024deepseek}) or sparse attention (e.g., DSA~\citep{liu2025deepseek}) with linear attention (e.g., GDN~\citep{yang2025gated} and KDA~\citep{team2025kimi}). Gemma4~\citep{team2026gemma}, Inkling~\citep{inkling}, and Mimo-v2~\citep{mimo2026v25pro} mix full softmax attention with sliding-window softmax attention. GLM-5.3~\citep{zeng2026glm,bai2026indexcache} and MiniMax-M3~\citep{minimax-m3,lai2026minimax} adopt sparse-attention hybrids~\citep{liu2025deepseek,gao2026hysparse,hu2026hierarchical}, while DeepSeek-v4~\citep{xu2026deepseek} uses compressed-attention hybrids~\citep{chu2026kwai} with different compression intensity. Hybrid attention often comes with hybrid positional biases and has position-embedding designs that differ from those in traditional models. For example, Jamba~\citep{team2024jamba}, Kimi-K3~\citep{team2026kimi}, SWAN-GPT~\citep{puvvada2025swan}, and RNoPE~\citep{yang2026rope} also show that full-attention layers in hybrid models do not necessarily require explicit position embeddings like RoPE~\citep{su2024roformer}.

These hybrid models can be more computationally and memory efficient than traditional models, while achieving downstream performance close to or even better than traditional full-attention models. As a result, hybrid models have received increasing attention~\citep{wang2025systematic,team2026kimi,merrill2026olmo,qiao2026rethinking}. Among them, hybrids with fixed memory complexity, especially linear attention (LA) and sliding-window attention (SWA) hybrids, after the early exploration~\citep{de2024griffin,botev2024recurrentgemma,ren2025samba}, have since been adopted by LLMs including Jamba~\citep{team2024jamba}, GPT-OSS~\citep{agarwal2025gpt}, MiniMax-01~\citep{li2025minimax}, Qwen~\citep{qwen35blog,qwen3.8flashnext}, Kimi~\citep{team2025kimi,team2026kimi}, Nemotron-H~\citep{blakeman2025nemotron}, and MiniCPM-SALA~\citep{team2026minicpm}. They have also inspired fine-tuning attempts~\citep{lan2025liger,chen2026hybrid,lan2026morphing,jolicoeur2026sliding}, and refined efficiency optimizations~\citep{sun2024you,hu2026hardware}. Therefore, we start from this class of hybrids to study why hybrid models are effective and how to design them more effectively.

\subsection{Long-Context Performance of Hybrid Models}

Beyond efficiency, previous work also suggests that hybrid models can bring performance gains within the training context length~\citep{team2026kimi,merrill2026olmo,yang2026rope}. Early works~\citep{fu2022hungry,cooper2026expressivity} show that hybrid models can combine the advantages of traditional softmax attention in retrieval with the advantages of linear attention in state tracing. Kimi-K3~\citep{team2026kimi} and OLMo Hybrid~\citep{merrill2026olmo} further verify at larger scales that LA hybrids can improve model expressivity and pretraining scaling efficiency. In addition to layer-wise hybrids, Hymba~\citep{dong2024hymba} and HydraHead~\citep{tan2026hydrahead} also verify the effectiveness of head-wise hybrids.

Beyond empirical validation, some works have begun to analyze SWA and LA hybrids theoretically. For example, \citet{wang2025systematic} suggests that a ratio around 3:1 suits hybrid models, while \citet{wu2023rethinking} points out that retrieval is mainly handled by full attention, whereas SWA or LA affects the training trajectory. Although these works explain the benefits of hybrid models, there is still a lack of mechanistic analysis of how attention modules cooperate based on their different positional inductive biases in length extrapolation and context extension. Based on this, can we combine the strengths and weaknesses of different attention modules in their positional-bias designs and derive new hybrid strategies that advance length extrapolation and context extension? These are the questions we study in this paper.

\subsection{Extrapolation based on Hybrid Models}

Besides improving long-context performance within the training length, hybrid models have also been shown to improve length extrapolation, or length generalization~\citep{presstrain}. For example, iRoPE~\citep{meta2025llama}, RNoPE~\citep{yang2026rope}, and SWAN-GPT~\citep{puvvada2025swan} use RoPE only in some layers and restrict the attention window of these layers during extrapolation, while retaining access to full context only in NoPE layers, achieving strong extrapolation performance. This hybrid position design has also proven to improve retrieval performance within the training length~\citep{yang2026rope}. Similarly, in LA hybrids, Jamba~\citep{team2024jamba}, Kimi-K3~\citep{team2026kimi}, and HypeNet~\citep{chen2026hybrid} find that RoPE can also be removed from full-attention layers, which also helps direct length generalization~\citep{presstrain}.

Although LA hybrids do not use explicit position embeddings, this does not mean there is no positional bias. Linear attention also contains various position designs, some of which have already been used to improve traditional softmax attention and build stronger position embeddings or attentions~\citep{linearattn}, such as FoX~\citep{lin2025forgetting}, DeltaFormer~\citep{zhong2025understanding}, and PaTH~\citep{yang2025path}. Accordingly, we can extrapolate LA hybrids in the same way as RoPE-NoPE hybrids by limiting position-biased attention to local windows while strengthening global perception in NoPE attention, which is also one of the contributions in this paper.

\section{Observation}\label{sec-obs}

\subsection{Setup}

We conduct 4k short-context pretraining with 50B tokens and 32k long-context continual pretraining with 5B tokens at 376M, 776M, and 1B model sizes, and we also verify key conclusions by pretraining a 3B model with the configuration detailed in Table~\ref{tab-config} in Appendix~\ref{app-more}. By default, we use length extrapolation performance to refer to long-context performance after 4k short-context pretraining and context extension performance to refer to that after 32k long-context continual pretraining. We abbreviate layer-wise hybrid as LH and head-wise hybrid as HH, and use RoPE and NoPE to denote full attention with and without RoPE, respectively. For example, SWA-NoPE-LH denotes a model where SWA and NoPE full attention are mixed by layer with a ratio of 3:1~\citep{wang2025systematic,qwen35blog,puvvada2025swan} by default. We set the window size of SWA to 128, following GPT-OSS~\citep{agarwal2025gpt} and Mimo-v2.5~\citep{mimo2026v25pro}, and set the rotary base of SWA and RoPE full attention to 10000~\citep{su2024roformer}, while GLA and GDN follow their original papers~\citep{yang2024gated,yang2025gated,yang2024fla}. We only use MHA, since GDN does not support GQA.

We use the PG19 dataset~\citep{rae2019compressive} for perplexity (PPL) and LongPPL~\citep{fang2024wrong} evaluation, and use RULER~\citep{hsieh2024ruler} and BABILong~\citep{kuratov2024babilong} to evaluate retrieval and state-tracking abilities, respectively. In addition, we also report the average scores within 4k and beyond 4k as references for in-domain and out-of-domain performance. 

\subsection{Short-Context Training}

\begin{figure}[!thb]
  \noindent
  \centering
  \begin{subfigure}[b]{0.48\linewidth}
    \centering
    \includegraphics[width=\linewidth]{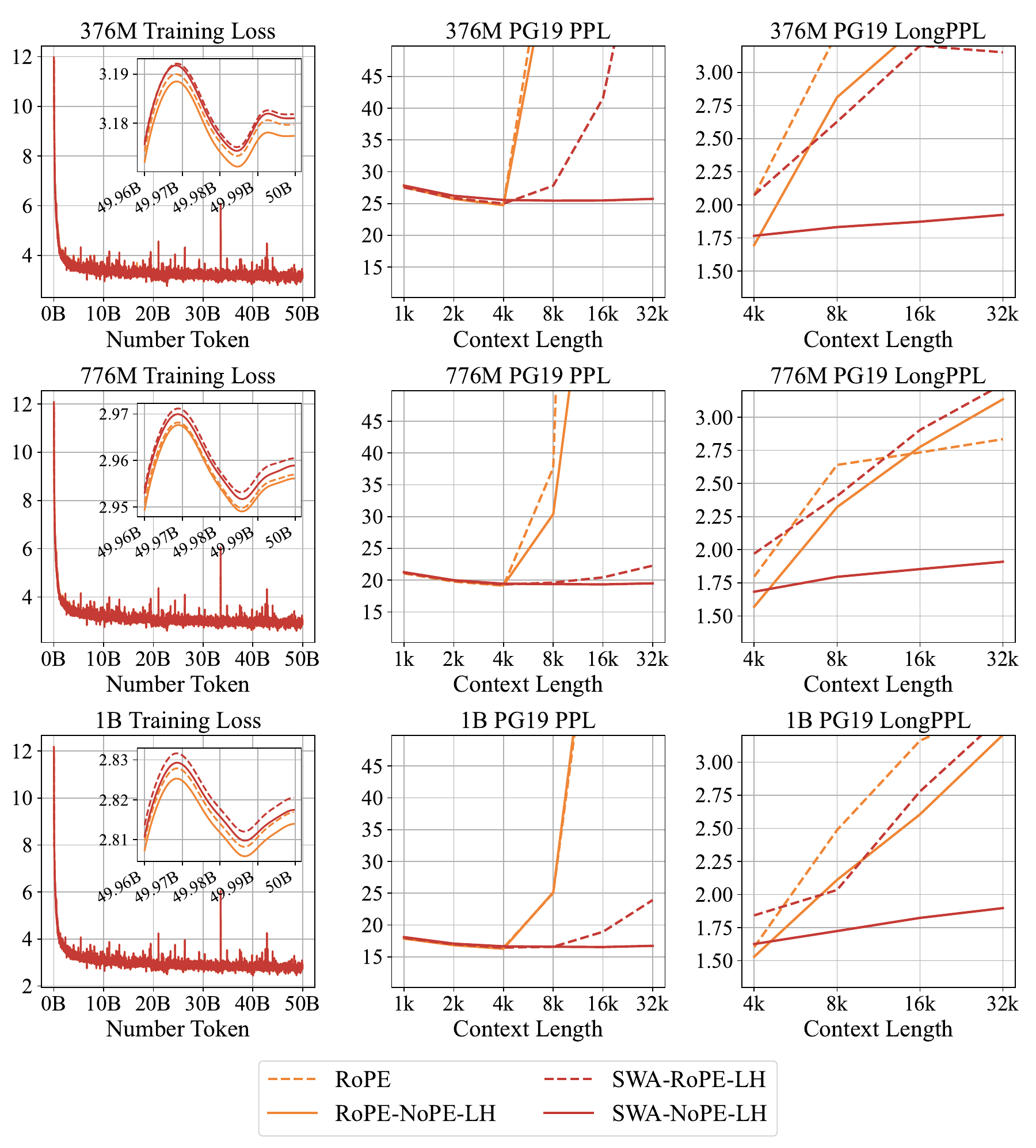}
    \caption{Results on Full Attention and SWA Hybrid.}
  \end{subfigure}
  \hfill
  \begin{subfigure}[b]{0.48\linewidth}
    \centering
    \includegraphics[width=\linewidth]{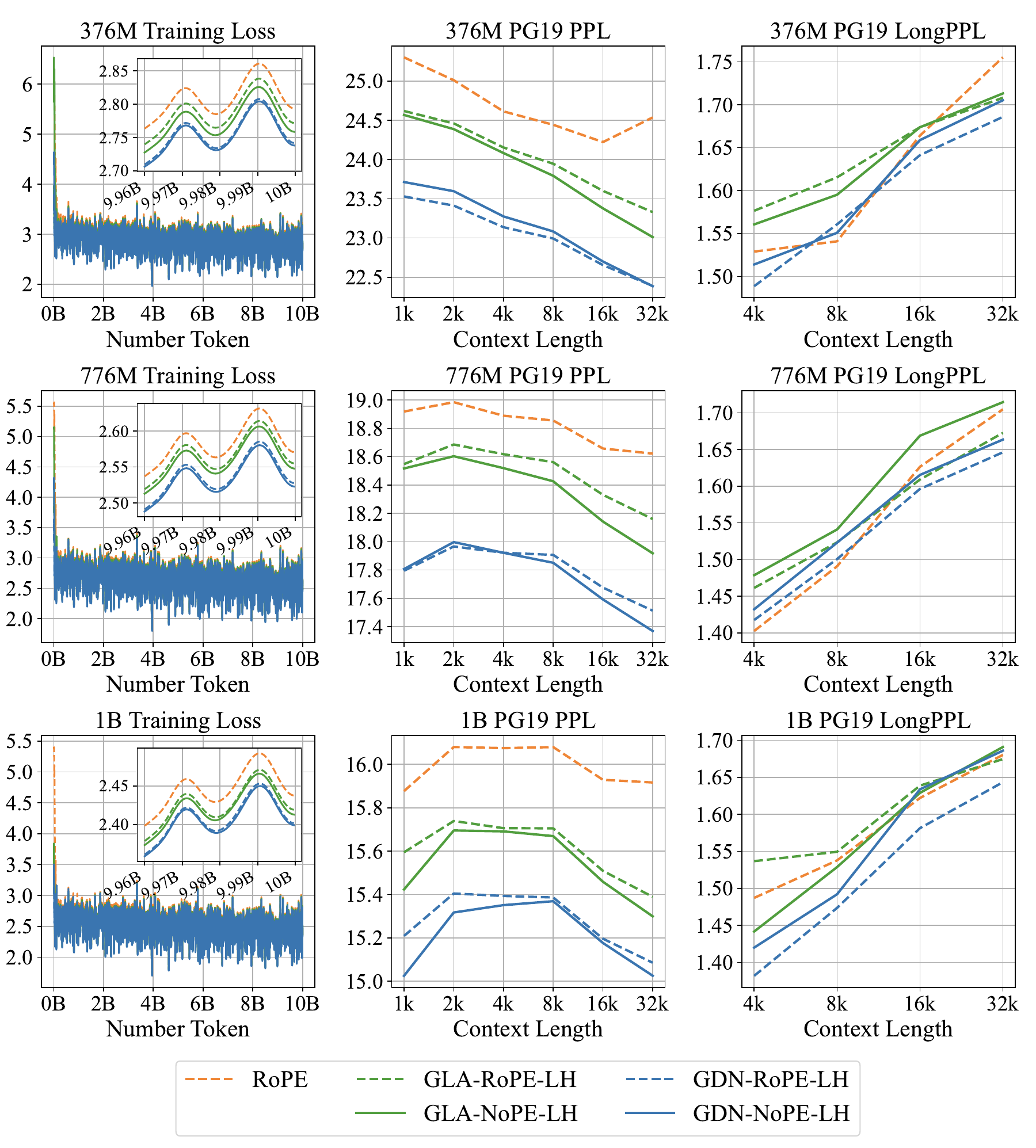}
    \caption{Results on Full Attention and GLA/GDN Hybrid.}
  \end{subfigure}
  \caption{Comparison of training loss, perplexity (PPL), and LongPPL among different layer-wise hybrid models after 4k context length pretraining. SWA/GLA/GDN-NoPE hybrids achieve stronger long-context language modeling.}
  \label{fig-main_ppl-lh}
\end{figure}

\begin{figure}[!thb]
  \noindent
  \centering
  \begin{subfigure}[b]{0.48\linewidth}
    \centering
    \includegraphics[width=\linewidth]{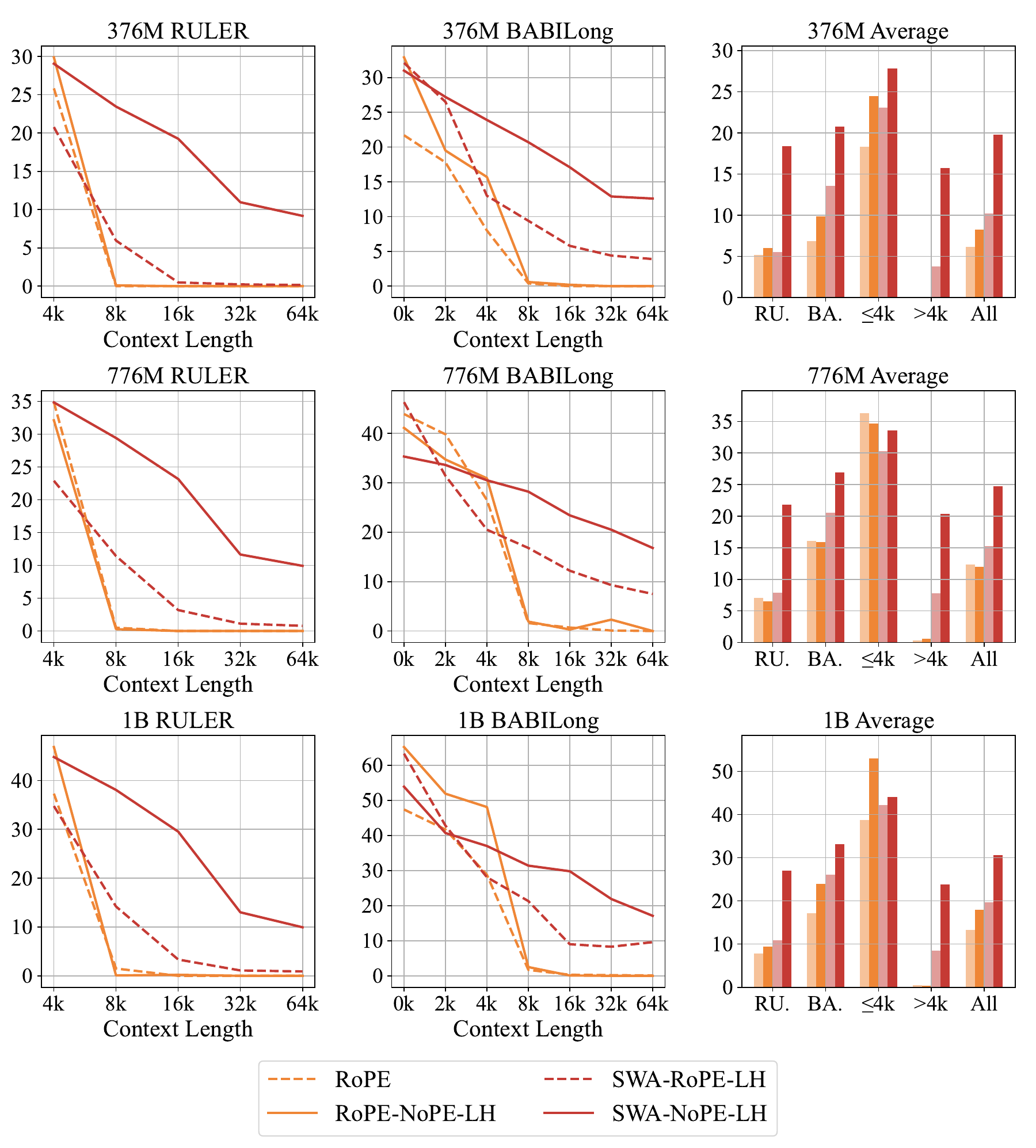}
    \caption{Results on Full Attention and SWA Hybrid.}
  \end{subfigure}
  \hfill
  \begin{subfigure}[b]{0.48\linewidth}
    \centering
    \includegraphics[width=\linewidth]{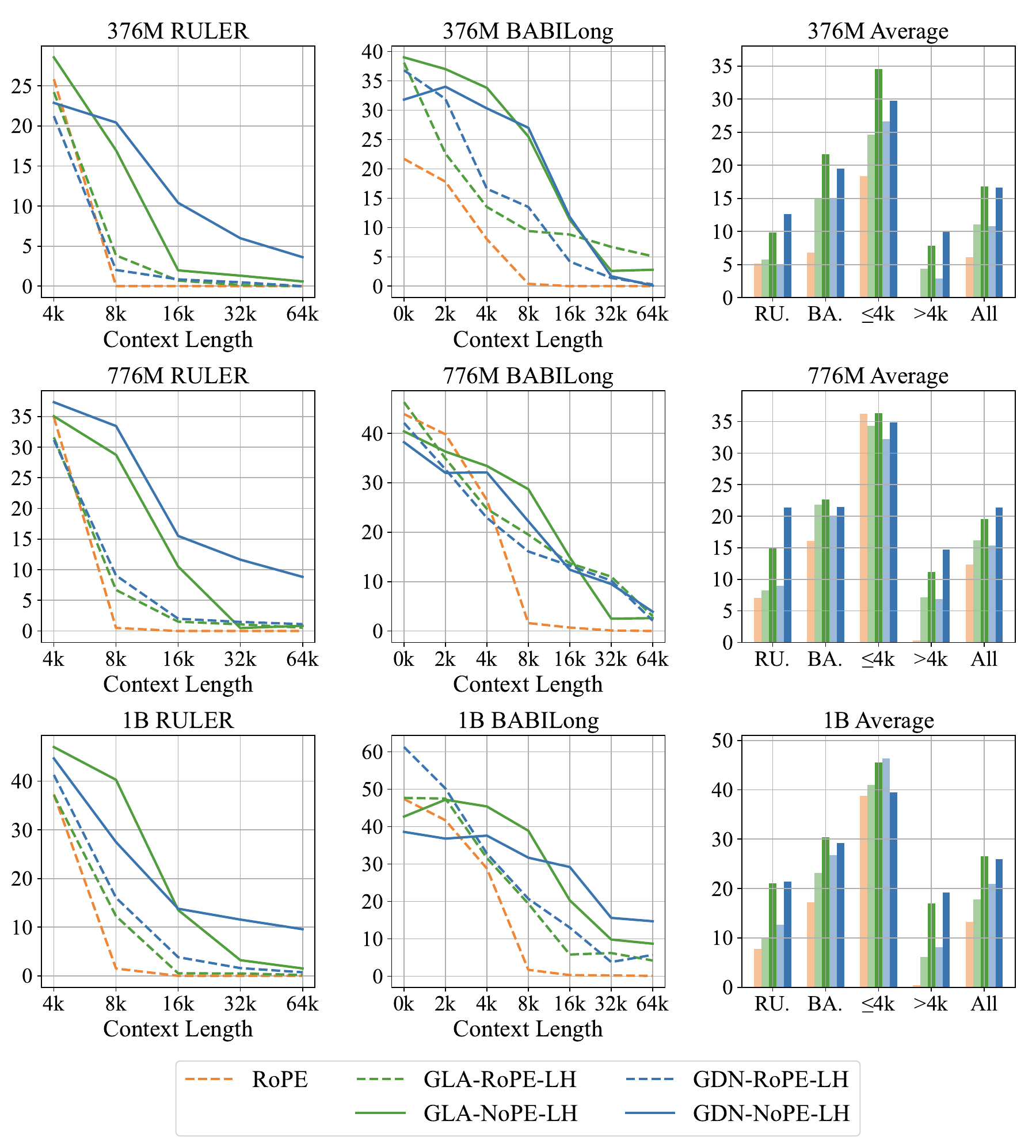}
    \caption{Results on Full Attention and GLA/GDN Hybrid.}
  \end{subfigure}
  \caption{Comparison of long-context performance under RULER and BABILong benchmarks among different layer-wise hybrid models after 4k context length pretraining. SWA/GLA/GDN-NoPE hybrids achieve both stronger long-context downstream performance within and beyond the training context length.}
  \label{fig-main_ruler-lh}
\end{figure}

\begin{figure}[!tb]
  \noindent
  \centering
  \begin{subfigure}[b]{0.48\linewidth}
    \centering
    \includegraphics[width=\linewidth]{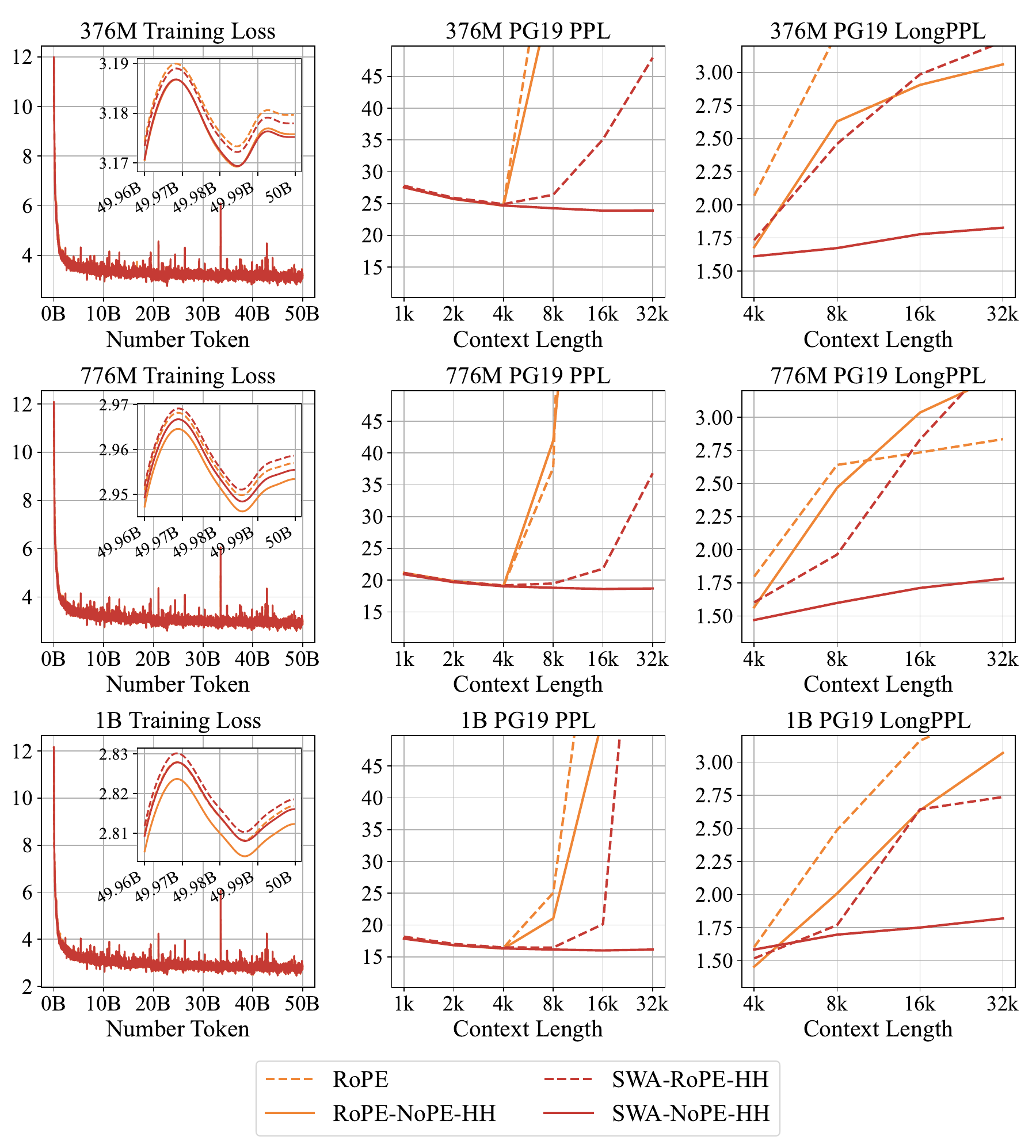}
    \caption{Results on Full Attention and SWA Hybrid.}
  \end{subfigure}
  \hfill
  \begin{subfigure}[b]{0.48\linewidth}
    \centering
    \includegraphics[width=\linewidth]{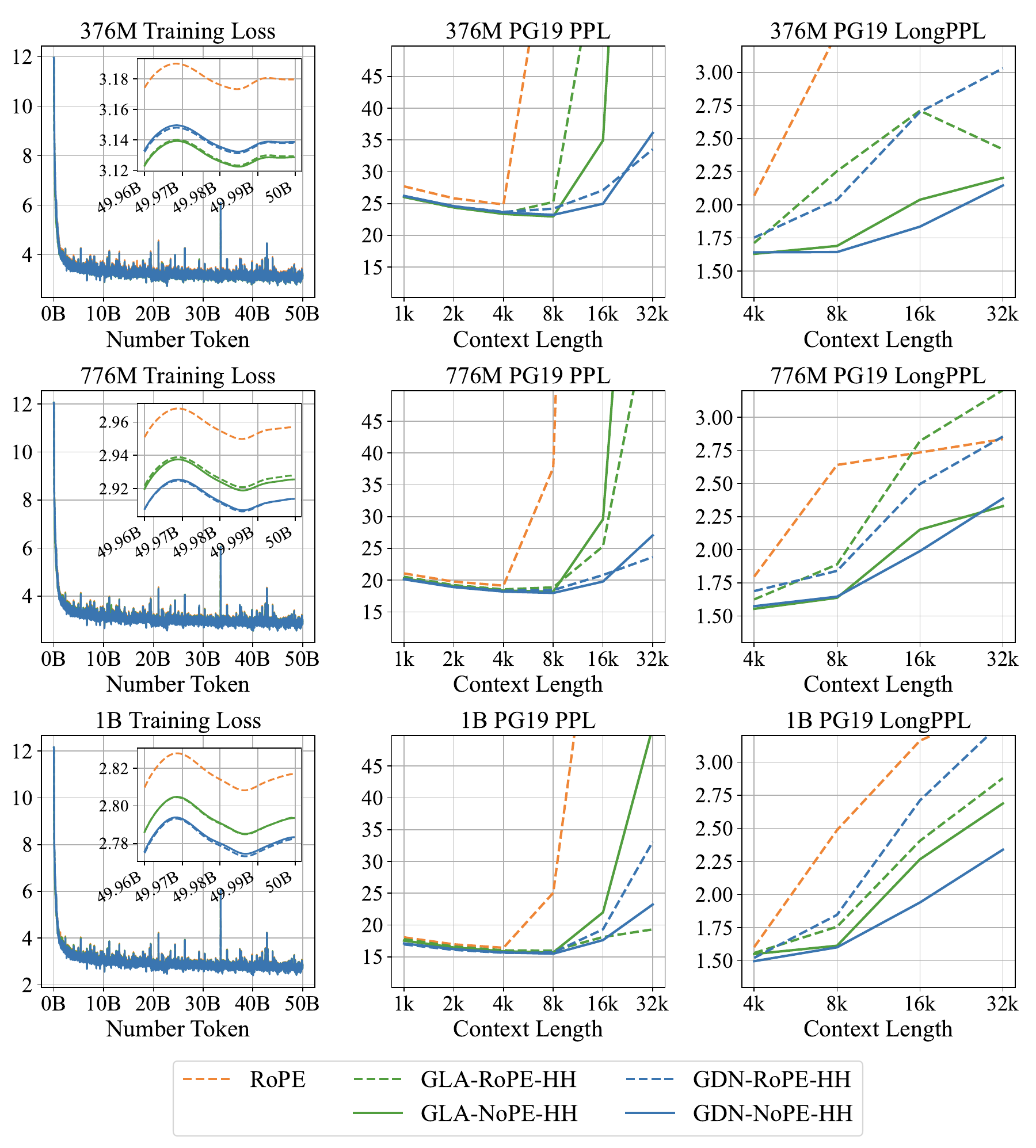}
    \caption{Results on Full Attention and GLA/GDN Hybrid.}
  \end{subfigure}
  \caption{Comparison of training loss, perplexity (PPL), and LongPPL among different head-wise hybrid models after 4k context length pretraining. SWA/GLA/GDN-NoPE hybrids achieve stronger long-context language modeling.}
  \label{fig-main_ppl-hh}
\end{figure}

\begin{figure}[!tb]
  \noindent
  \centering
  \begin{subfigure}[b]{0.48\linewidth}
    \centering
    \includegraphics[width=\linewidth]{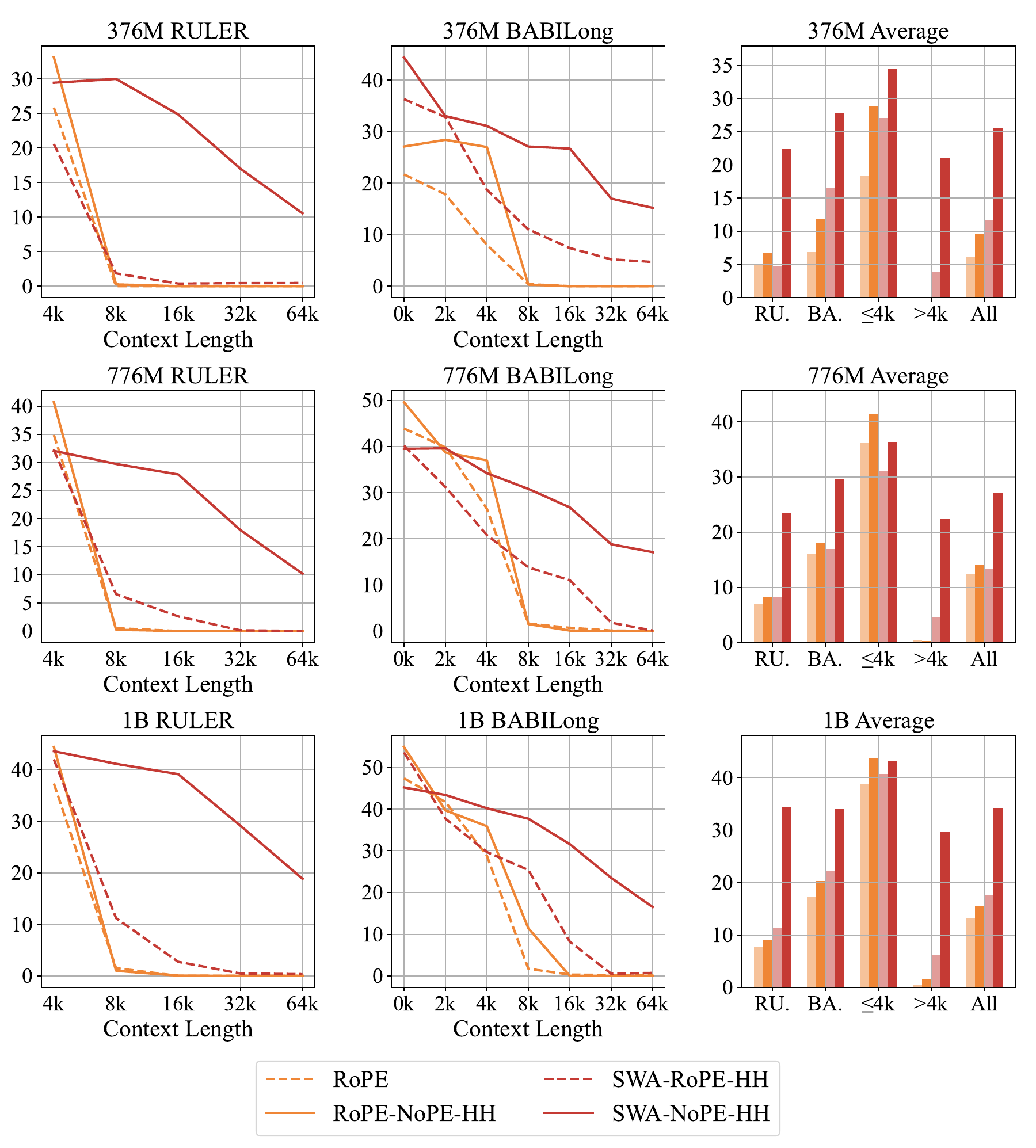}
    \caption{Results on Full Attention and SWA Hybrid.}
  \end{subfigure}
  \hfill
  \begin{subfigure}[b]{0.48\linewidth}
    \centering
    \includegraphics[width=\linewidth]{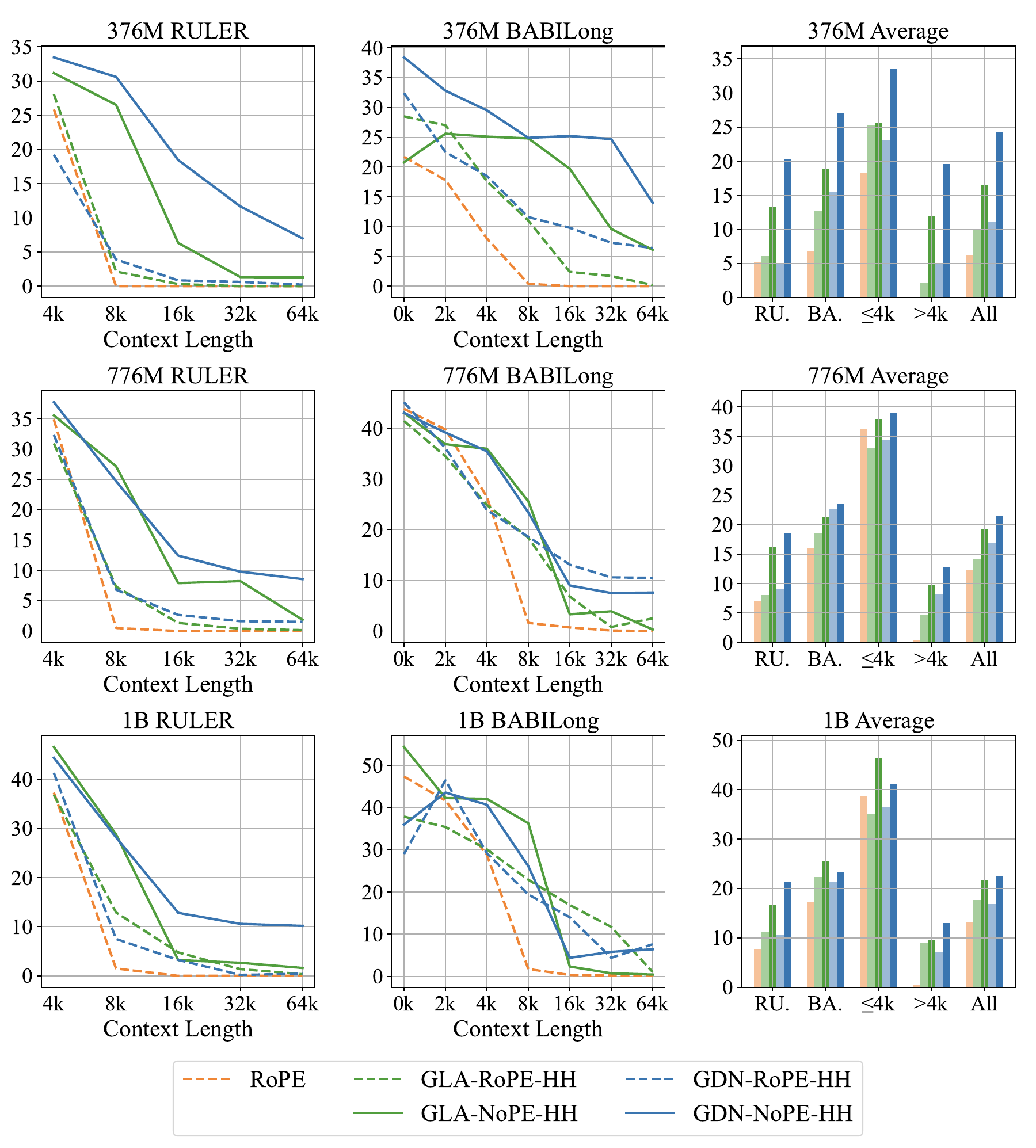}
    \caption{Results on Full Attention and GLA/GDN Hybrid.}
  \end{subfigure}
  \caption{Comparison of long-context performance under RULER and BABILong benchmarks among different head-wise hybrid models after 4k context length pretraining. SWA/GLA/GDN-NoPE hybrids achieve both stronger long-context downstream performance within and beyond the training context length.}
  \label{fig-main_ruler-hh}
\end{figure}

Figure~\ref{fig-main_ppl-lh} and Figure~\ref{fig-main_ruler-lh} show the results of layer-wise hybrids, while Figure~\ref{fig-main_ppl-hh} and Figure~\ref{fig-main_ruler-hh} show those of head-wise hybrids. Compared with the RoPE-only full attention model, the hybrid models exhibit lower training loss and perplexity within the training length, even for RoPE-NoPE hybrid full attention models~\citep{yang2026rope}. In SWA hybrids and LA hybrids, removing the position embedding in the full attention layer still enhances long-context performance and length extrapolation~\citep{puvvada2025swan,team2024jamba}, in both layer-wise and head-wise hybrids. 

\begin{figure}[!tb]
  \noindent
  \centering
  \begin{subfigure}[b]{0.48\linewidth}
    \centering
    \includegraphics[width=\linewidth]{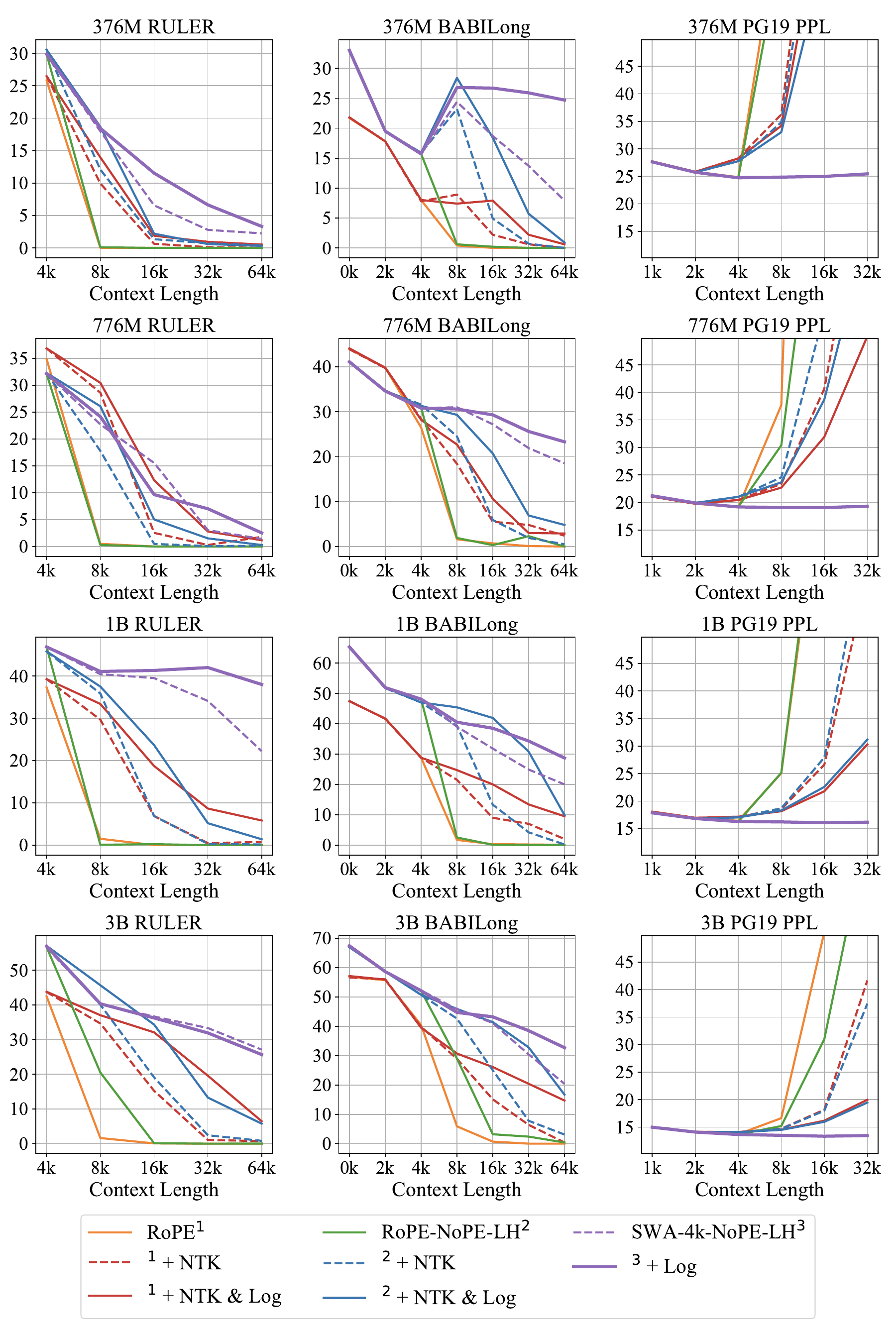}
    \caption{Results on Layer-wise Hybrid Model (LH).}
  \end{subfigure}
  \hfill
  \begin{subfigure}[b]{0.48\linewidth}
    \centering
    \includegraphics[width=\linewidth]{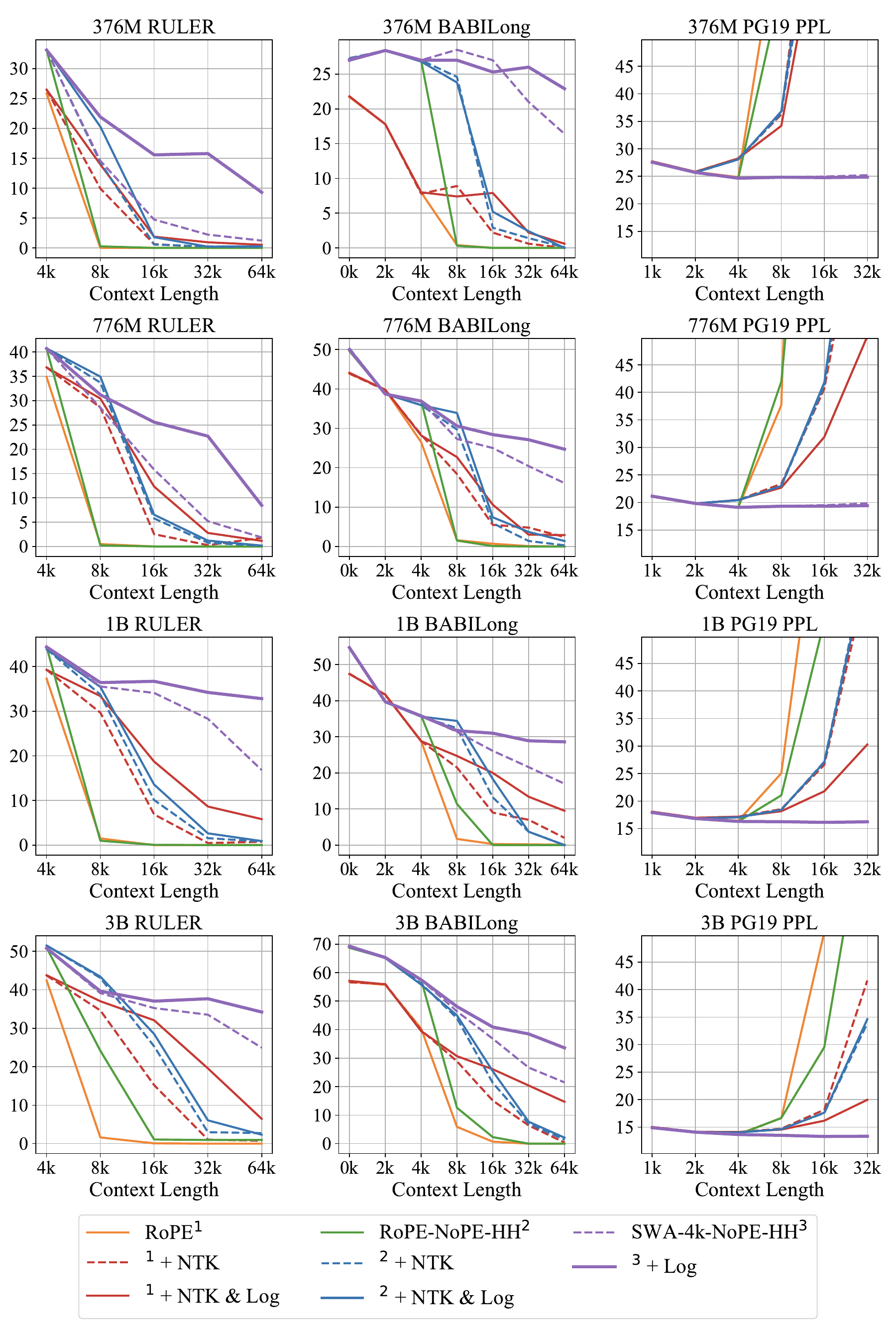}
    \caption{Results on Head-wise Hybrid Model (HH).}
  \end{subfigure}
  \caption{Comparison of long-context performance between RoPE-only and RoPE-NoPE hybrid full-attention models. Both layer-wise and head-wise RoPE-NoPE hybrids have better long-context performance and can achieve effective length extrapolation directly or under log-scaled NoPE extrapolation.}
  \label{fig-matthew-fa}
\end{figure}

We also verify the effectiveness of RoPE-NoPE extrapolation in an SWA-NoPE hybrid manner~\citep{yang2026rope,puvvada2025swan} rather than positional interpolation~\citep{fixedNTK,dynamicNTK,chen2023extending,pengyarn}. For RoPE attention, we retain only the attention window of the nearest pretraining context length, 4k. For NoPE attention, we use a log-scaled attention extrapolation, increasing the attention-logit scale to prevent attention entropy from increasing as the context grows, with the log base equal to the pretraining context length~\citep{log,liuscaling}. As shown in Figure~\ref{fig-matthew-fa}, this hybrid extrapolation method significantly outperforms RoPE or RoPE-NoPE hybrids using Dynamic NTK interpolation~\citep{dynamicNTK,liuscaling}.

This observation motivates \textbf{Takeaway 1} in this paper, \textbf{From Hybrid Attention to Hybrid Position}, because the gate mechanism in linear attention can be viewed as inducing a data-dependent position bias~\citep{lin2025forgetting,zhong2025understanding,yang2025path}. In hybrid models, the hybrid position between NoPE and other position biases plays an important role in long-context performance, and we will provide a detailed analysis in Section~\ref{sec-hybrid} and Section~\ref{sec-la}.

\begin{tcolorbox}[colback=blue!5!white,colframe=blue!75!black,title=\paragraph{Takeaway 1: From Hybrid Attention to Hybrid Position}\ ]
Hybrid models improve long-context performance and length extrapolation through a hybrid position of NoPE and other position biases. Even in RoPE-NoPE hybrid models, training-free effective length extrapolation is achieved for LLMs pretrained with full attention.
\end{tcolorbox}

\subsection{Long-Context Training}

\begin{figure}[!thb]
  \noindent
  \centering
  \begin{subfigure}[b]{0.48\linewidth}
    \centering
    \includegraphics[width=\linewidth]{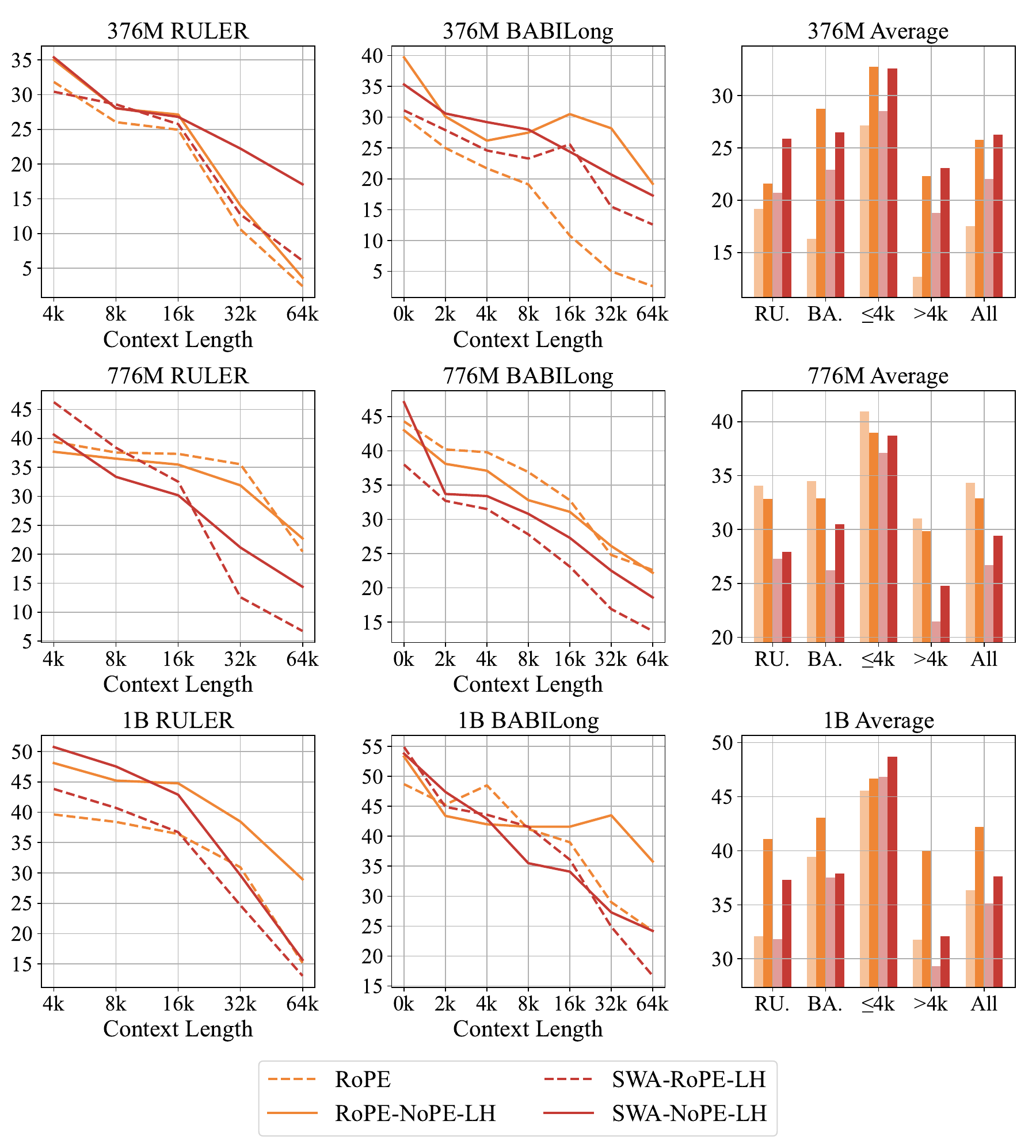}
    \caption{Results on Full Attention and SWA Hybrid.}
  \end{subfigure}
  \hfill
  \begin{subfigure}[b]{0.48\linewidth}
    \centering
    \includegraphics[width=\linewidth]{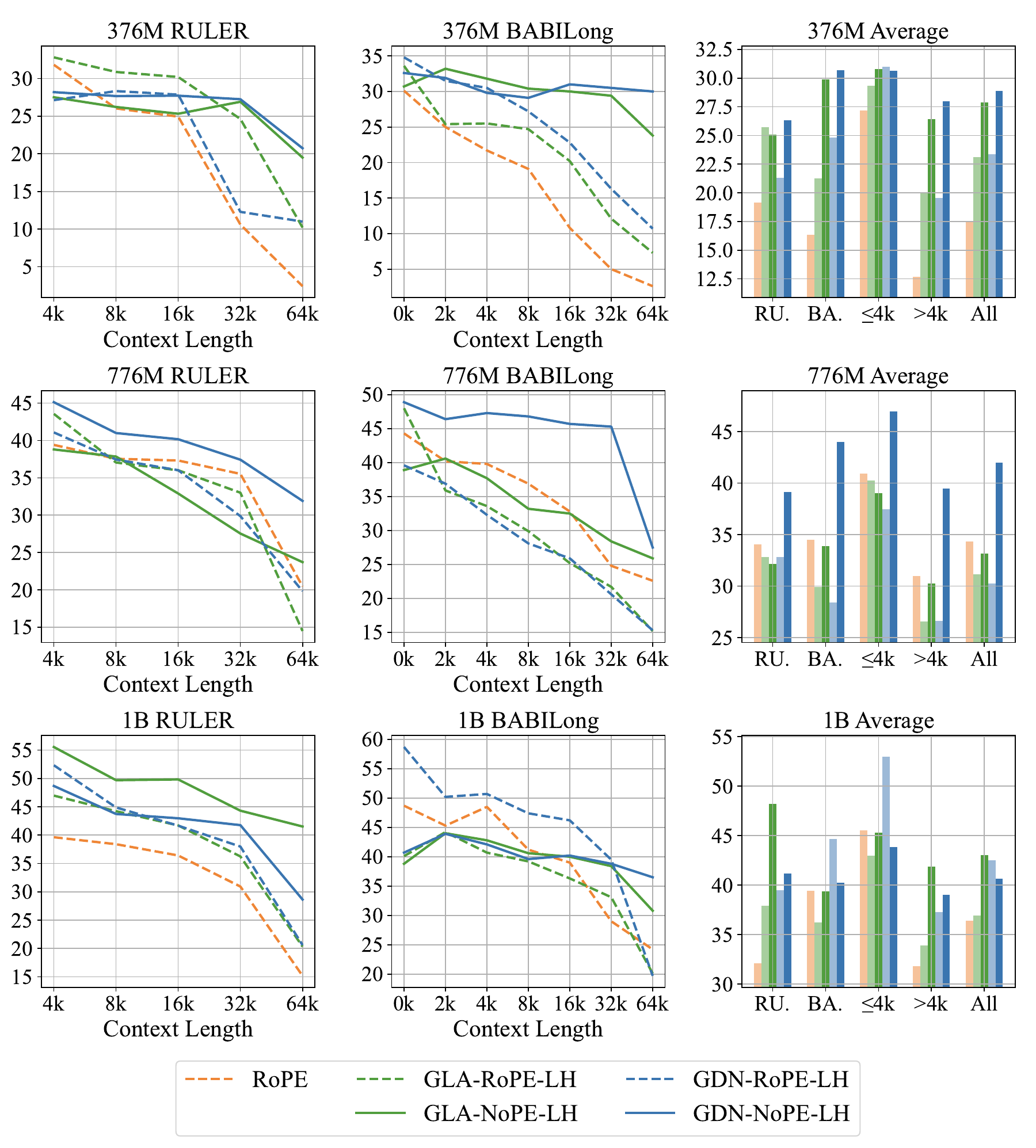}
    \caption{Results on Full Attention and GLA/GDN Hybrid.}
  \end{subfigure}
  \caption{Comparison of long-context performance under RULER and BABILong benchmarks among different layer-wise hybrid models after 32k context length continual pretraining.}
  \label{fig-main_ruler-lh-long}
\end{figure}

\begin{figure}[!thb]
  \noindent
  \centering
  \begin{subfigure}[b]{0.48\linewidth}
    \centering
    \includegraphics[width=\linewidth]{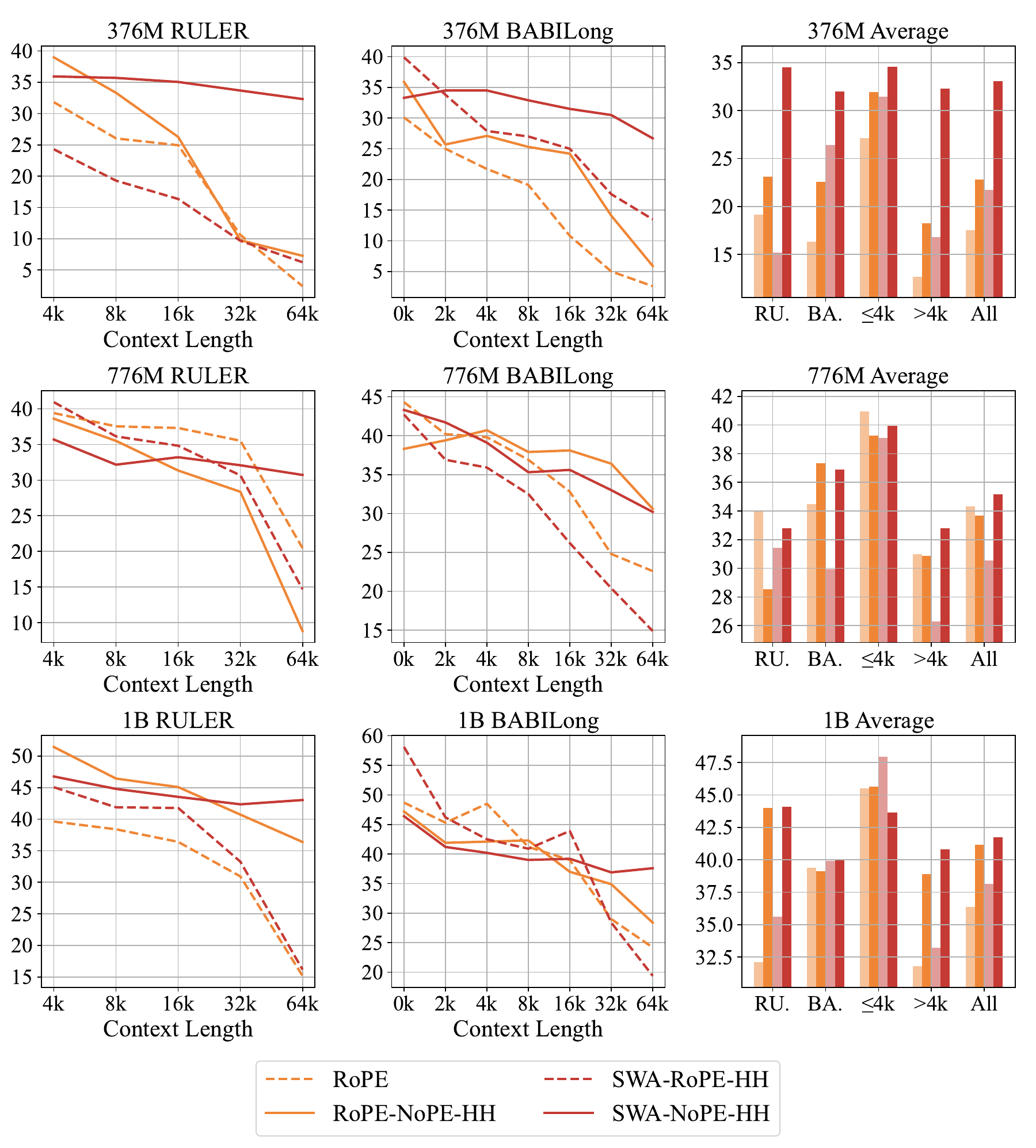}
    \caption{Results on Full Attention and SWA Hybrid.}
  \end{subfigure}
  \hfill
  \begin{subfigure}[b]{0.48\linewidth}
    \centering
    \includegraphics[width=\linewidth]{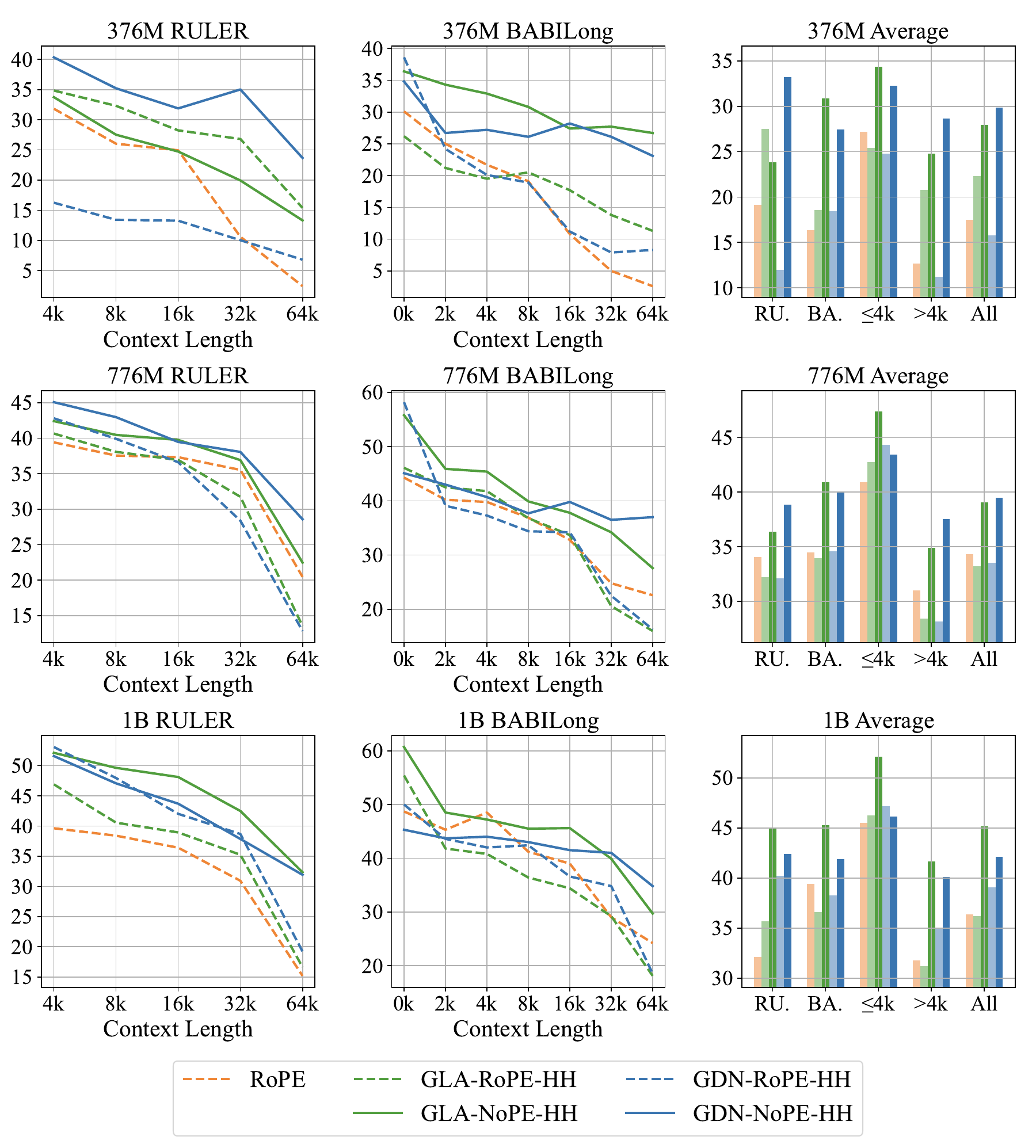}
    \caption{Results on Full Attention and GLA/GDN Hybrid.}
  \end{subfigure}
  \caption{Comparison of long-context performance under RULER and BABILong benchmarks among different head-wise hybrid models after 32k context length continual pretraining.}
  \label{fig-main_ruler-hh-long}
\end{figure}

\begin{figure}[!tb]
  \noindent
  \centering
  \begin{subfigure}[b]{0.48\linewidth}
    \centering
    \includegraphics[width=\linewidth]{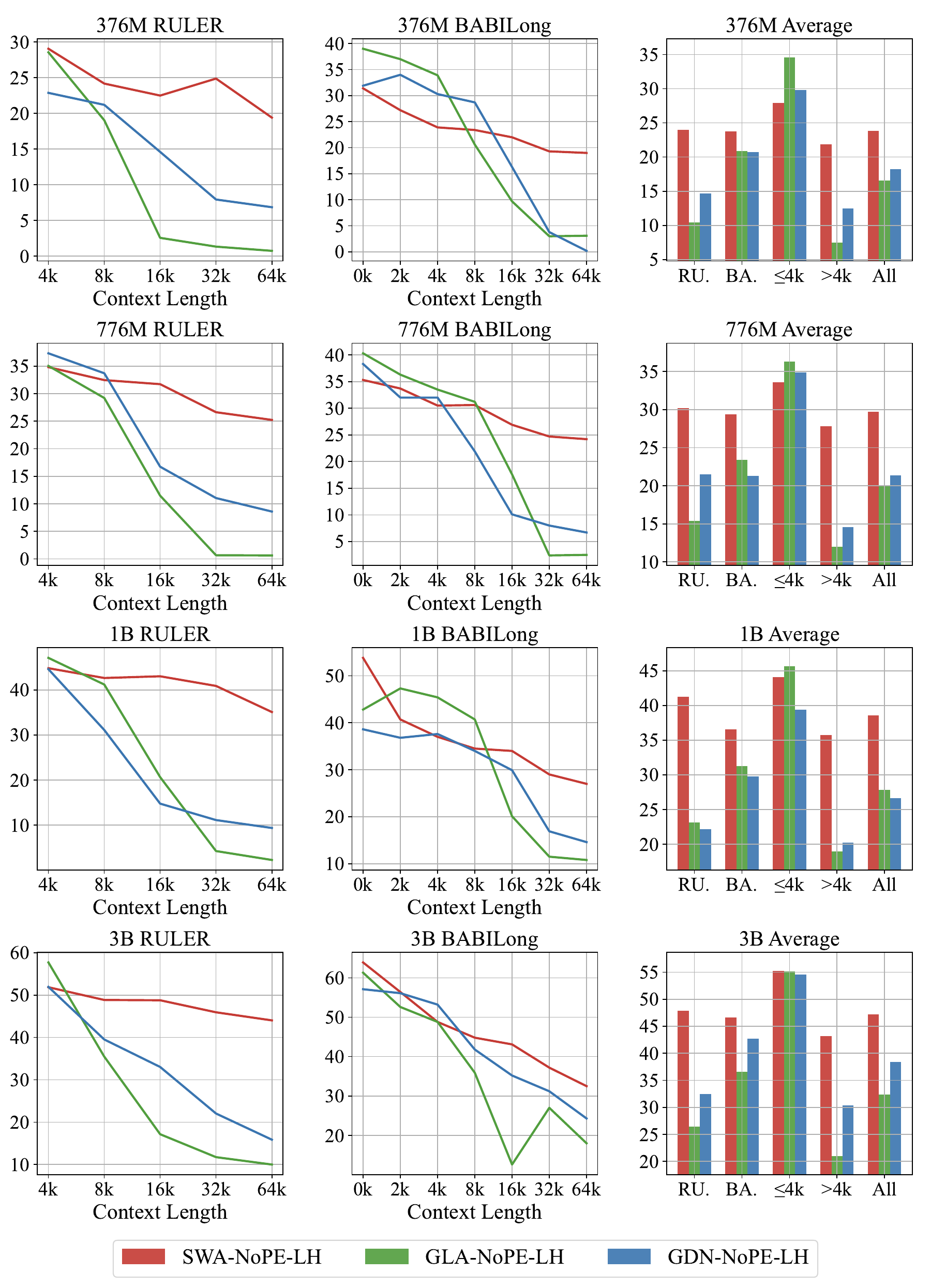}
    \caption{Performance of log-scale NoPE extrapolation.}
  \end{subfigure}
  \hfill
  \begin{subfigure}[b]{0.48\linewidth}
    \centering
    \includegraphics[width=\linewidth]{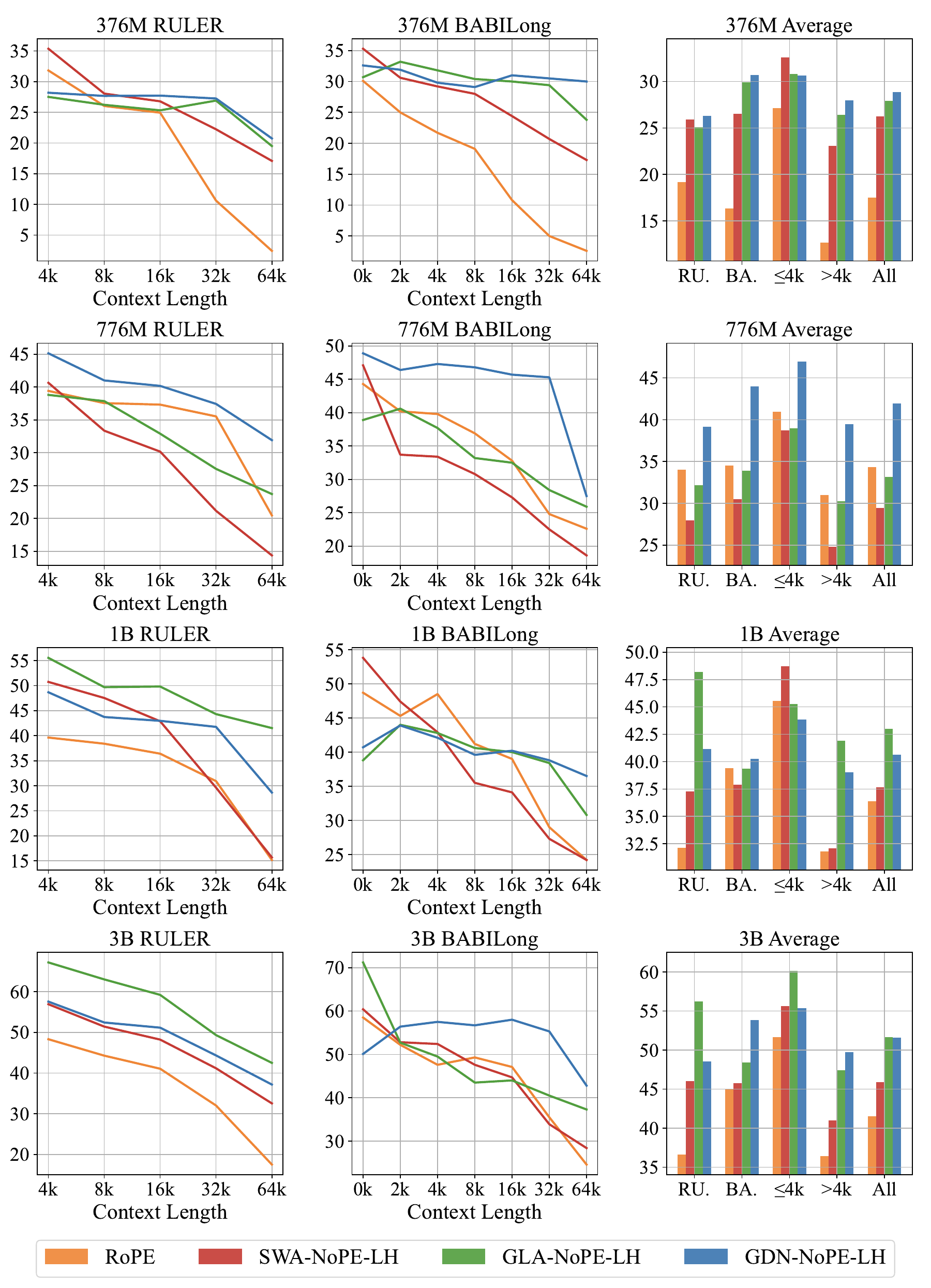}
    \caption{Performance after long-context continual pretraining.}
  \end{subfigure}
  \caption{Comparison of the long-context performance of the SWA/GLA/GDN-NoPE layer-wise hybrid models under log-scale NoPE extrapolation after short-context pretraining (left) and after long-context continual pretraining (right). The results of SWA hybrids exhibit a remarkable seesaw effect, as detailed in Takeaway 2.}
  \label{fig-main_seesaw-lh}
\end{figure}

\begin{figure}[!tb]
  \noindent
  \centering
  \begin{subfigure}[b]{0.48\linewidth}
    \centering
    \includegraphics[width=\linewidth]{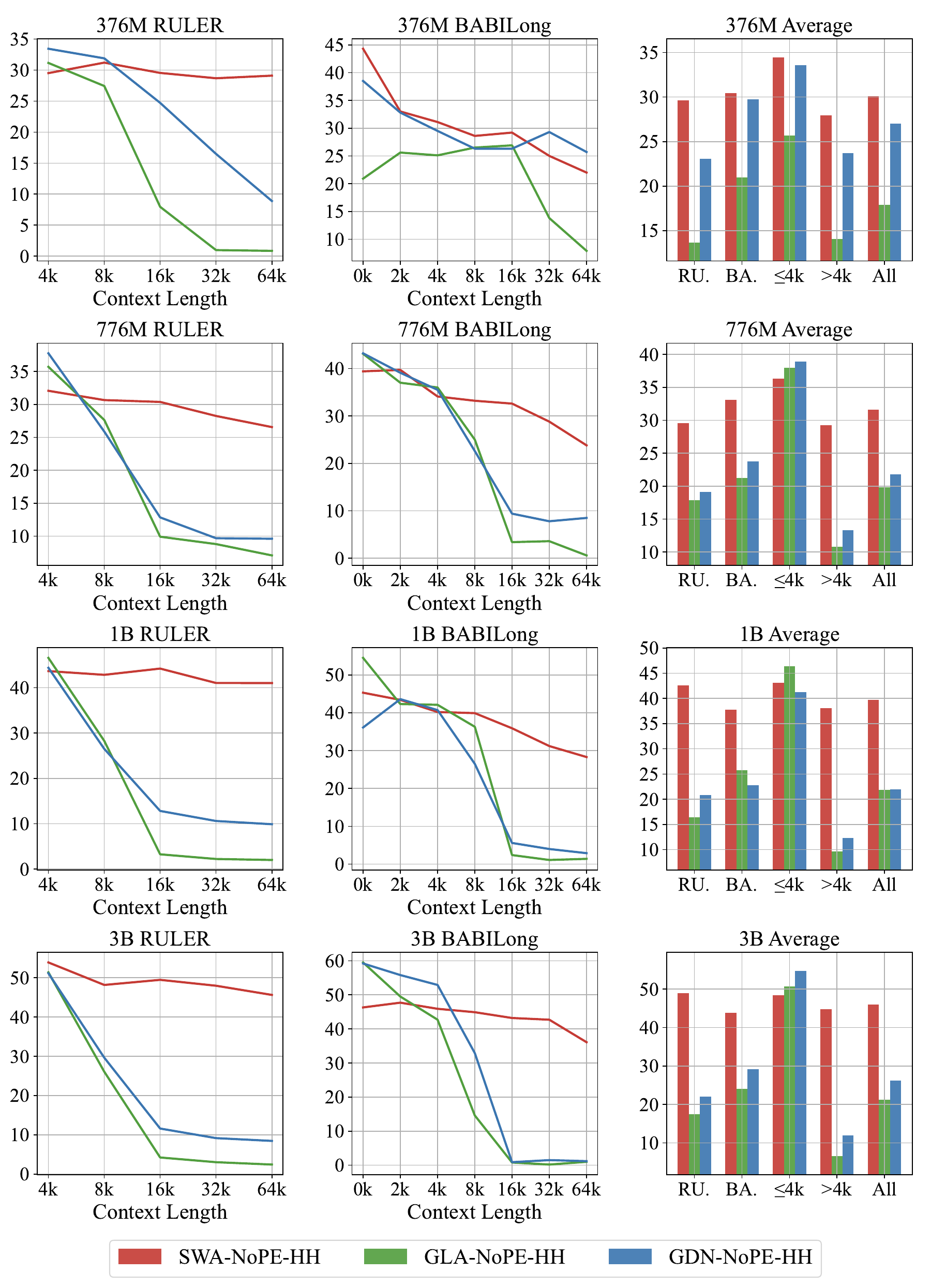}
    \caption{Performance of log-scale NoPE extrapolation.}
  \end{subfigure}
  \hfill
  \begin{subfigure}[b]{0.48\linewidth}
    \centering
    \includegraphics[width=\linewidth]{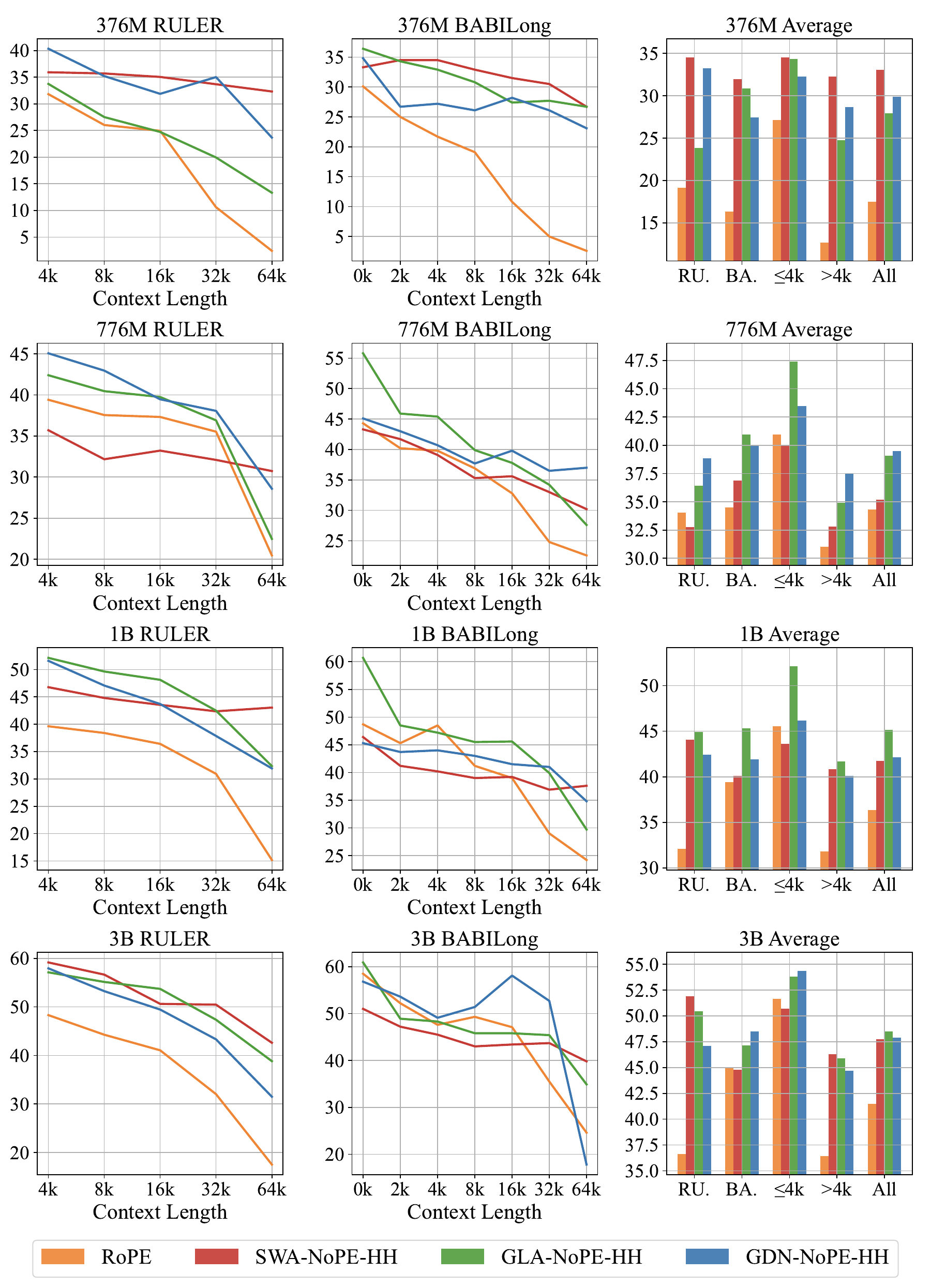}
    \caption{Performance after long-context continual pretraining.}
  \end{subfigure}
  \caption{Comparison of the long-context performance of the SWA/GLA/GDN-NoPE head-wise hybrid models under log-scale NoPE extrapolation after short-context pretraining (left) and after long-context continual pretraining (right). The results of SWA hybrids exhibit a remarkable seesaw effect, as detailed in Takeaway 2.}
  \label{fig-main_seesaw-hh}
\end{figure}

We further conduct long-context continual pretraining. Figure~\ref{fig-main_ruler-lh-long} shows results on layer-wise hybrids, while Figure~\ref{fig-main_ruler-hh-long} shows those on head-wise hybrids. Notably, there exists a very interesting phenomenon. SWA hybrids, which perform best in length extrapolation, underperform the LA hybrids after context extension, as shown in Figure~\ref{fig-main_seesaw-lh} and Figure~\ref{fig-main_seesaw-hh}, and may even fail to surpass their performance under log-scaled NoPE extrapolation, which is more remarkable in layer-wise hybrids. We call this phenomenon of SWA hybrids being outperformed by LA hybrids in long-context pretraining the \textbf{Seesaw Effect of Context Extension in SWA Hybrid}, which is \textbf{Takeaway 2} in this paper. In Section~\ref{sec-swa}, we will try to find the cause of this phenomenon and enhance the context extension of SWA hybrids through a deeper analysis.

\begin{tcolorbox}[colback=blue!5!white,colframe=blue!75!black,title=\paragraph{Takeaway 2: Seesaw Effect in Context Extension of SWA Hybrid}\ ]
For the SWA-NoPE hybrid model, the advantage in long-context performance over the LA-NoPE hybrid is inverted after long-context continual pretraining. Moreover, this phenomenon is more remarkable in the SWA-NoPE layer-wise hybrid model.
\end{tcolorbox}

In contrast, LA hybrids benefit more from context extension during long-context continual pretraining, but lag behind SWA hybrids in length extrapolation. In other words, compared with SWA hybrids, LA hybrids exhibit stronger fitting capability within the training context length, yet weaker generalization beyond it. This phenomenon is consistent with the No-Free-Lunch principle in machine learning and neural network design, leading to \textbf{Takeaway 3} in this paper, \textbf{No-Free-Lunch Effect in LA Hybrids}. In section~\ref{sec-la}, we will try to figure out the position bias in LA hybrids and enhance their length extrapolation.

\begin{tcolorbox}[colback=blue!5!white,colframe=blue!75!black,title=\paragraph{Takeaway 3: No-Free-Lunch Effect in Length Extrapolation of LA Hybrid}\ ]
The GLA/GDN-NoPE hybrid model shows its disadvantages in extrapolation (OOD) compared to the SWA-NoPE hybrid model and advantages within the training context (in-domain). Particularly, 
\begin{itemize}
    \item GLA/GDN-NoPE underperforms SWA-NoPE under direct extrapolation and log-scaled NoPE extrapolation after short-context pretraining, while
    \item GLA/GDN-NoPE outperforms SWA-NoPE within the training context length in both short- and long-context pretraining.
\end{itemize}
\end{tcolorbox}

\section{Discussion on Hybrid Position}\label{sec-hybrid}

The effectiveness of hybrid models raises thought-provoking questions: why do RoPE/NoPE full attention, sliding-window attention, and linear attention each have shortcomings, but hybrid models can achieve better long-context performance within and beyond the training context length? What specific roles do NoPE and other position-biased attention play in the same model? 

\subsection{Ratio Ablation and Noise Analysis}

To answer these questions, we first analyze NoPE head-wise hybrid models with different NoPE ratios. We choose head-wise hybrids because they let NoPE and other attention types collaborate in parallel within the same layer, making them better suited for role attribution than layer-wise hybrids, and avoiding the impact of layer order and functional differences at different depths.

\begin{figure}[!b]
  \noindent
  \centering
  \begin{subfigure}[b]{0.49\linewidth}
    \centering
    \includegraphics[width=\linewidth]{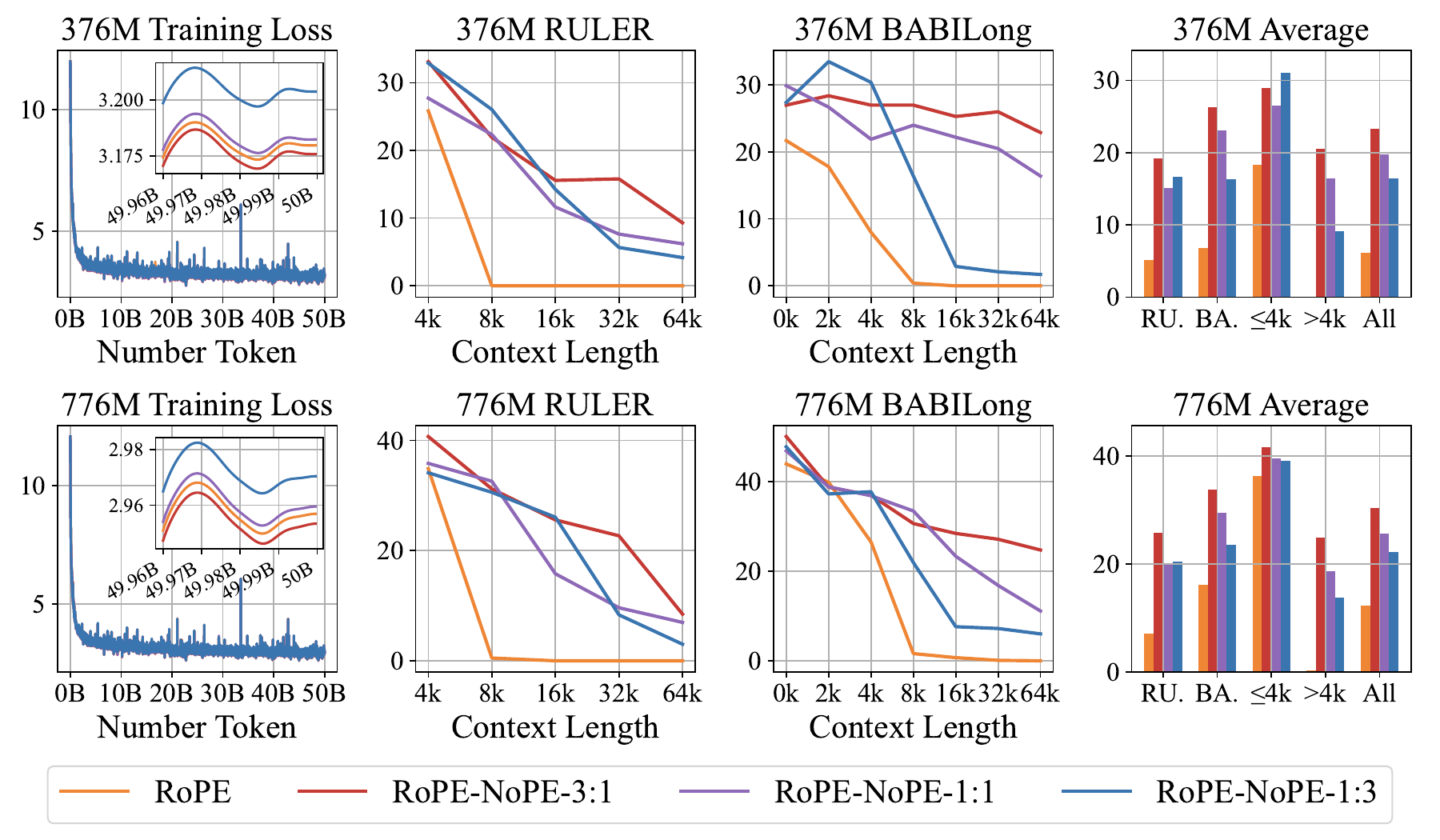}
    \caption{Results on RoPE-NoPE hybrids.}
  \end{subfigure}
  \hfill
  \begin{subfigure}[b]{0.49\linewidth}
    \centering
    \includegraphics[width=\linewidth]{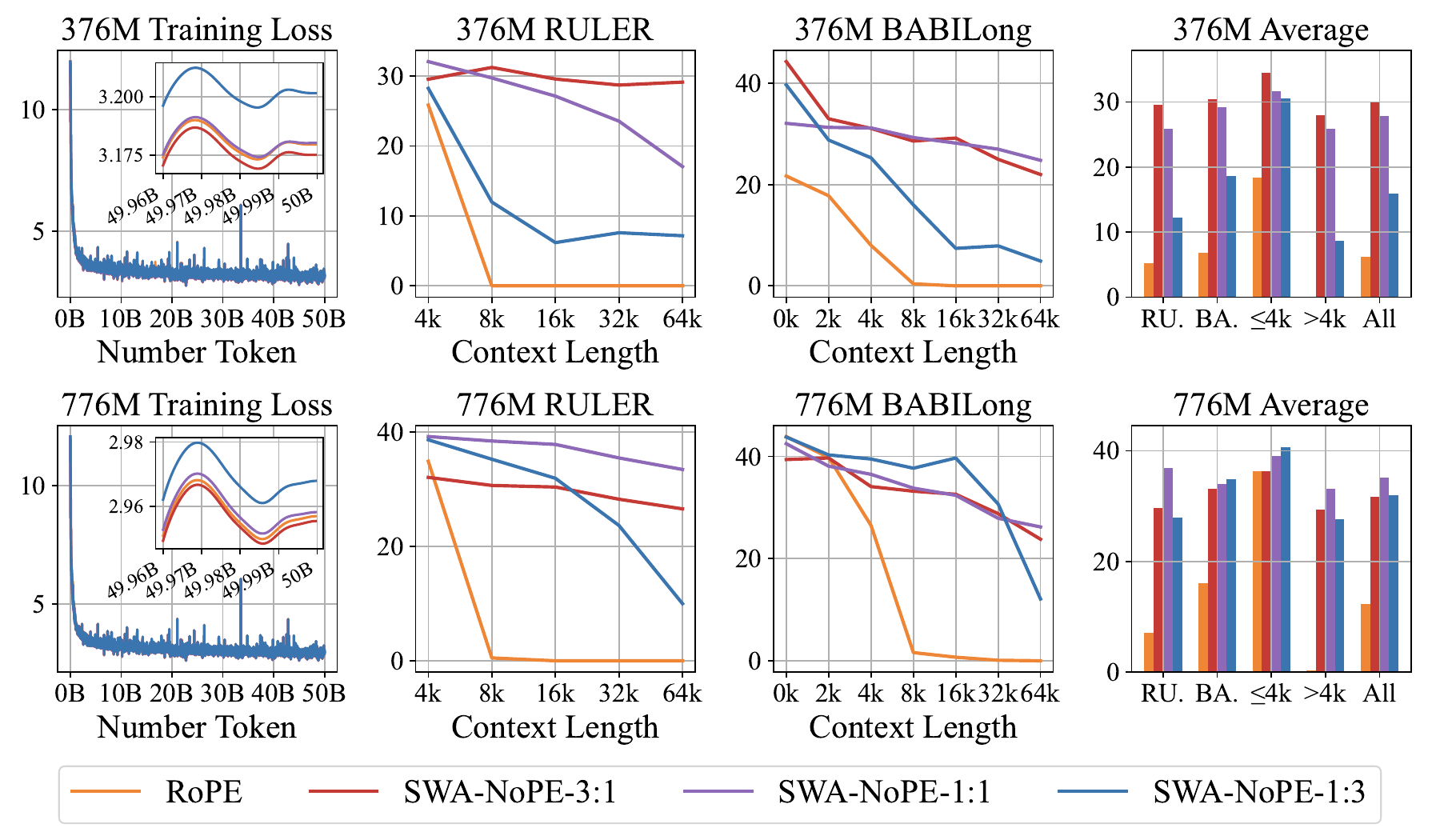}
    \caption{Results on SWA-NoPE hybrids.}
  \end{subfigure}
  \vskip0.6\baselineskip
  \begin{subfigure}[b]{0.49\linewidth}
    \centering
    \includegraphics[width=\linewidth]{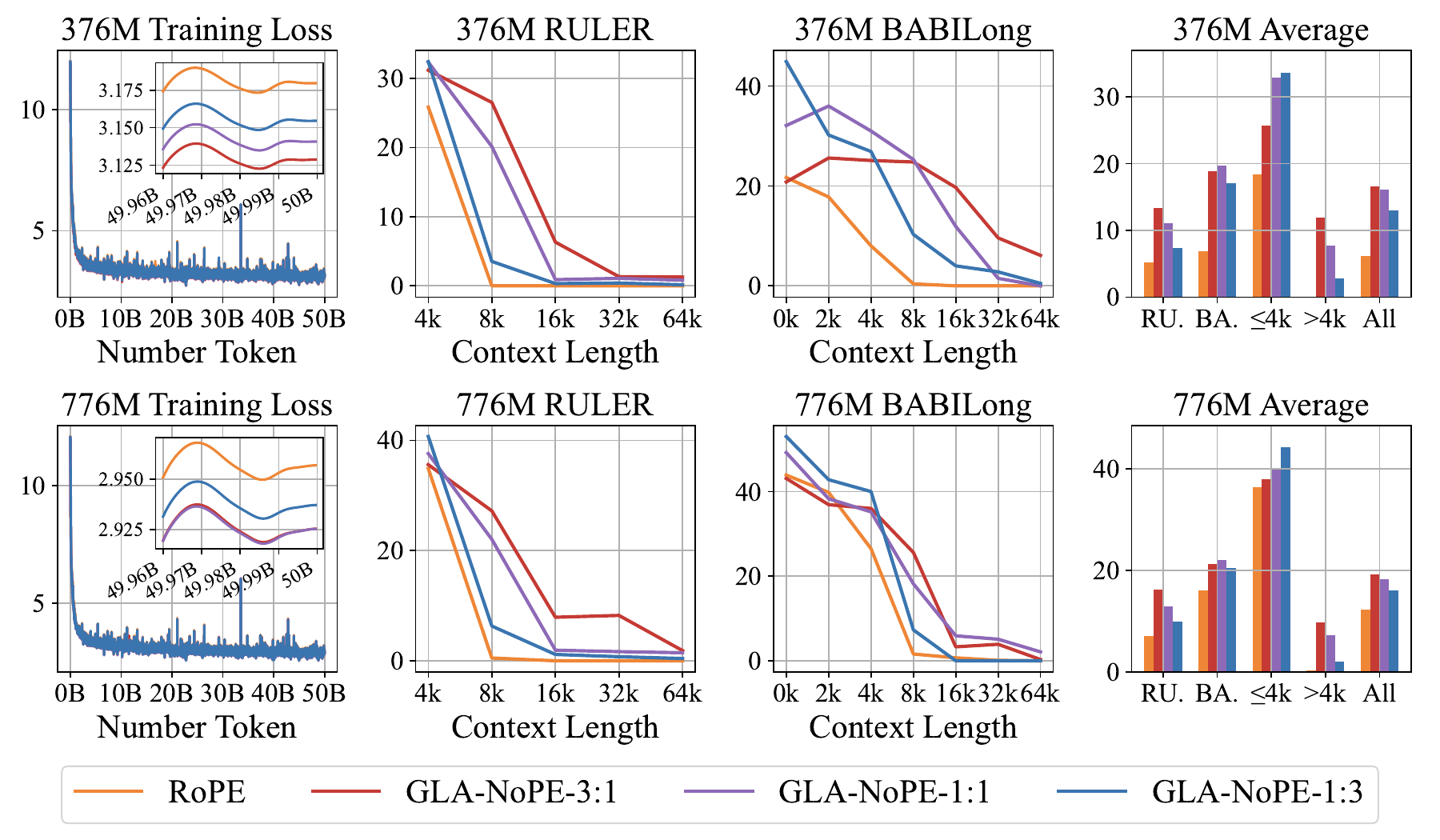}
    \caption{Results on GLA-NoPE hybrids.}
  \end{subfigure}
  \hfill
  \begin{subfigure}[b]{0.49\linewidth}
    \centering
    \includegraphics[width=\linewidth]{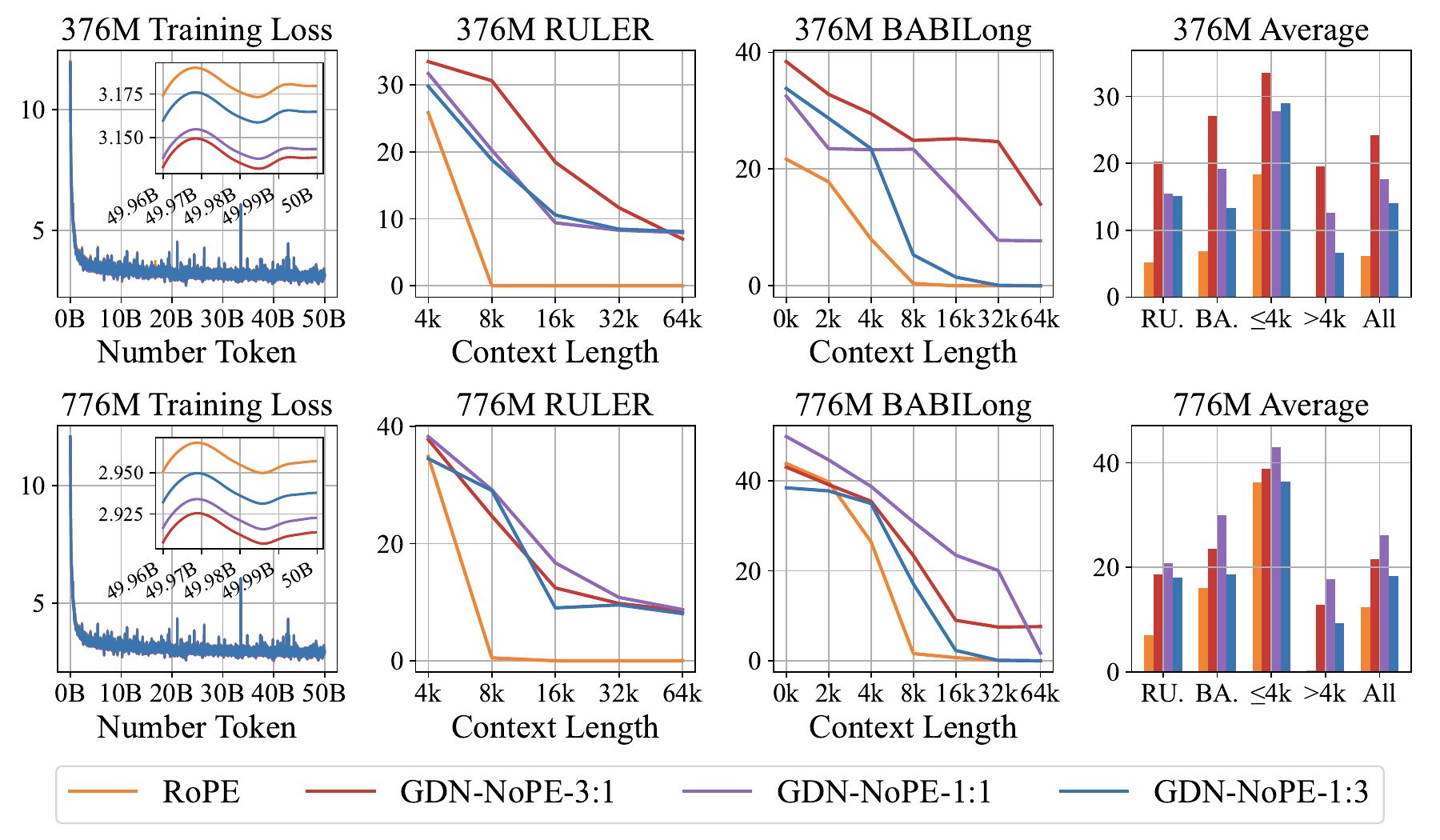}
    \caption{Results on GDN-NoPE hybrids.}
  \end{subfigure}
  \caption{Comparison of training loss, RULER, and BABILong among different hybrid ratios in head-wise hybrid models.}
  \label{fig-hybrid-ratio_hh}
\end{figure}

We compare RoPE-NoPE-HH, SWA-NoPE-HH, GLA-NoPE-HH, and GDN-NoPE-HH with different hybrid ratios, 3:1, 1:1, and 1:3, using the same setup for pretraining on model sizes of 376M and 776M. We only conduct the first stage of pretraining with a context length of 4k, and apply a sliding window with a window size of 4k for RoPE attention in RoPE-NoPE-HH~\citep{yang2026rope,puvvada2025swan}, as well as log attention scaling for NoPE attention in all hybrids to achieve better length extrapolation~\citep{log,liuscaling}. The results are shown in Figure~\ref{fig-hybrid-ratio_hh}. The results indicate that a 3:1 RoPE-NoPE ratio performs best within and beyond the training length in most cases. While NoPE attention plays an important role in long-context performance and extrapolation capability, more NoPE attention actually weakens the extrapolation effect.

\begin{figure}[!tb]
  \noindent
  \centering
  \begin{subfigure}[b]{0.48\linewidth}
    \centering
    \includegraphics[width=\linewidth]{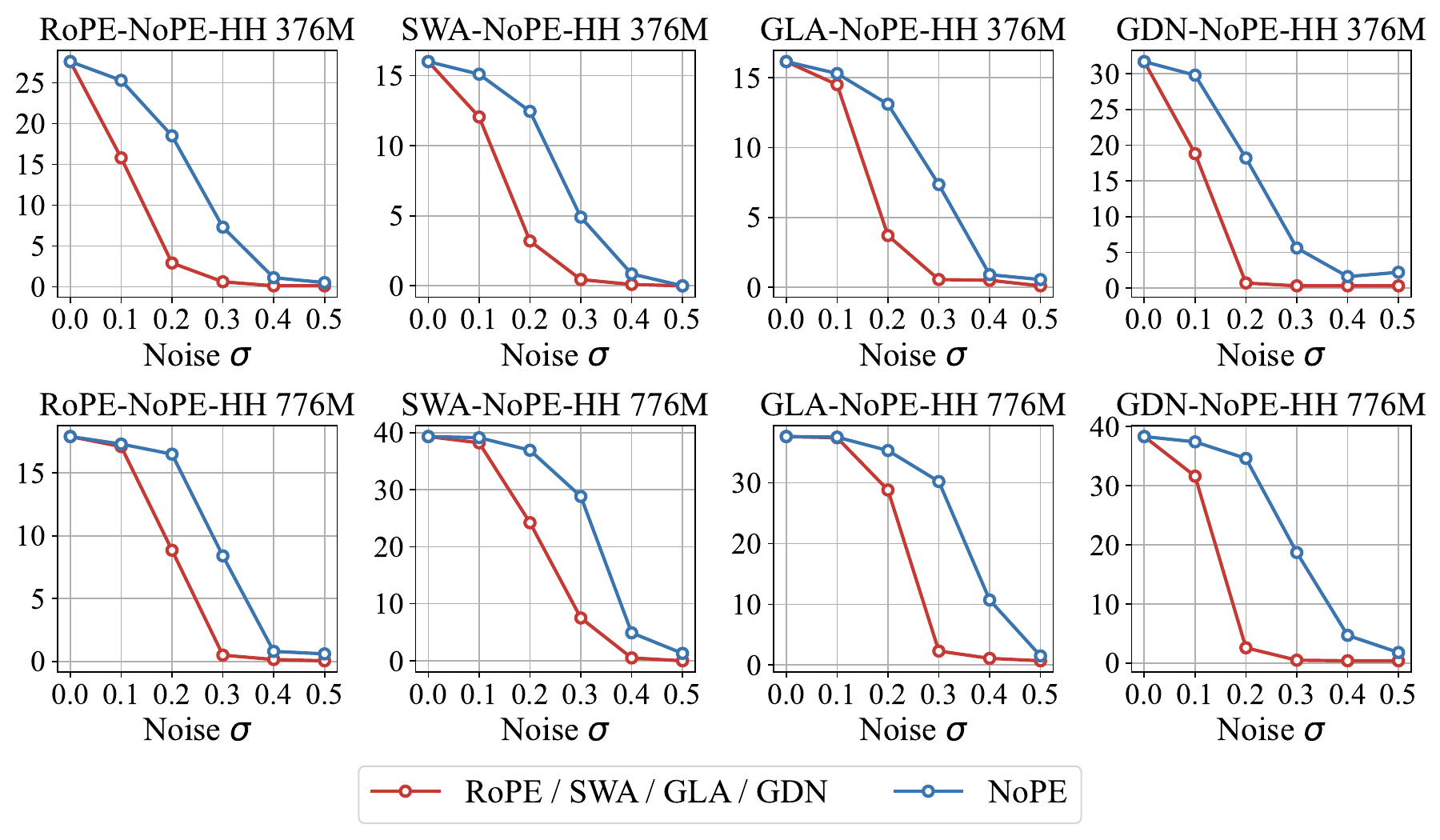}
    \caption{Results on RULER Benchmark.}
  \end{subfigure}
  \hfill
  \begin{subfigure}[b]{0.48\linewidth}
    \centering
    \includegraphics[width=\linewidth]{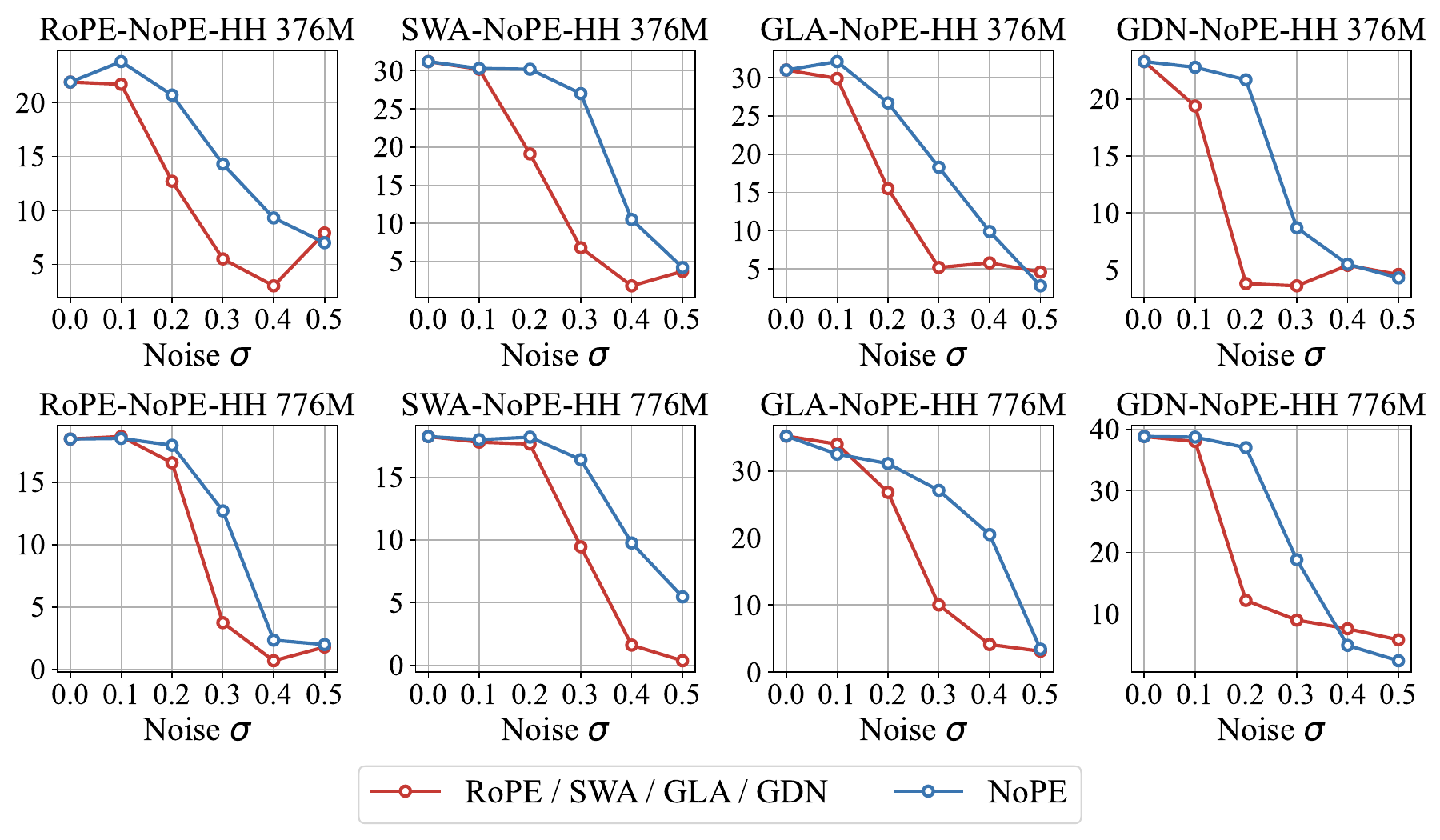}
    \caption{Results on BABILong Benchmark.}
  \end{subfigure}
  \caption{Results of noise experiments on RULER and BABILong benchmark. Adding Gaussian noise to RoPE full attention or SWA/GLA/GDN degrades long-context performance more severely than the same perturbation applied to NoPE full attention in the corresponding hybrid models.}
  \label{fig-hybrid-noise}
\end{figure}

Particularly, we conduct noise experiments on the 1:1 NoPE head hybrids. We add Gaussian noise to NoPE heads or other position-biased heads and observe which part of the noise has a greater impact on downstream tasks such as RULER and BABILong in 4k context length as the variance increases. Surprisingly, as shown in Figure~\ref{fig-hybrid-noise}, adding noise to the position-biased heads, including RoPE/SWA/GLA/GDN heads, has a greater impact on downstream long-context tasks. This contradicts the intuition that NoPE is more responsible for global information retrieval, tending to dominate long-context performance, prompting us to investigate further: what role does NoPE full attention play in the hybrid models, and why could adding only a small amount of NoPE attention improve the performance of the original RoPE model?

\subsection{Attention Analysis of Hybrid Position}

\begin{figure}[!tb]
  \noindent
  \centering
  \begin{subfigure}[b]{0.403\linewidth}
    \centering
    \includegraphics[width=\linewidth]{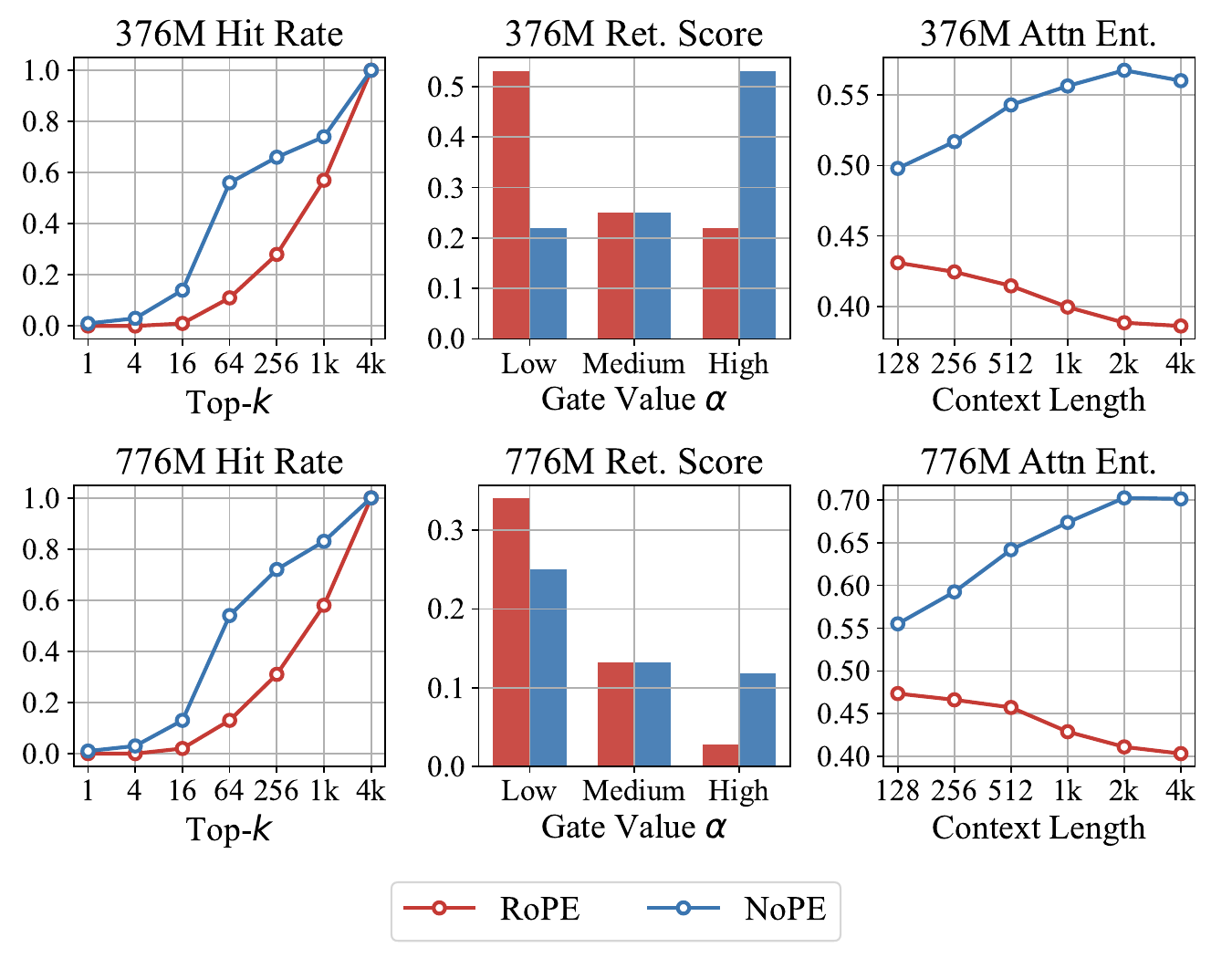}
    \caption{Comparison within 1:1 RoPE-NoPE-HH.}
  \end{subfigure}
  \hfill
  \begin{subfigure}[b]{0.587\linewidth}
    \centering
    \includegraphics[width=\linewidth]{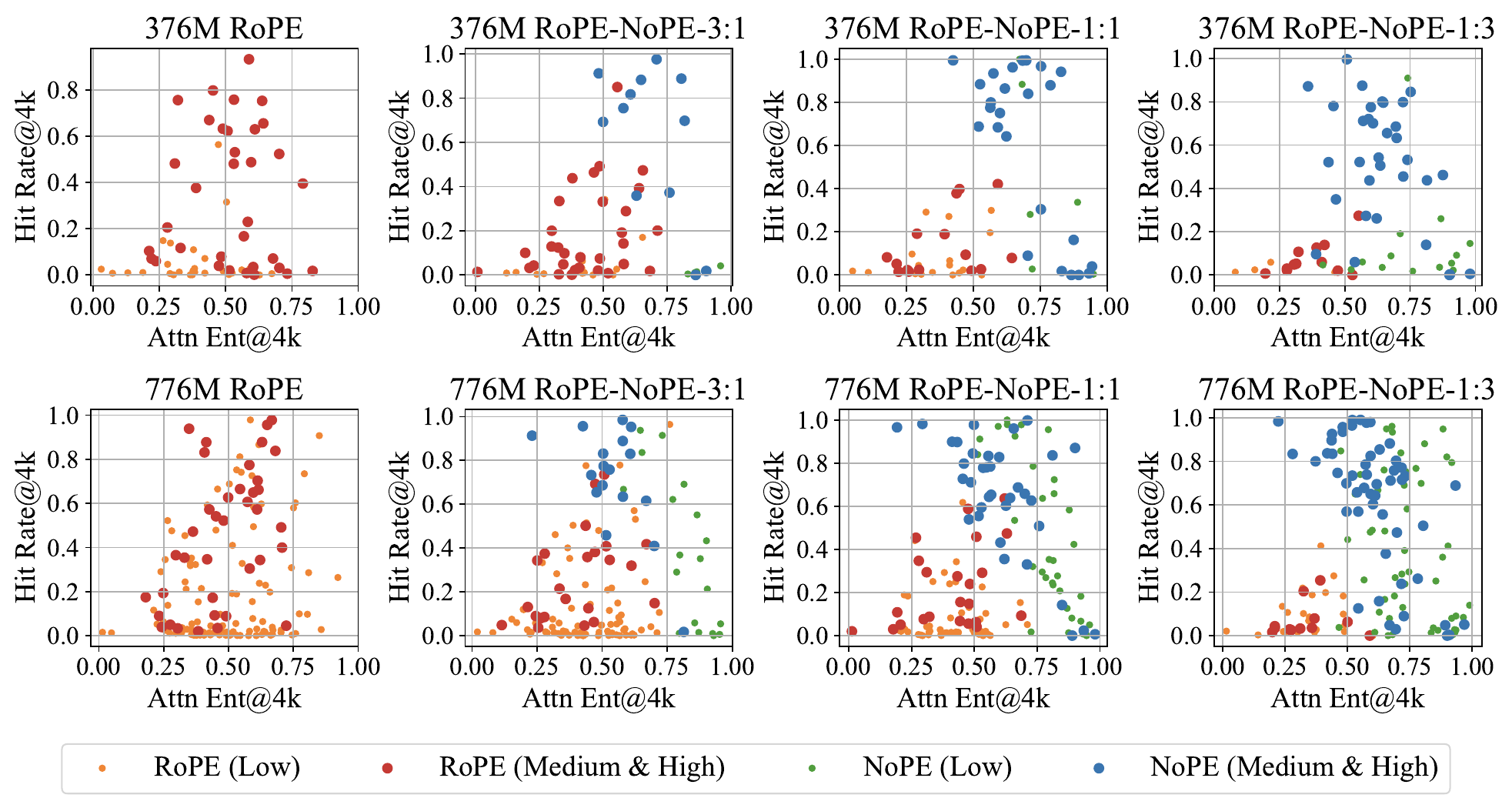}
    \caption{Comparison among RoPE-NoPE-HH}
  \end{subfigure}
  \caption{The results of attention analysis within 1:1 RoPE-NoPE-HH and among RoPE-NoPE-HH with different hybrid ratio. NoPE full attention shows a stronger tendency for coarse global information localization, given the higher attention score hit rate and larger attention entropy, compared with RoPE full attention in RoPE-NoPE hybrids.}
  \label{fig-scatter-fa_hh}
\end{figure}

To analyze the role of NoPE full attention in the hybrid models more clearly, we carry out three more in-depth experiments on attention distribution for the RoPE-NoPE hybrids.
\begin{enumerate}
    \item \textbf{Hit Rate Experiment}. Input the prompt of the 4k SK1 task in RULER~\citep{hsieh2024ruler}, and calculate the probability that the top-$k$ attention score in RoPE or NoPE attention hits the position to be retrieved for the query tokens at the end of the input. This experiment determines whether RoPE or NoPE is more responsible for direct information localization. 
    \item \textbf{Retrieval Head Experiment}. Follow the definition of the retrieval head and streaming head in DuoAttention~\citep{xiao2024duoattention} and compare the retrieval scores in RoPE and NoPE attention. According to \citet{xiao2024duoattention}, a retrieval head is an attention head that preserves distant tokens in addition to the initial sink and the recent local tokens, whereas a streaming head primarily preserves the sink and local tokens. This experiment determines whether RoPE or NoPE reads distant content more frequently. 
    \item \textbf{Attention Entropy Experiment}: Count the attention entropy of RoPE and NoPE attention on the PG19 dataset~\citep{rae2019compressive}. We normalize the entropy between 0 and 1 by dividing by the maximum possible entropy value, the logarithm of the context length. This experiment determines whether RoPE or NoPE attention has a broader attention range. 
\end{enumerate} 
Since retrieval scores are concentrated near 0 within $(0,1)$, we divide them into three bins: high $[10^{-3},1)$, medium $[10^{-5},10^{-3})$, and low $(0,10^{-5})$, and report each proportion. The results are shown in Figure~\ref{fig-scatter-fa_hh}, where NoPE and RoPE show the biggest gap in the hit-rate plot when $k=64$. To better reflect the distinct roles of RoPE and NoPE attentions, we add a scatter plot where each point represents an attention head, using its attention entropy at 4k and its top-64 hit rate as its coordinates. In this \textbf{entropy-hit-rate diagram}, we also use different colors to distinguish the retrieval head and streaming head in RoPE and NoPE attention.

\begin{figure}[!tb]
  \noindent
  \centering
  \begin{subfigure}[b]{0.403\linewidth}
    \centering
    \includegraphics[width=\linewidth]{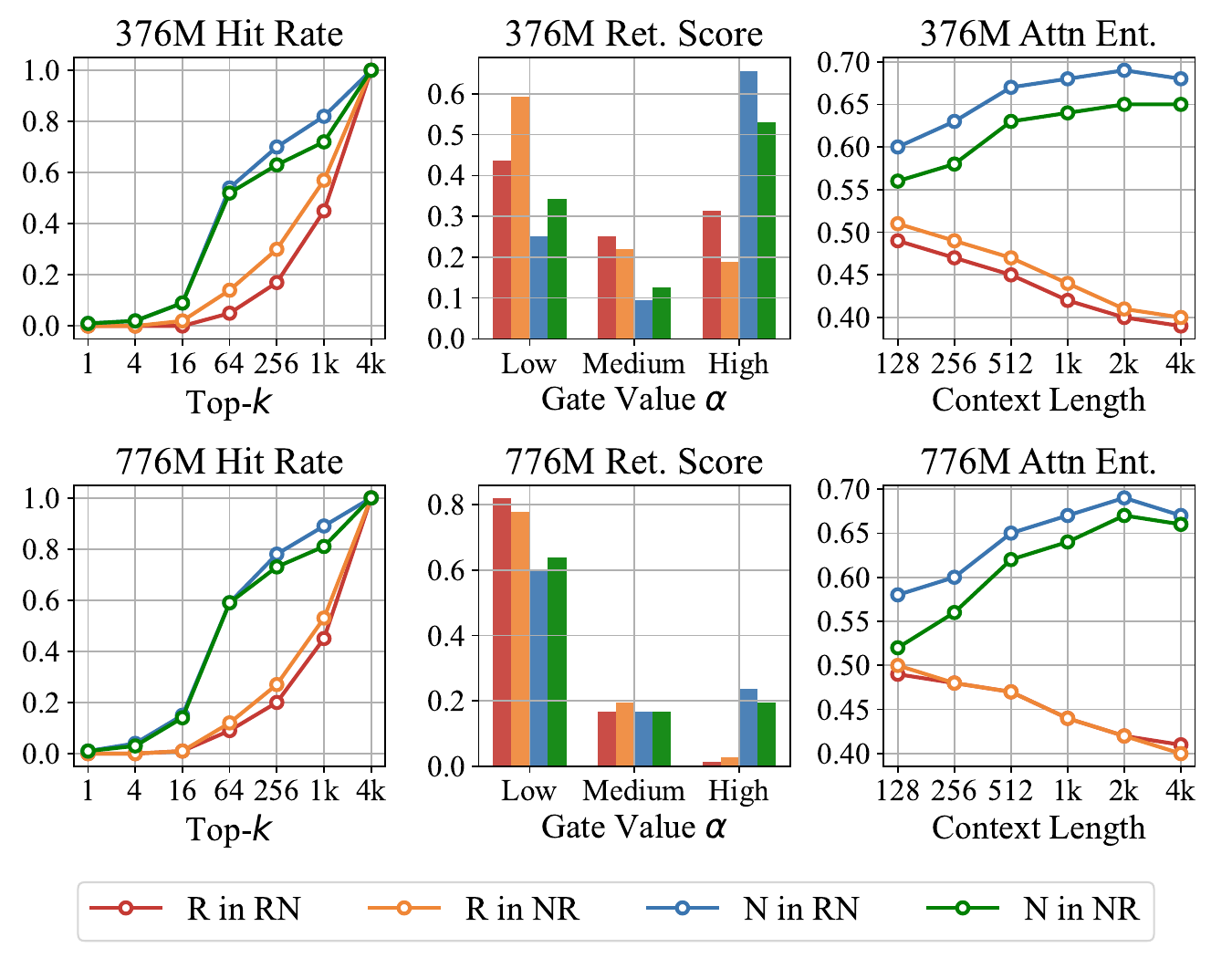}
    \caption{Comparison within 1:1 RoPE-NoPE-LH.}
  \end{subfigure}
  \hfill
  \begin{subfigure}[b]{0.587\linewidth}
    \centering
    \includegraphics[width=\linewidth]{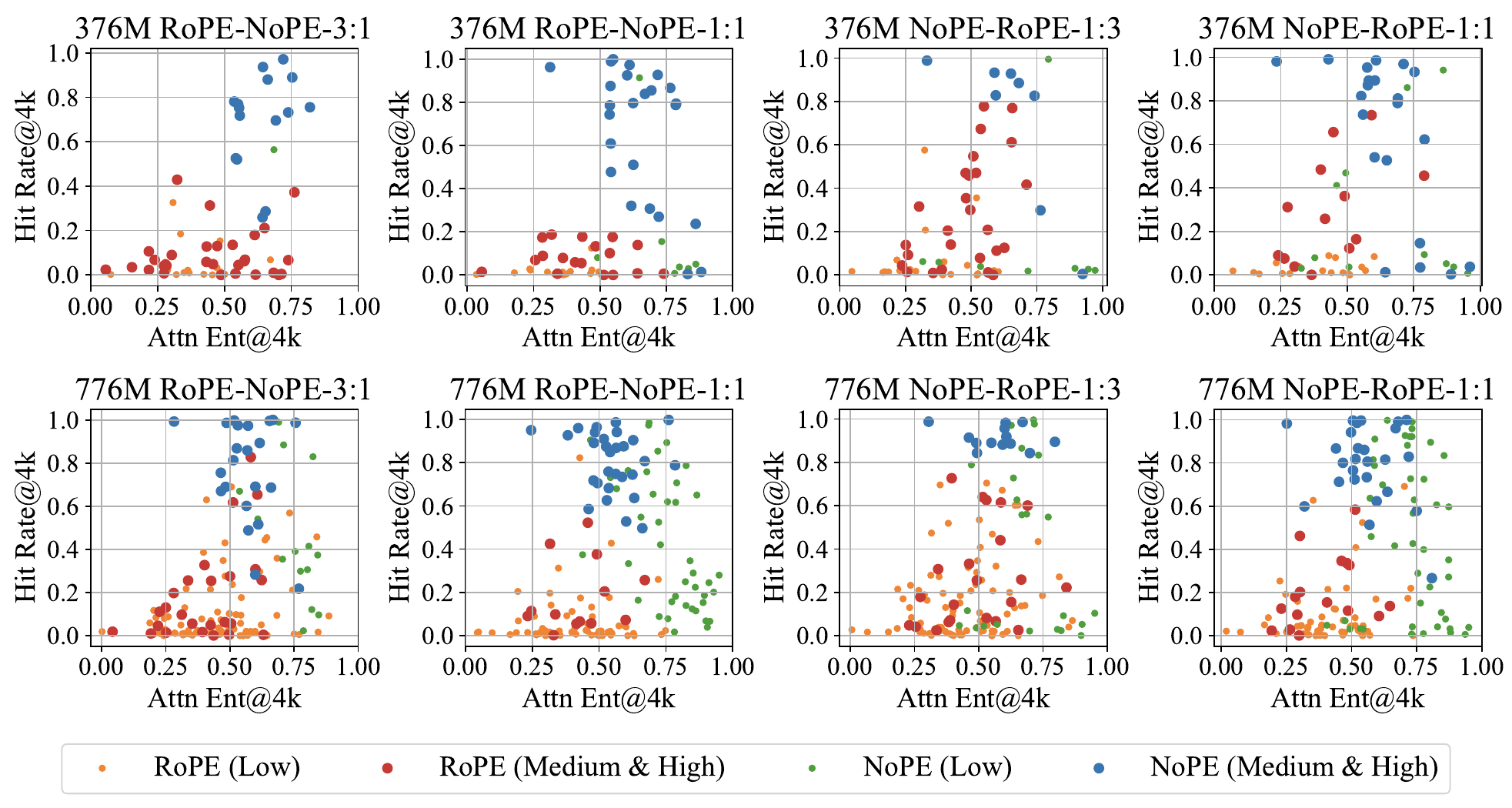}
    \caption{Comparison among RoPE-NoPE-LH}
  \end{subfigure}
  \caption{The results of attention analysis within 1:1 RoPE-NoPE-LH and among RoPE-NoPE-LH with different hybrid ratios. The tidal effect of hybrid position is still applicable in RoPE-NoPE layer-wise hybrid models.}
  \label{fig-scatter-fa_lh}
\end{figure}

We first compare the RoPE-NoPE head-wise hybrid models with a 1:1 head mix. We find that \textbf{NoPE attention has a higher hit rate, a higher retrieval-head proportion, and a higher attention entropy, mainly occupying the strip area from the upper middle to the lower right} in the entropy-hit-rate diagram. In contrast, \textbf{RoPE attention has a lower hit rate, a higher streaming-head proportion, and a lower attention entropy, mainly occupying the sector area extending outward from the lower-left corner} in the entropy-hit-rate diagram. This indicates that \textbf{NoPE mainly supports coarse global localization} in the hybrid models, since it dominates the high-entropy region. However, the negative correlation between entropy and hit rate indicates that when NoPE achieves a medium or lower hit rate, its higher attention entropy may introduce more noise. That is why localization relying solely on NoPE is unreliable, and a higher NoPE ratio may imply worse performance. 

This is even more evident in comparing RoPE-NoPE hybrids with different RoPE-NoPE ratios. As the NoPE ratio increases, the distribution of RoPE attentions in the entropy-hit-rate diagram gradually shrinks towards the lower-left corner. In contrast, NoPE attentions occupy the high-hit-rate region and extend to the low-hit-rate and high-entropy region simultaneously. This is just like a tidal lane, where traffic may travel in either direction, depending on certain conditions. Therefore, we name this redistribution, under the collaboration between position-biased attention and NoPE attention, the \textbf{Tidal Effect of Hybrid Position}, which is \textbf{Takeaway 4} in this paper: position-biased and NoPE attention heads occupy distinct regions in the entropy-hit-rate diagram, and their functional boundaries shift with the hybrid ratio. 

\begin{tcolorbox}[colback=blue!5!white,colframe=blue!75!black,title=\paragraph{Takeaway 4: Tidal Effect in Collaboration of Hybrid Position}\ ]
 Effective collaboration of hybrid position relies on a minority of high-hit-rate NoPE attention with coarse aggregation and a majority of low-entropy position-biased attention (i.e., RoPE attention and gated linear attention) for noise reduction. In the entropy-hit-rate diagram, position-biased attention occupies a sector region extending from the lower-left corner, while NoPE attention occupies a strip region from the upper middle to the lower right, and their boundaries shift with the hybrid ratio.
\end{tcolorbox}

Notably, here we extend this effect from RoPE-NoPE hybrid modes to other hybrid models, including GLA-NoPE and GDN-NoPE hybrids. We will verify such an extension in Section~\ref{sec-la}. Here, we first verify the Tidal Effect in RoPE-NoPE layer-wise hybrid models. Considering the impact of layer order and functional differences at different depths, we conduct the validation experiment on 3:1 RoPE-NoPE, 1 NoPE layer after 3 RoPE layers, 1:1 RoPE-NoPE, 1 NoPE after 1 RoPE layer, 1:3 NoPE-RoPE, 3 RoPE layers after 1 NoPE layer, and 1:1 NoPE-RoPE, 1 RoPE after 1 NoPE layer. We compare the hit-rate curve, retrieval score distribution, and attention entropy curve in both 1:1 RoPE-NoPE (NR) and 1:1 NoPE-RoPE (RN), and present the entropy-hit-rate diagram for 4 hybrids in 376M and 776M model sizes in Figure~\ref{fig-scatter-fa_lh}. The relative rankings of hit rate, retrieval score, and attention entropy, and the distribution pattern between RoPE and NoPE attention are consistent with those in head-wise hybrid models. In addition, we find that attentions in upper layers tend to have higher hit rates and larger entropy for both NoPE and RoPE attentions.

To sum up, we clarify how RoPE and NoPE attention collaborate, extending Takeaway 1 to a more fine-grained level. Although NoPE attention enables training-free effective length extrapolation and a higher attention hit rate, its coarse global localization still requires collaboration with position-biased attention to maintain performance in long-context tasks as the distribution shifts with the hybrid ratio.

\begin{tcolorbox}[colback=blue!5!white,colframe=blue!75!black,title=\textbf{Takeaway 1: From Hybrid Attention to Hybrid Position (Complete Version)}]
Hybrid models improve long-context performance and length extrapolation through a hybrid position of NoPE and other position biases. Particularly, 
\begin{itemize}
    \item On one hand, training-free effective length extrapolation can be achieved even for LLMs pretrained with full attention by limiting the attention scope of position bias.
    \item On the other hand, although NoPE attention consistently aggregates global information, downstream performance still relies on collaboration with other modules.
\end{itemize}
\end{tcolorbox}

\section{Discussion on SWA Hybrid}~\label{sec-swa}

For SWA hybrids, we focus on why the seesaw effect appears during context extension, and why a model that extrapolates well from short-context pretraining can become less competitive after long-context continual pretraining. We first use perplexity curves and parameter change ratios to analyze the source of the problem. Then, we improve context-extension performance by enlarging the window size in long-context continual pretraining and strengthening long-context dependency learning with LongCE~\citep{fang2024wrong}.

\subsection{Short-Context Learning Trap of SWA Hybrids}

\begin{figure}[!b]
  \noindent
  \centering
  \begin{subfigure}[b]{0.48\linewidth}
    \centering
    \includegraphics[width=\linewidth]{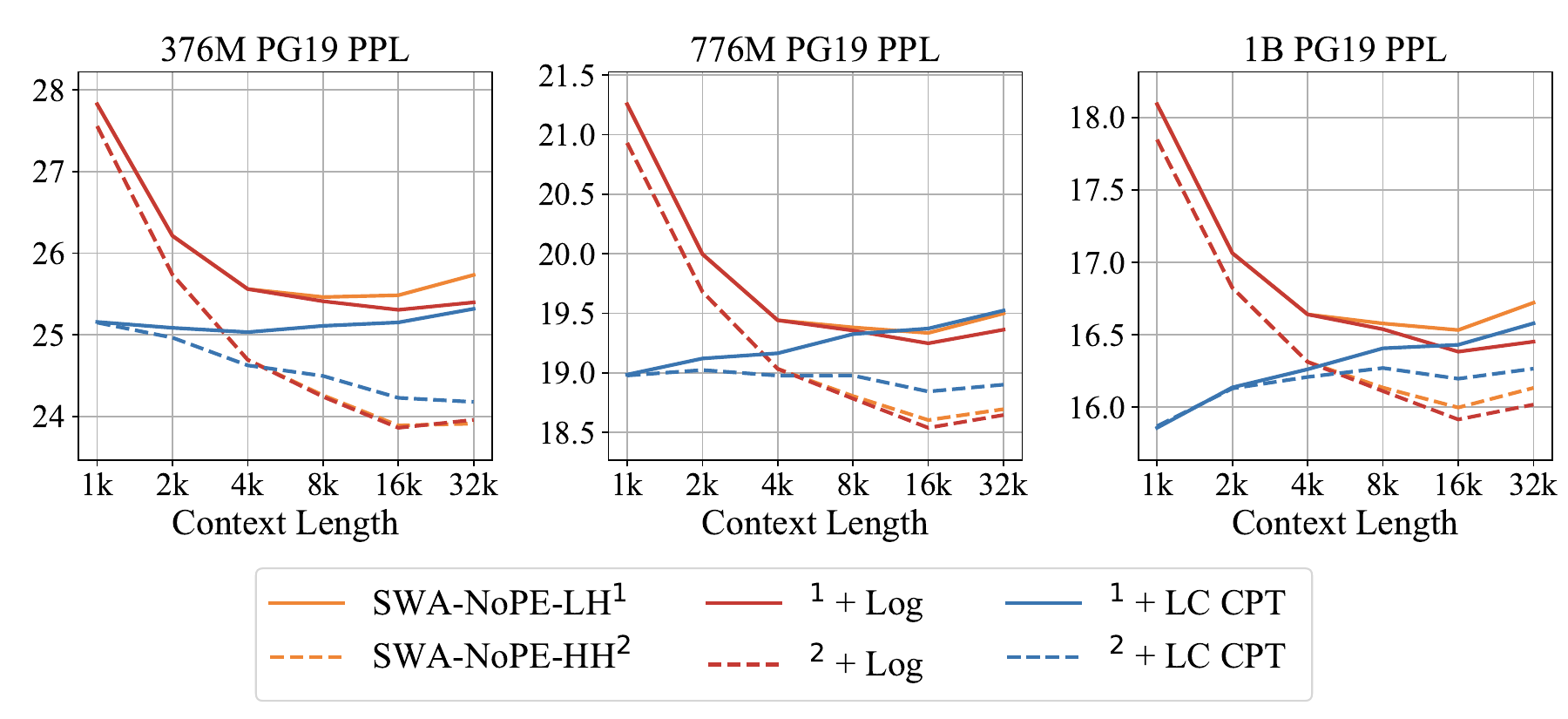}
    \caption{Perplexity of SWA Hybrids.}
  \end{subfigure}
  \hfill
  \begin{subfigure}[b]{0.48\linewidth}
    \centering
    \includegraphics[width=\linewidth]{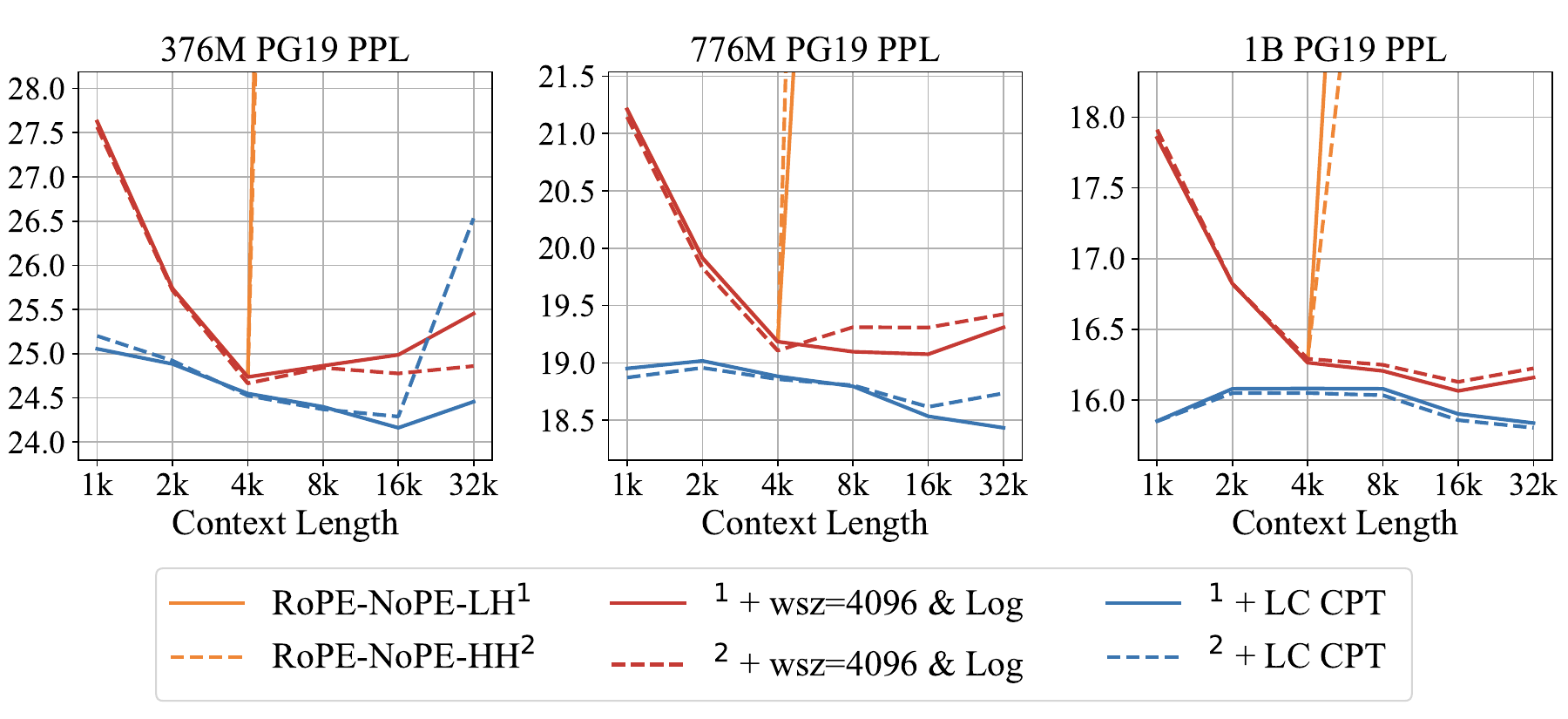}
    \caption{Perplexity on RoPE-NoPE hybrids.}
  \end{subfigure}
  \vskip0.3\baselineskip
  \begin{subfigure}[b]{0.48\linewidth}
    \centering
    \includegraphics[width=\linewidth]{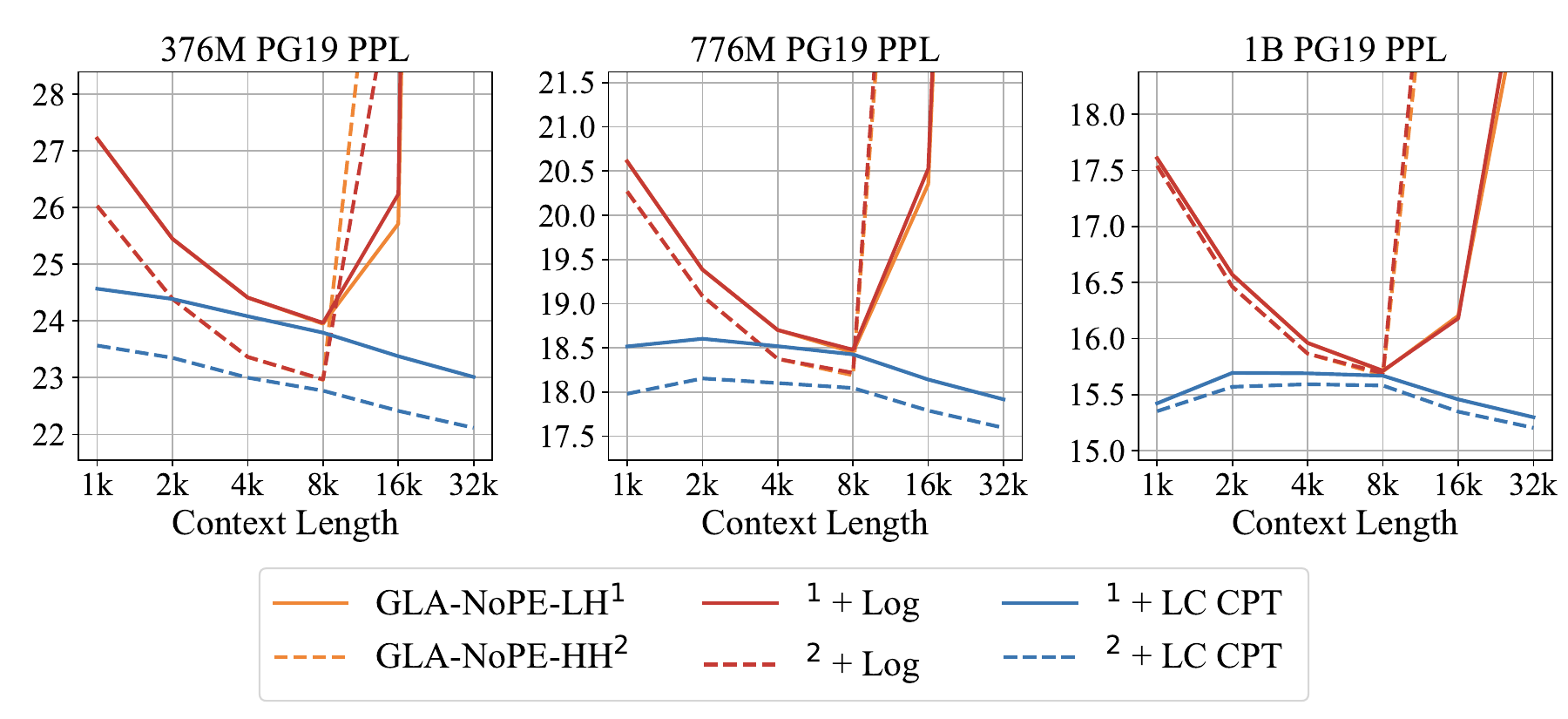}
    \caption{Perplexity of GLA Hybrids.}
  \end{subfigure}
  \hfill
  \begin{subfigure}[b]{0.48\linewidth}
    \centering
    \includegraphics[width=\linewidth]{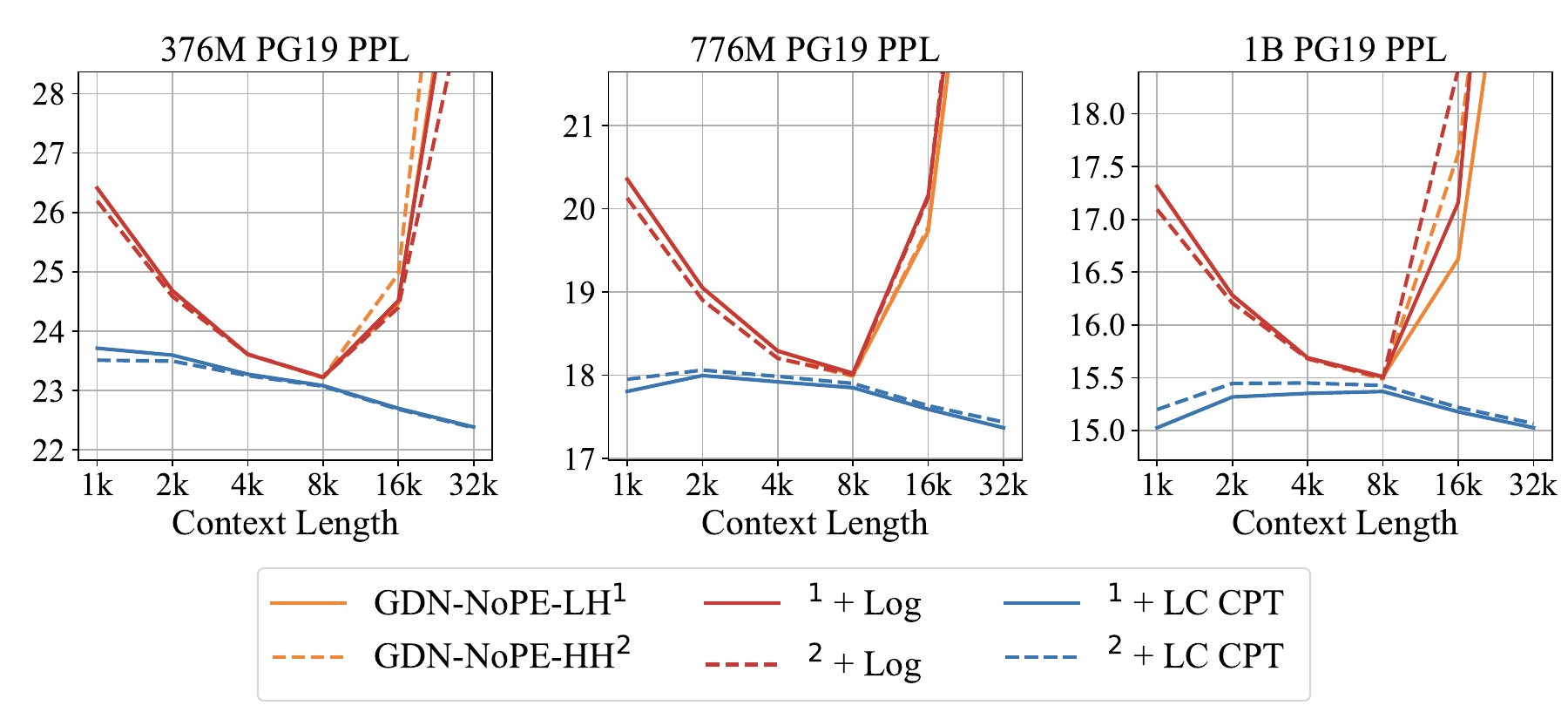}
    \caption{Perplexity of GDN Hybrids.}
  \end{subfigure}
  \caption{Perplexity curves of different hybrid models under direct extrapolation, log-scale NoPE extrapolation (Log), and long-context continual pretraining (LC-CPT). The perplexity curves of SWA-NoPE-LH show a remarkable increasing trend as the context length grows, different from other hybrid models.}
  \label{fig-seesaw-ppl}
\end{figure}

We start the discussion by observing the perplexity curves, and this reveals a potential reason behind the reversal in downstream task ranking. Figure~\ref{fig-seesaw-ppl} compares the perplexity curves of SWA-NoPE, RoPE-NoPE, GLA-NoPE, and GDN-NoPE hybrids before and after long-context continual pretraining. For most models, the perplexity curve before long-context training rises clearly beyond the pretraining context length. After long-context continual pretraining, the curve becomes much flatter or steadily decreases, indicating that the model can support longer contexts and use context to predict the next token better.

However, SWA-NoPE hybrids show the opposite behavior. Because they already have strong extrapolation ability, their perplexity curves keep decreasing within 32k before long-context continual pretraining. After long-context continual pretraining, especially for layer-wise SWA-NoPE hybrids, the perplexity curve instead shows an upward trend: \textbf{the loss on long-context positions decreases only slightly}, while \textbf{the loss on short-context positions drops abnormally strongly}, and this phenomenon becomes more obvious as the model size increases. This means that long-context dependencies do not help SWA-NoPE hybrids make better predictions. Long-context continual pretraining of SWA-NoPE hybrids focuses too much on short contexts, and as a result, LA-NoPE hybrids surpass SWA-NoPE hybrids after long-context continual pretraining. We refer to this phenomenon as the \textbf{Short-Context Learning Trap in Context Extension of SWA Hybrid}. This explains why the seesaw effect emerges during context extension of SWA hybrids and refines Takeaway 2.

\begin{tcolorbox}[float,colback=blue!5!white,colframe=blue!75!black,title=\textbf{Takeaway 2: Seesaw Effect in Context Extension of SWA Hybrid (Complete Version)}.]
For the SWA-NoPE hybrid model, the advantage in long-context performance over the LA-NoPE hybrid is inverted after long-context continual pretraining, which is more remarkable in SWA-NoPE layer-wise hybrids, due to the \textbf{Short-Context Learning Trap in Context Extension of SWA Hybrid}.
\end{tcolorbox}

\begin{figure}[!tb]
  \noindent
  \centering
  \begin{subfigure}[b]{0.48\linewidth}
    \centering
    \includegraphics[width=\linewidth]{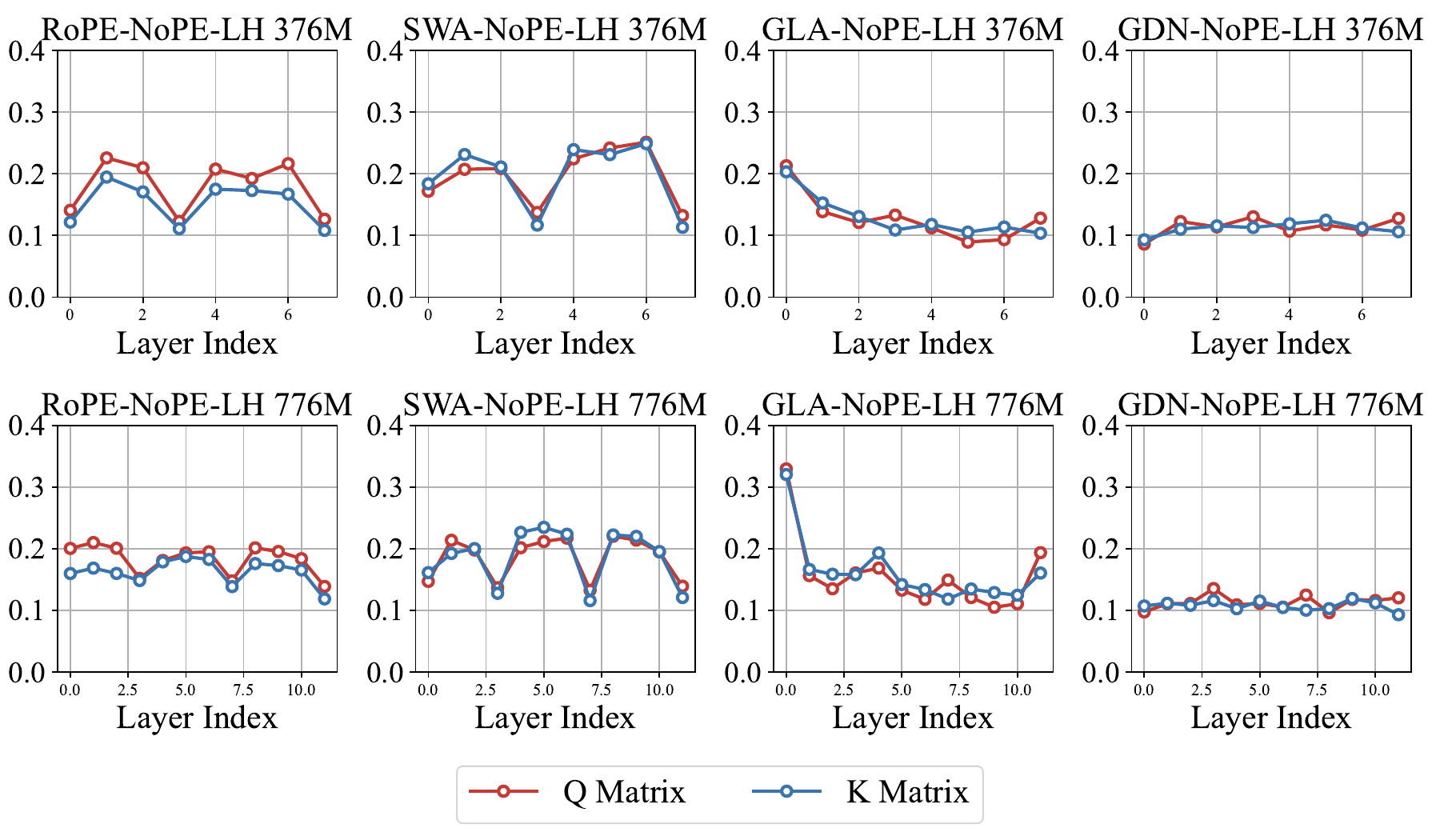}
    \caption{Comparison of layer-wise hybrid models.}
  \end{subfigure}
  \hfill
  \begin{subfigure}[b]{0.48\linewidth}
    \centering
    \includegraphics[width=\linewidth]{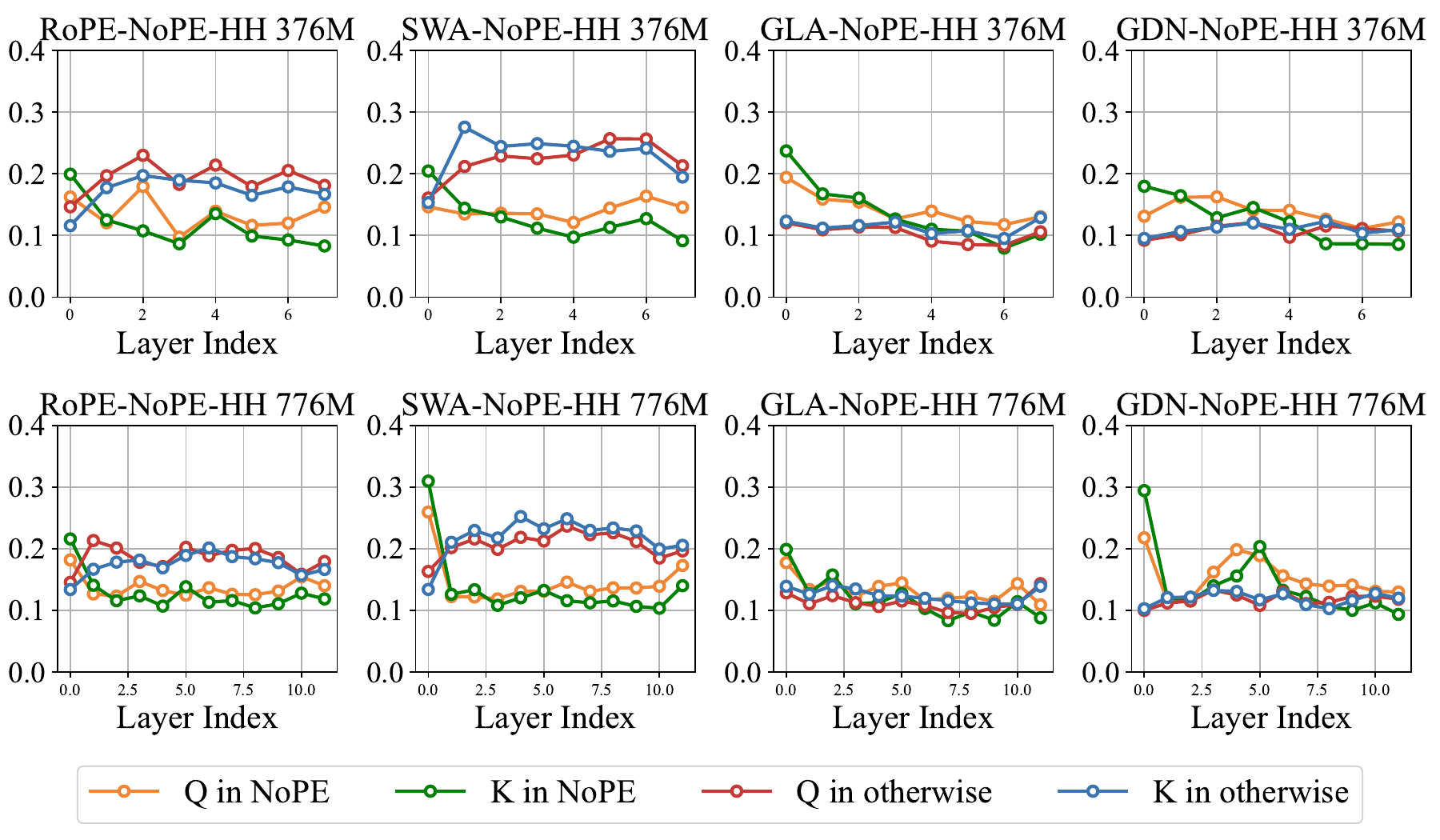}
    \caption{Comparison of head-wise hybrid models.}
  \end{subfigure}
  \caption{Comparison of the change rates of the $\bm{W}_Q$ and $\bm{W}_K$ matrices in attention modules across different hybrid models before and after long-context training. In layer-wise hybrids, 1 NoPE layer is inserted every 3 RoPE/SWA/GLA/GDN layers. In head-wise hybrids, each layer contains the $\bm{W}_Q$ and $\bm{W}_K$ matrices of NoPE as well as those of the other RoPE/SWA/GLA/GDN modules. Remarkably, in SWA-NoPE hybrids, the SWA modules receive anomalous emphasis during long-context training. However, their limited window size cannot support modeling longer contexts.}
  \label{fig-seesaw-param}
\end{figure}

To further verify the existence of the short-context learning trap, we plot the parameter change ratios in Figure~\ref{fig-seesaw-param}, comparing SWA-NoPE, RoPE-NoPE, GLA-NoPE, and GDN-NoPE hybrids. The x-axis represents different layers, and the y-axis represents the parameter change ratios of the $\bm{W}_Q$ and $\bm{W}_K$ matrices. Taking $\bm{W}_Q$ as an example, let $\bm{W}_Q$ and $\bm{W}_Q'$ denote its values before and after long-context continual pretraining. We define the change ratio as $\|\bm{W}_Q'-\bm{W}_Q\|_2/\|\bm{W}_Q\|_2$. We first compute this ratio for each head within the same layer, and then average over heads. For head-wise hybrids, we separately aggregate NoPE heads and non-NoPE heads in each layer. For layer-wise hybrids, we do not make an additional distinction within each layer, and the non-NoPE layers and NoPE layers are interleaved at a 3:1 ratio.

We find that for GLA-NoPE and GDN-NoPE hybrids, the parameter change ratios of GLA/GDN and NoPE attention are close. In contrast, for SWA and RoPE attention, especially SWA, the parameter change ratio is significantly larger than that of NoPE attention in both layer-wise and head-wise hybrids. This suggests that \textbf{during long-context continual pretraining, SWA and RoPE attention receive much more optimization} in SWA-NoPE and RoPE-NoPE hybrids. For SWA, however, the window size remains fixed, so the range of context dependencies that it can model is still limited. Therefore, its abnormal focus on this bounded local branch is what drives SWA-NoPE hybrids into the short-context learning trap during context extension.

\subsection{Short-Window Weariness and Long-Window Laziness}

\begin{figure}[!tb]
  \noindent
  \centering
  \begin{minipage}[c]{0.48\linewidth}
  \centering
  \includegraphics[width=\linewidth]{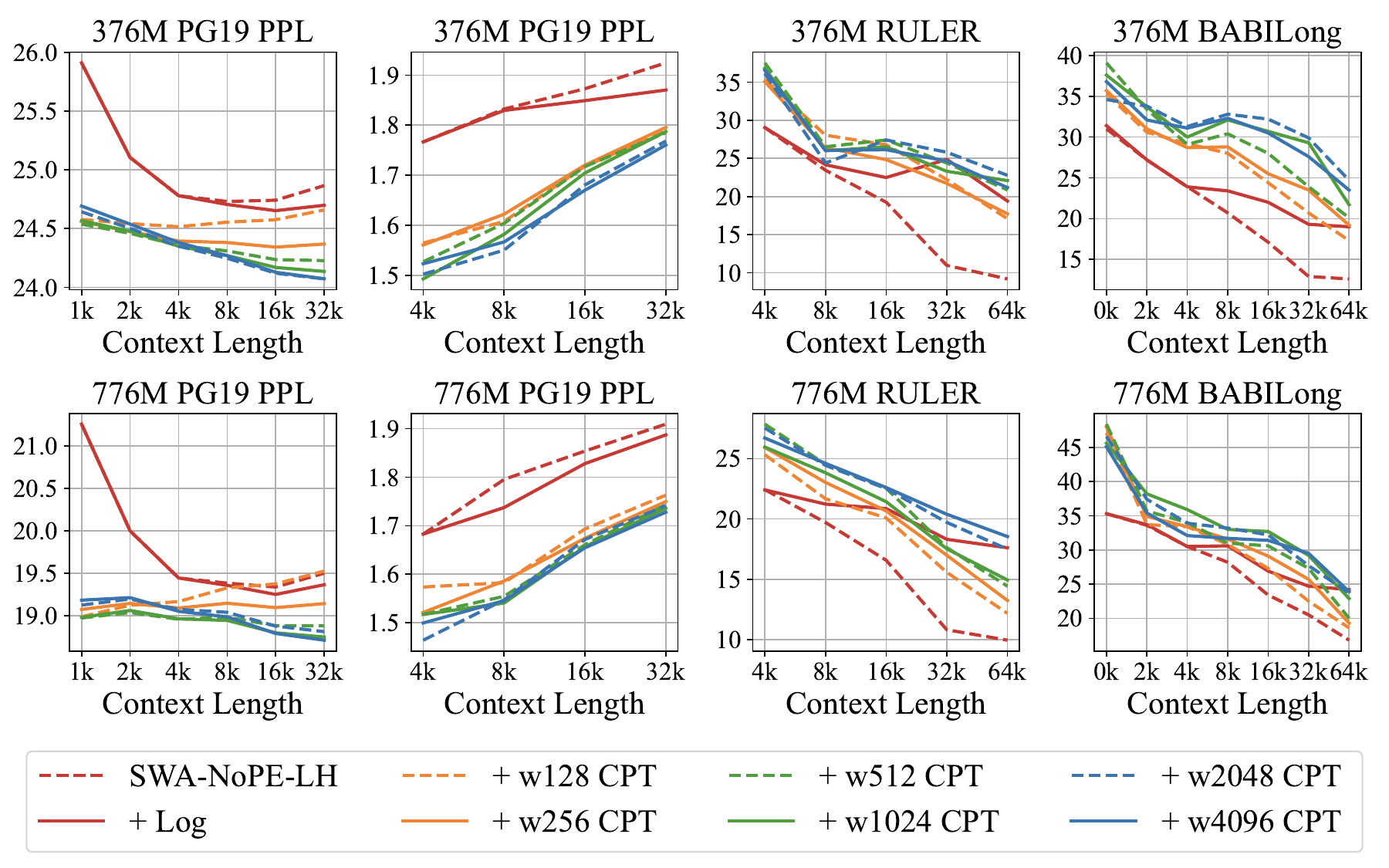}
  \vspace{-18pt}
  \caption{Comparison of long-context perplexity and performance of SWA-NoPE-LH with different window sizes (wsz) in long-context continual pretraining (CPT).}
  \label{fig-seesaw-sww1}
  \end{minipage}
\hfill
  \begin{minipage}[c]{0.48\linewidth}
  \centering
  \includegraphics[width=\linewidth]{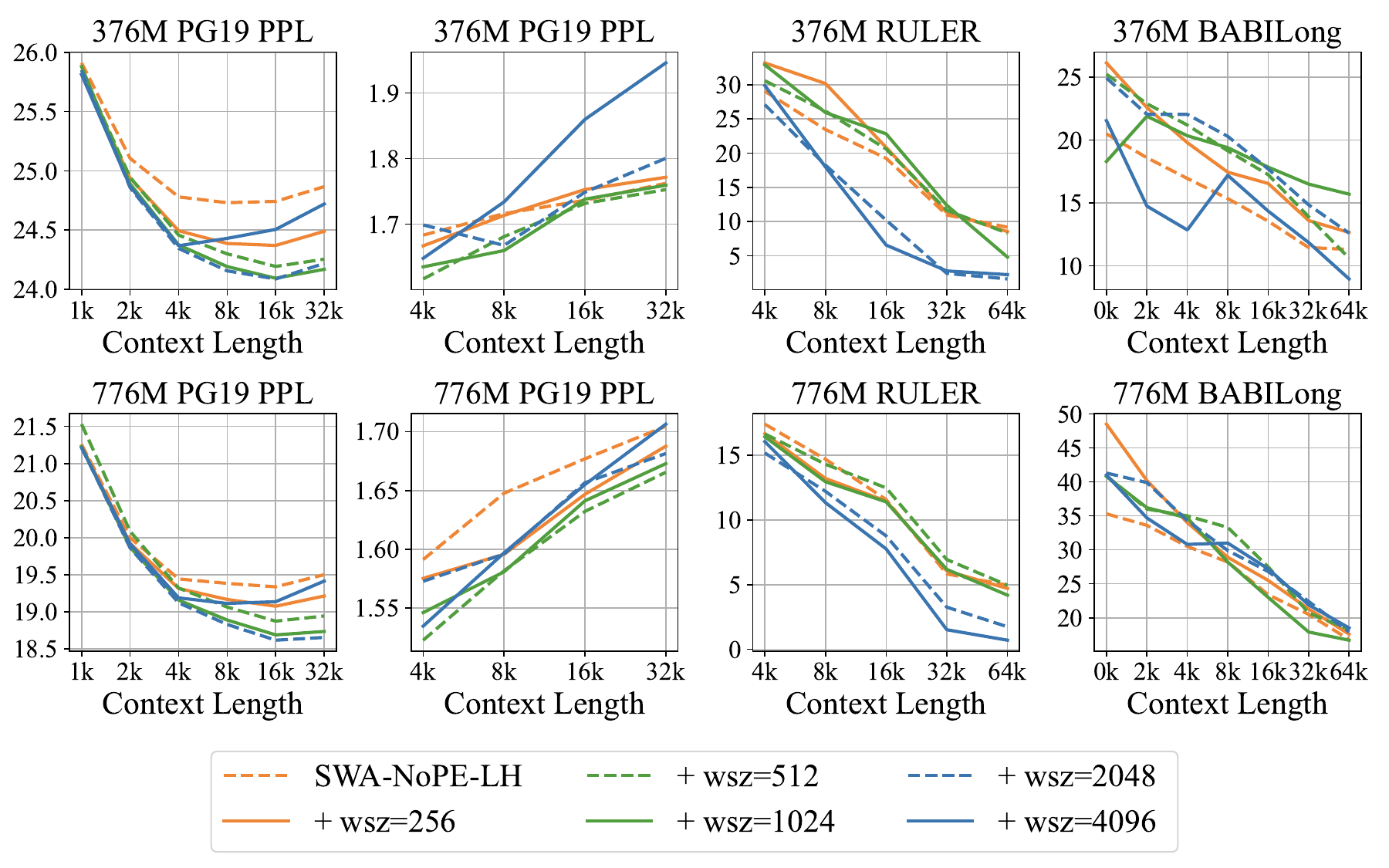}
  \vspace{-18pt}
  \caption{Comparison of long-context perplexity and performance of SWA-NoPE-LH with different window sizes (wsz) in short-context pretraining.}
  \label{fig-seesaw-lwl1}
  \end{minipage}
\end{figure}

\begin{figure}[!tb]
  \noindent
  \centering
  \includegraphics[width=\linewidth]{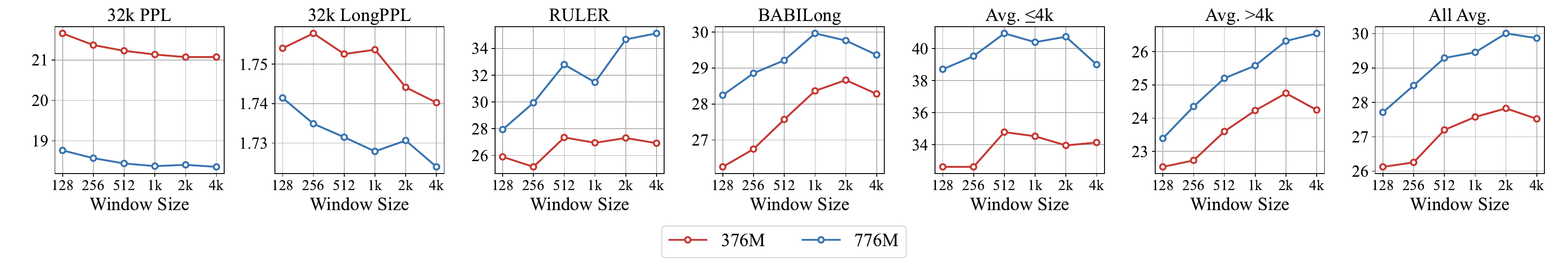}
  \vspace{-18pt}
  \caption{Tendency of long-context perplexity and performance as the window size of SWA-NoPE-LH increases in long-context pretraining. Larger window sizes show better context extension, implying short-window weariness.}
  \label{fig-seesaw-sww2}
\end{figure}

Motivated by Figure~\ref{fig-seesaw-param}, we enlarge the SWA window size during long-context continual pretraining of layer-wise SWA-NoPE hybrids. The idea is to use the abnormal focus on the SWA branch to capture longer context dependencies. Specifically, we increase the window size from 128 to 256, 512, 1024, 2048, and 4096. Following the scaling law of RoPE-based extrapolation~\cite{liuscaling}, we also compute the rotary base corresponding to each enlarged window size when the initial window size is 128, and the original rotary base is 10000. Figure~\ref{fig-seesaw-sww1} and Figure~\ref{fig-seesaw-sww2} compare the PG19 perplexity, RULER, and BABILong performance of SWA hybrids after enlarging the window. The results show that increasing the window size indeed improves downstream long-context performance, and a window size of 2048 gives the best results at both 376M and 776M scales.

In addition, enlarging the window only exploits the short-context learning trap rather than emphasizing long-context dependencies directly. Therefore, we also use LongCE~\citep{fang2024wrong} as the training objective for layer-wise SWA-NoPE hybrids in long-context continual pretraining. As shown in Figure~\ref{fig-seesaw-longce}, both enlarging the window to 2048 and using LongCE improve the long-context performance of SWA-NoPE layer-wise hybrids, and the two methods can be combined. With both methods, SWA-NoPE-LH becomes competitive with GDN-NoPE-LH across different model sizes. Since the short-context trap in SWA hybrid long-context continual pretraining may not be limited to SWA-NoPE layer-wise hybrids, we further apply the same two strategies to SWA-RoPE layer-wise hybrids. Enlarging the window and using LongCE also improves the long-context training performance of SWA-RoPE-LH at different scales, making it close to GDN-RoPE-LH.

\begin{figure}[!tb]
  \noindent
  \centering
  \includegraphics[width=\linewidth]{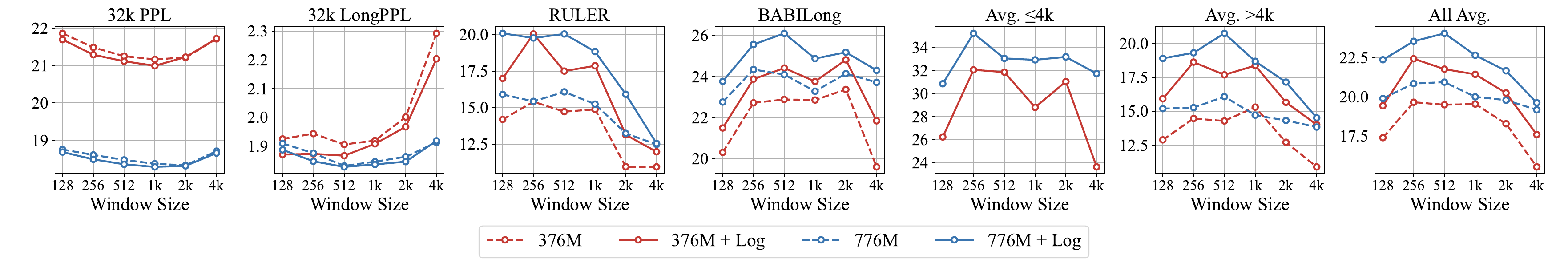}
  \vspace{-18pt}
  \caption{Tendency of long-context perplexity and performance as the window size of SWA-NoPE-LH increases in short-context pretraining. Shorter window sizes show better length extrapolation, implying long-window laziness.}
  \label{fig-seesaw-lwl2}
\end{figure}

\begin{figure}[!tb]
  \noindent
  \centering
  \begin{subfigure}[b]{0.48\linewidth}
    \centering
    \includegraphics[width=\linewidth]{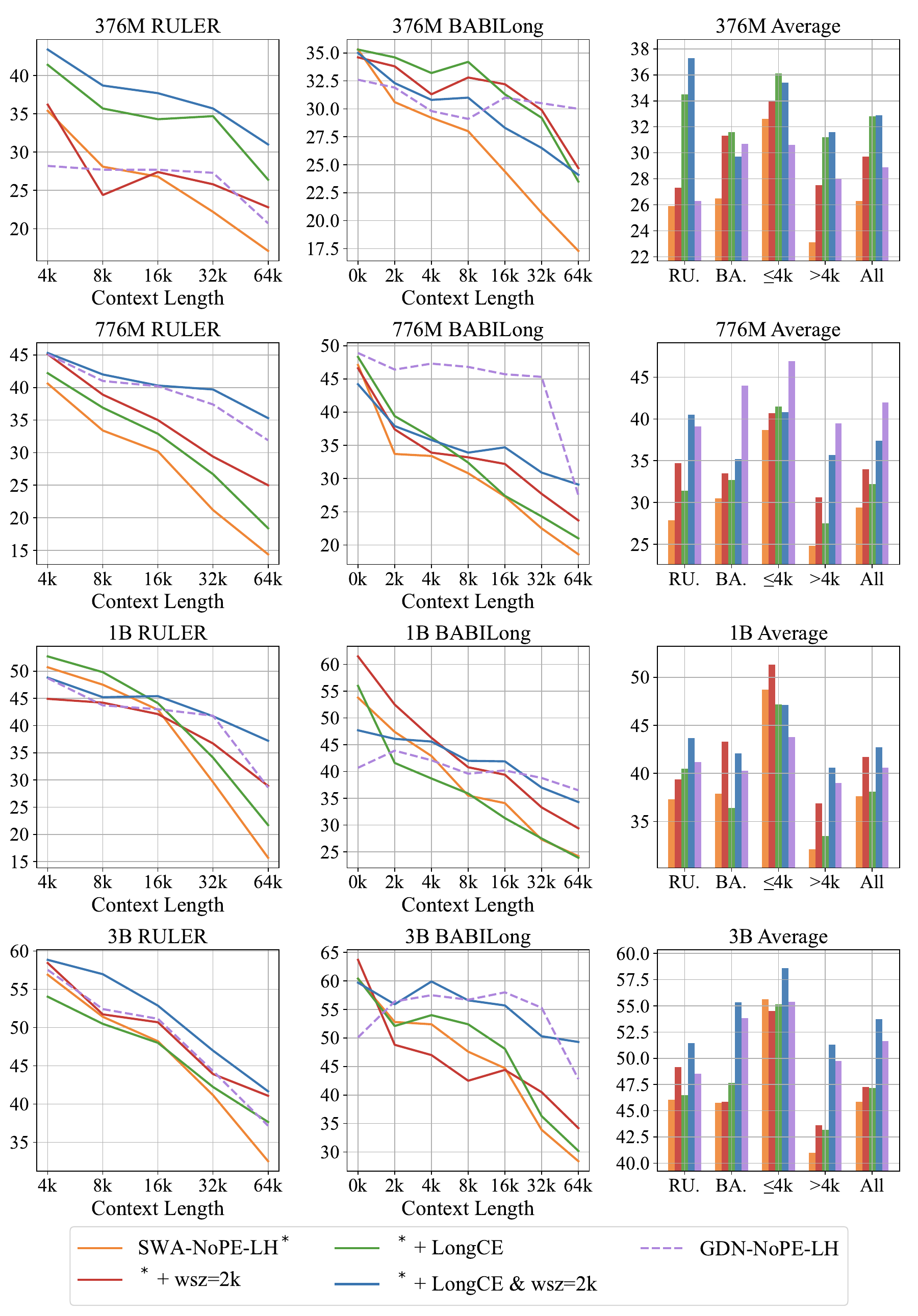}
    \caption{Results of SWA-NoPE-LH.}
  \end{subfigure}
  \hfill
  \begin{subfigure}[b]{0.48\linewidth}
    \centering
    \includegraphics[width=\linewidth]{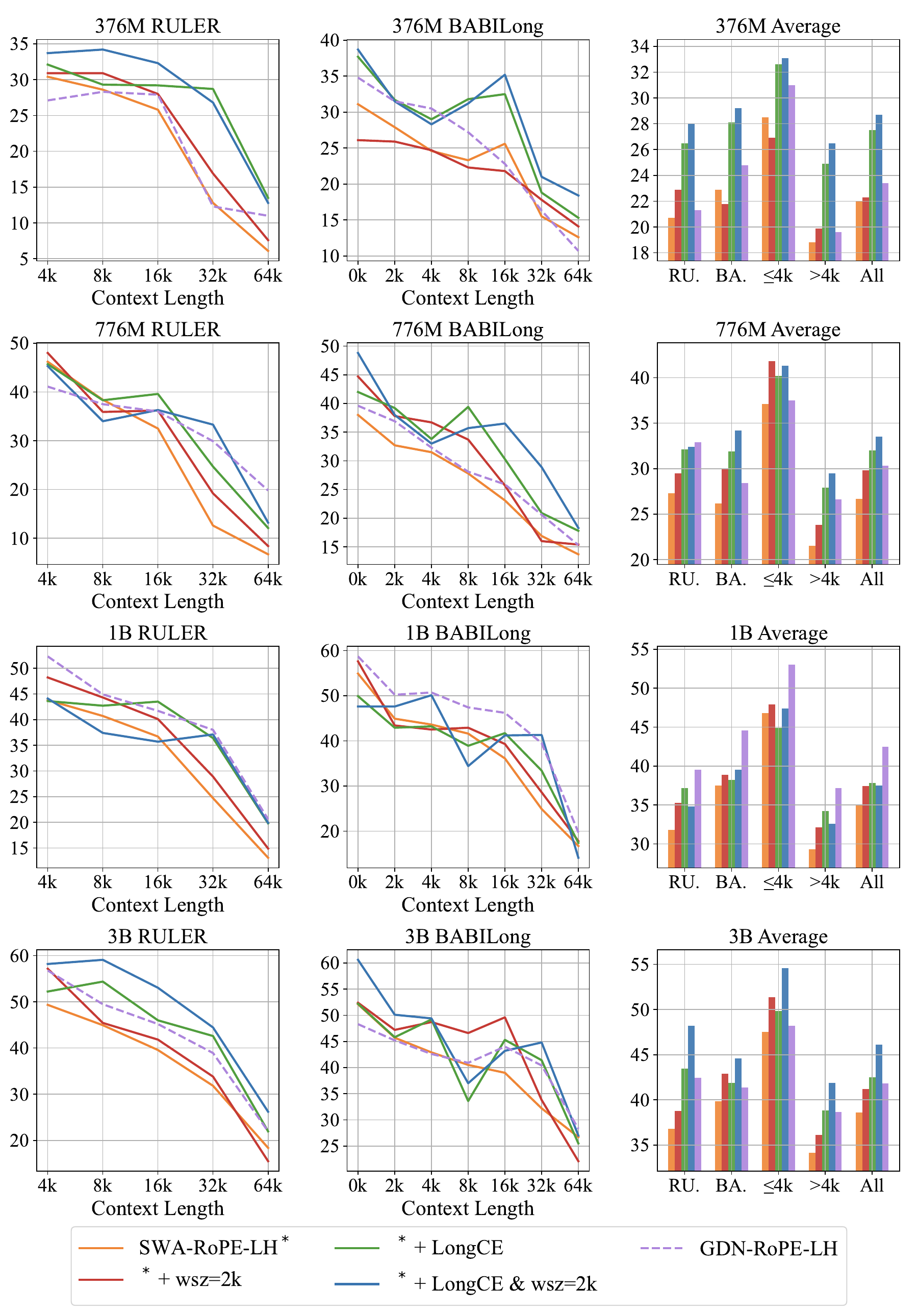}
    \caption{Results of SWA-RoPE-LH.}
  \end{subfigure}
  \caption{Comparison of long-context performance of SWA-NoPE-LH after long-context continual pretraining enhanced by LongCE and extended window size. Both clearly weaken the seesaw effect and also contribute to SWA-RoPE-LH.}
  \label{fig-seesaw-longce}
\end{figure}

These results show that keeping a small SWA window during long-context continual pretraining harms context extension. \textbf{SWA hybrids suffer from the short-context learning trap and require enlarged windows as well as training strategies that emphasize long-context learning}. We call this phenomenon \textbf{Short-Window Weariness of SWA Hybrid}. At the same time, some previous works have reported that for SWA hybrids, shorter windows can be more advantageous than longer windows during the initial training stage, because they better activate the potential of NoPE full-attention layers~\citep{hu2026every,qiao2026rethinking}; this phenomenon is sometimes called \textbf{Long-Window Laziness}. We argue that the two observations are not contradictory. 
In fact, we can also verify this tendency as shown in Figure~\ref{fig-seesaw-lwl1} and Figure~\ref{fig-seesaw-lwl2}. 
The coexistence of short-window weariness and long-window laziness indicates that SWA hybrids may require fine-grained adjustment of positional bias across short-context and long-context training stages, just as RoPE full-attention models need to tune the rotary base and scaling factor in context extension~\citep{chen2023extending,pengyarn,liuscaling,xiong2024effective}. We summarize the coexistence of short-window weariness and long-window laziness as \textbf{Takeaway 5} in this paper.

\begin{tcolorbox}[colback=blue!5!white,colframe=blue!75!black,title=\paragraph{Takeaway 5: Short-Window Weariness and Long-Window Laziness of SWA Hybrid}\ ]
\begin{itemize}[leftmargin=*]
\item In short-context pretraining, SWA hybrids with shorter window size show better length extrapolation.
\item In long-context pretraining, SWA hybrids with larger window size show better context extension.
\end{itemize}
\end{tcolorbox}


\section{Discussion on LA Hybrid}~\label{sec-la}

For LA hybrids, we first ask how to present the collaboration of LA variants with NoPE attention, and whether this collaboration differs from that between RoPE attention and NoPE attention. Moreover, since GLA/GDN-NoPE hybrids still underperform SWA hybrids in length extrapolation, revealing the No-Free-Lunch Effect in Takeaway 3, we therefore need to consider how to improve their extrapolation performance.

\subsection{Extended Hybrid Position Analysis}

The biggest difference between linear attention and softmax attention is that linear attention does not have a standard attention distribution, and its attention scores are not necessarily non-negative. Therefore, we cannot compute the attention entropy directly. Of course, one could apply softmax to the attention scores of linear attention and obtain a distribution, but this is inconsistent with the original computation of linear attention. Before analyzing linear attention, we therefore need a method that extends concepts such as attention distribution and attention entropy to arbitrary attention mechanisms, compatible with softmax attention and without extra operations on other attention algorithms.

\begin{figure}[!tb]
  \noindent
  \centering
  \begin{subfigure}[b]{0.403\linewidth}
    \centering
    \includegraphics[width=\linewidth]{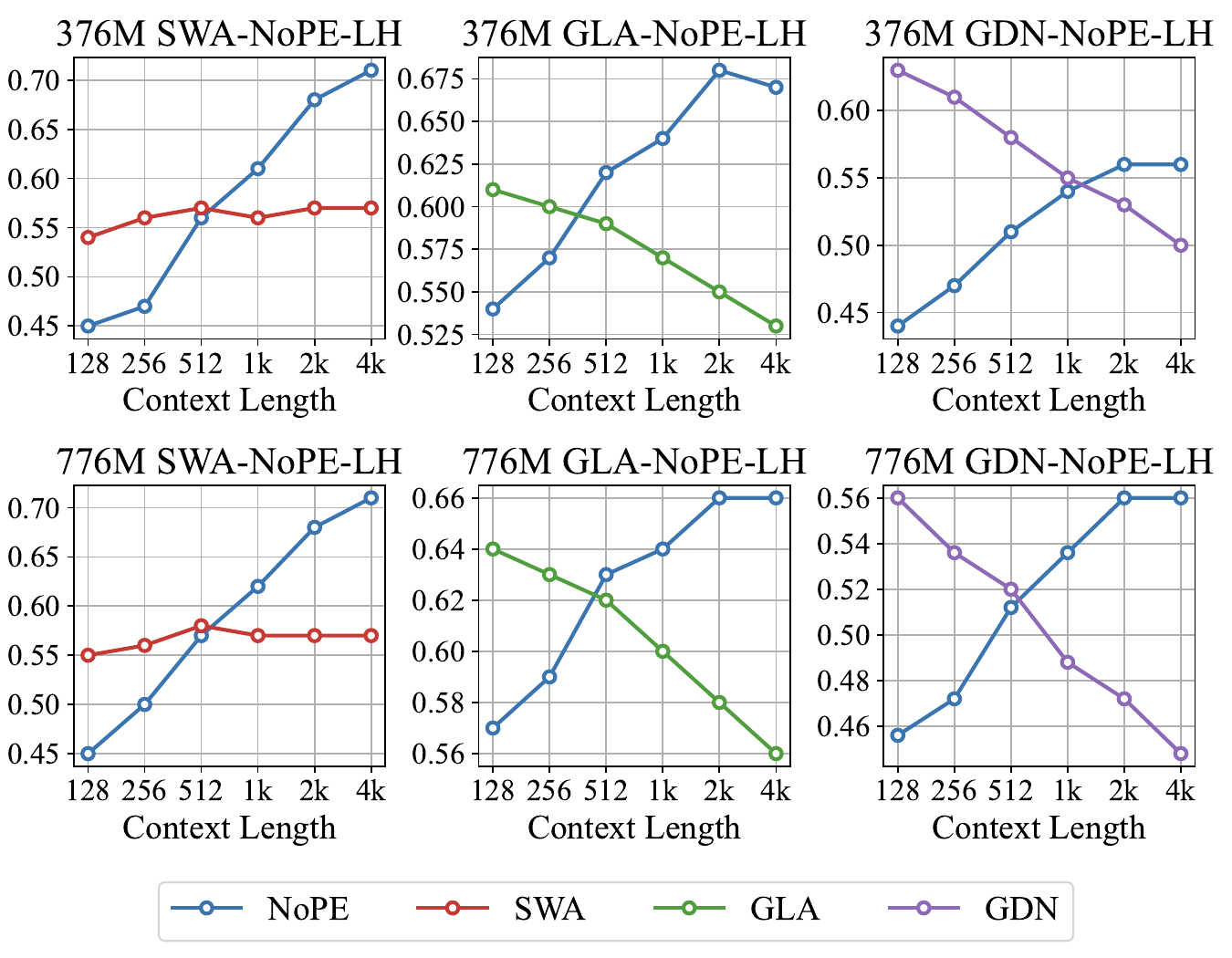}
    \caption{Entropy comparison within NoPE-LH.}
  \end{subfigure}
  \hfill
  \begin{subfigure}[b]{0.587\linewidth}
    \centering
    \includegraphics[width=\linewidth]{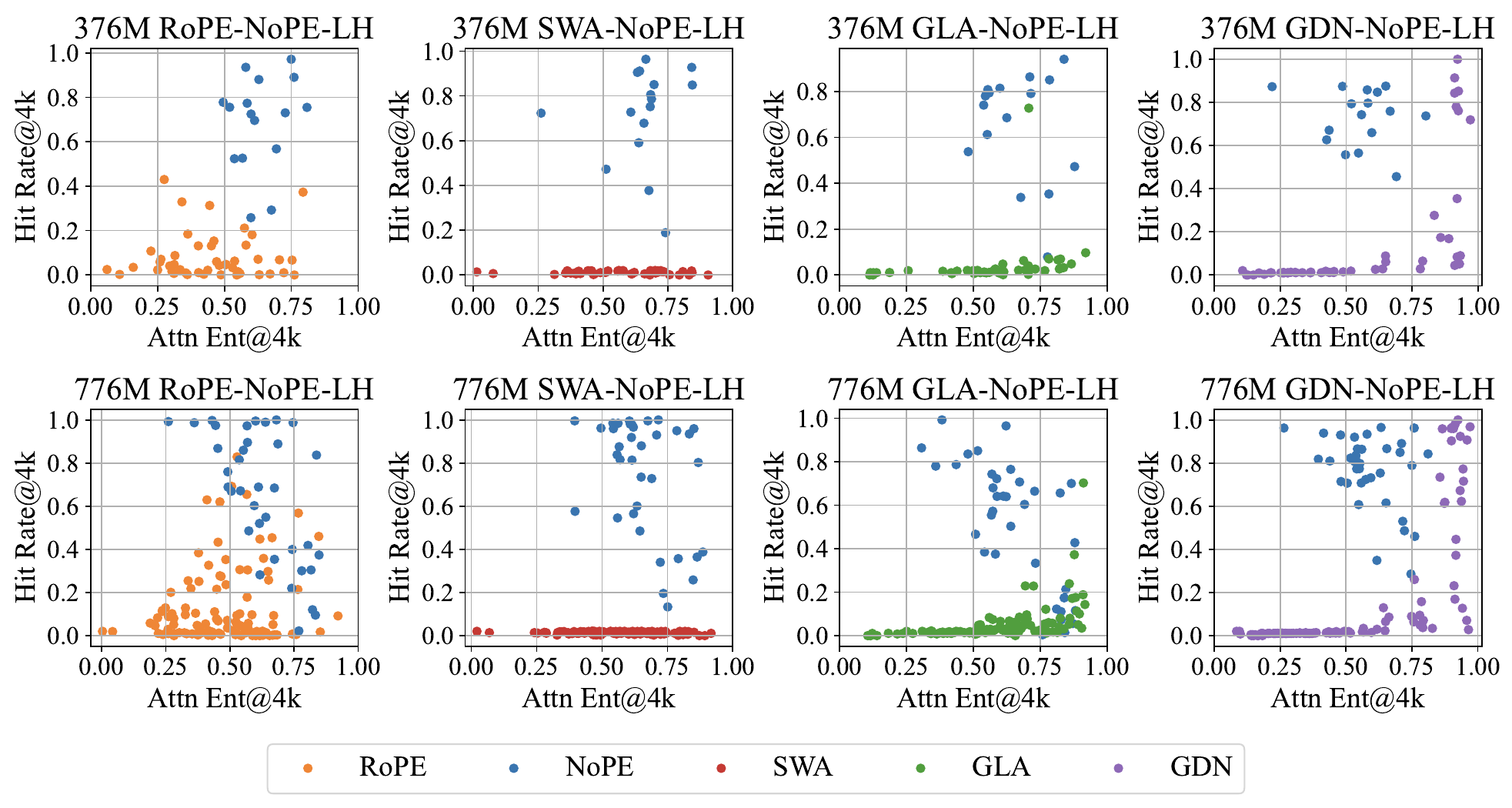}
    \caption{Entropy-hit-rate scatter diagram comparison among NoPE-LH.}
  \end{subfigure}
  \vskip0.6\baselineskip
  \begin{subfigure}[b]{0.403\linewidth}
    \centering
    \includegraphics[width=\linewidth]{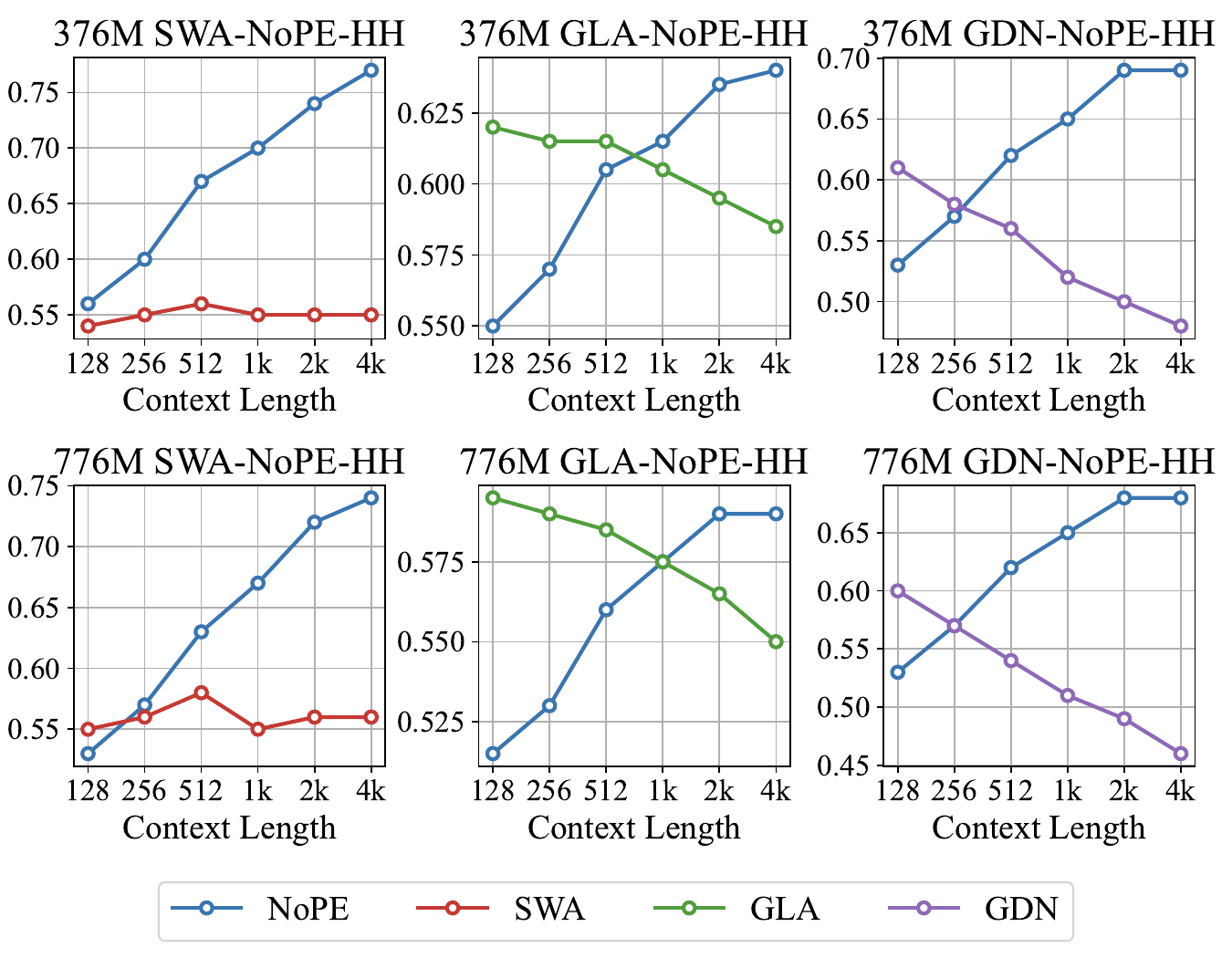}
    \caption{Entropy comparison within NoPE-HH.}
  \end{subfigure}
  \hfill
  \begin{subfigure}[b]{0.587\linewidth}
    \centering
    \includegraphics[width=\linewidth]{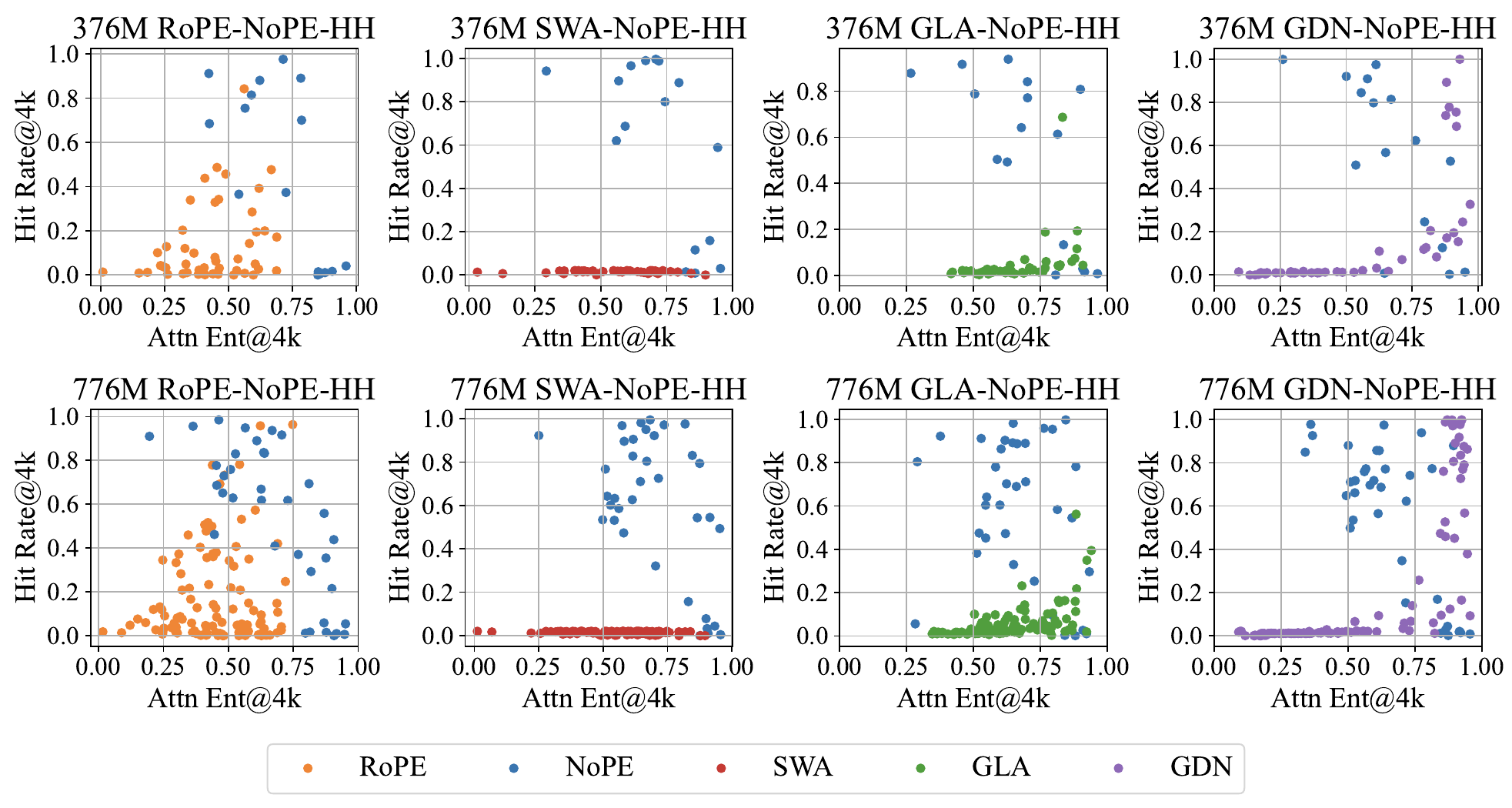}
    \caption{Entropy-hit-rate scatter diagram comparison among NoPE-HH.}
  \end{subfigure}
  \caption{Attention analysis results among 3:1 RoPE-NoPE, SWA-NoPE, GLA-NoPE, and GDN-NoPE hybrid models, with extended attention entropy and hit rate. In the entropy-hit-rate diagram, NoPE attention still shows a stronger tendency to aggregate global information coarsely, with a higher attention hit rate and larger attention entropy. In contrast, SWA, GLA, and GDN show a similar distribution pattern to RoPE attention.}
  \label{fig-scatter-all}
\end{figure}

\begin{figure}[!tb]
  \noindent
  \centering
  \begin{subfigure}[b]{0.49\linewidth}
    \centering
    \includegraphics[width=\linewidth]{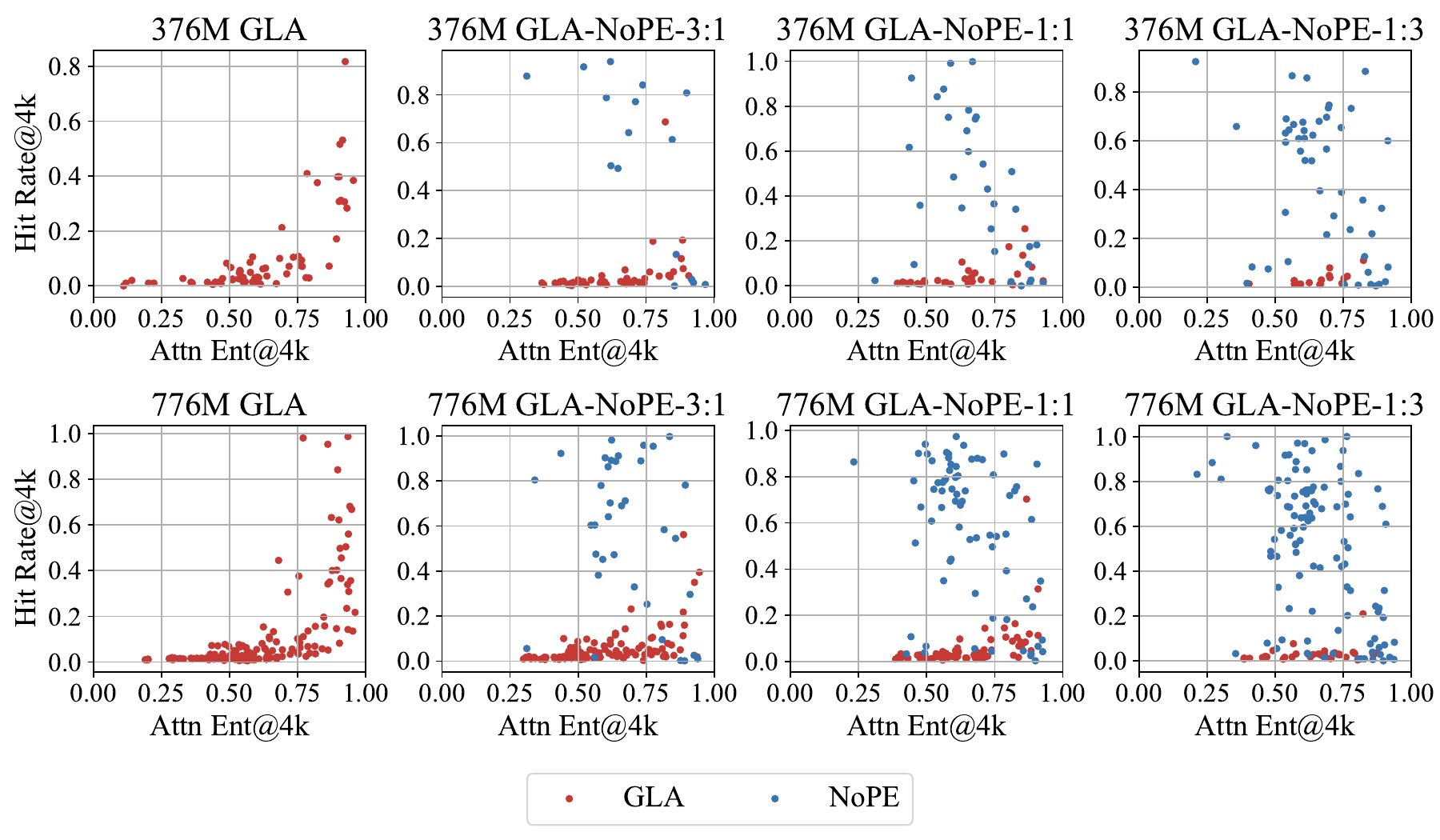}
    \caption{Comparison among GLA-NoPE-HH}
  \end{subfigure}
  \hfill
  \begin{subfigure}[b]{0.49\linewidth}
    \centering
    \includegraphics[width=\linewidth]{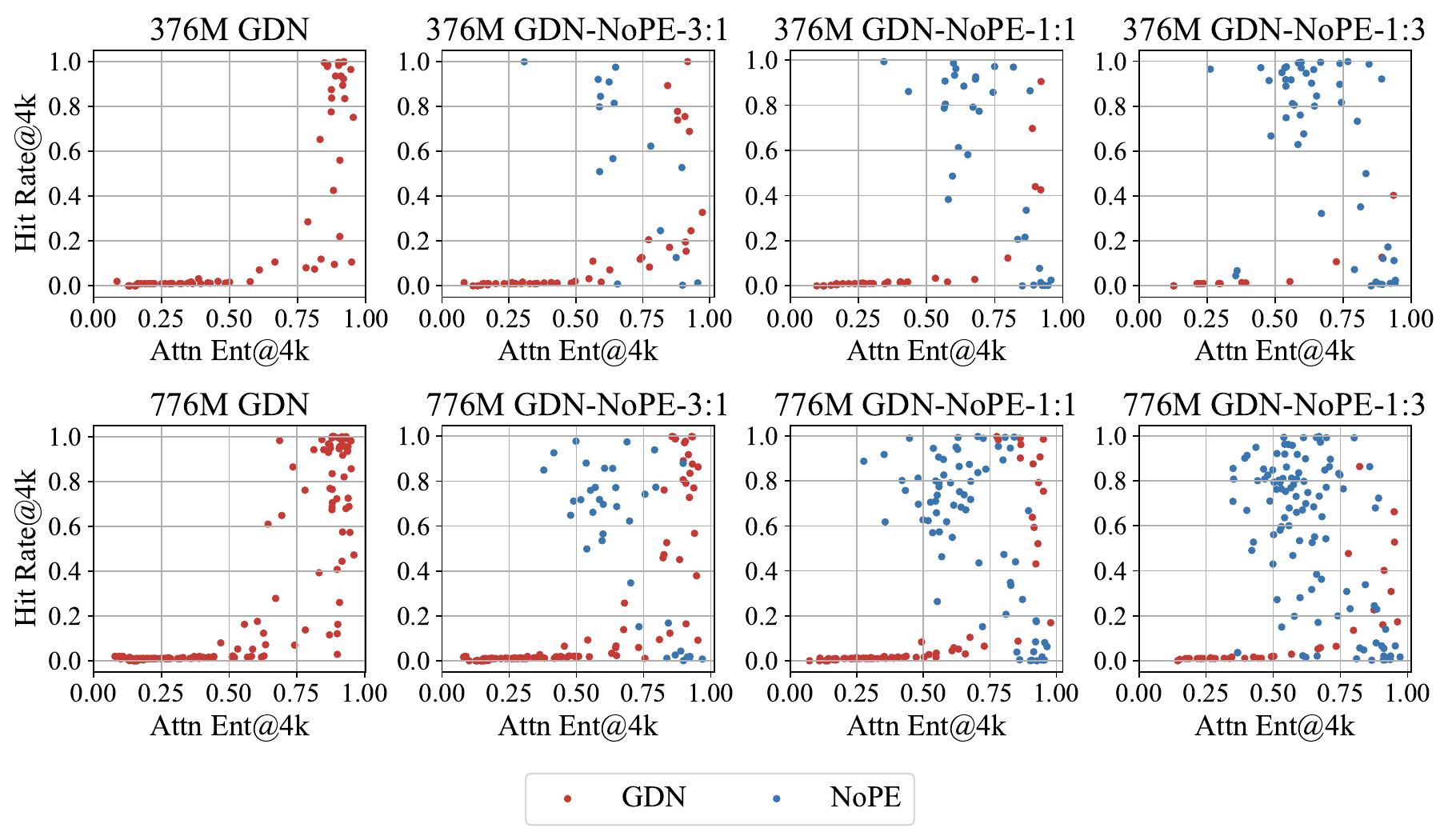}
    \caption{Comparison among GDN-NoPE-HH}
  \end{subfigure}
  \caption{Attention analysis results among GLA-NoPE-HH and GDN-NoPE-HH with different hybrid ratios. The tidal effect of hybrid position is still applicable in GLA-NoPE and GDN-NoPE head-wise hybrid models.}
  \label{fig-scatter-la_hh}
\end{figure}

We note that the original attention distribution $\alpha_{t,s}$ is not only the softmax of the scaled inner product between $\bm{q}_t$ and $\bm{k}_s$, but also the combination weight of $\bm{v}_s$ when forming $\bm{o}_t$. From this perspective, $\alpha_{t,s}$ depends on the internal attention computation only in the forward pass. In the backward view, it corresponds to the derivative of $\bm{o}_t$ with respect to $\bm{v}_s$, more precisely, the norm of the Jacobian matrix as shown in Equation~\ref{equ-dist}.
\begin{equation}
    \alpha_{t,s}=\mathrm{softmax}{\left(\frac{\bm{q}_t\bm{k}^\top_s}{\sqrt{d_k}}\right)}, \quad \bm{o}_t=\sum_{s=0}^{t}{\alpha_{t,s}\bm{v}_s}, \quad \frac{\mathrm{d}\bm{o}_t}{\mathrm{d}\bm{v}_s}=\alpha_{t,s}\bm{I}, \quad \left\|\frac{\mathrm{d}\bm{o}_t}{\mathrm{d}\bm{v}_s}\right\|_2=\alpha_{t,s}.\label{equ-dist}
\end{equation}
Since a matrix norm is non-negative, after normalization it yields a value in $[0,1]$ and can be used as a distribution to compute attention entropy. For softmax attention, because $\alpha_{t,s}$ is already normalized, this extension does not change the original result. For linear attention, this extension does not modify the original attention computation. We call the resulting distribution the extended attention distribution, and call its entropy the extended attention entropy. By default, we also normalize this entropy by dividing by the logarithm of the context length. We use the induced matrix 2-norm (spectral norm) for the Jacobian, compute extended attention entropy and top-k hit rate, and use them to discuss the characteristics of SWA, GLA, and GDN beyond RoPE attention. The results are shown in Figure~\ref{fig-scatter-all} and Figure~\ref{fig-scatter-la_hh}.

Figure~\ref{fig-scatter-all} first compares the characteristics of different attention modules in 3:1 layer-wise SWA-NoPE, RoPE-NoPE, GLA-NoPE, and GDN-NoPE hybrids. Because SWA always keeps a fixed attention range, its attention entropy changes only slightly as the context length increases. Correspondingly, SWA also struggles to hit key information in the previous context, and its points stay close to the x-axis in the entropy-hit-rate diagram. NoPE attention behaves similarly to Figure~\ref{fig-scatter-fa_hh}: its entropy still increases with context length, indicating a more uniform distribution and a coarser aggregation. In the entropy-hit-rate diagram, NoPE attention again occupies the strip region from the upper middle to the lower right. In contrast, \textbf{GLA and GDN behave more like the RoPE attention} in Figure~\ref{fig-scatter-fa_hh}: their entropy decreases as the context length increases, suggesting that they perceive only a limited amount of information and become more concentrated at longer lengths. The difference is that GLA and GDN start in a medium-entropy region compared with RoPE, which may be related to the absence of softmax in GLA and GDN: high-score positions are not explicitly amplified by softmax. Therefore, in the entropy-hit-rate diagram, GLA and GDN still occupy regions close to the lower-left area like RoPE attention, but overall shift toward the right, higher-entropy side.

We further compare head-wise GLA-NoPE and GDN-NoPE hybrids with different hybrid ratios in Figure~\ref{fig-scatter-la_hh}. Similar to Figure~\ref{fig-scatter-fa_hh}, as the NoPE ratio increases, the distribution of GLA or GDN in the entropy-hit-rate diagram gradually shrinks toward the lower-left region, while NoPE attention simultaneously occupies the high-hit-rate region. A difference from RoPE attention is that because the distributions of GLA and GDN are overall compressed toward the right, they also have points in the lower-right region with low hit rate and high entropy. This corresponds to the observation in Figure~\ref{fig-hybrid-ratio_hh} that RoPE-NoPE hybrids are more sensitive to the increase of the NoPE ratio than LA-NoPE hybrids. Therefore, we answer the hypothesis left in Section~\ref{sec-hybrid}, and extend the Tidal Effect of Hybrid Position from RoPE-NoPE hybrids to mixtures of NoPE attention and position-biased attention, including RoPE attention as well as gated linear attention such as GLA and GDN. This naturally leads to the next question: can we also use the extrapolation strategy of RoPE-NoPE hybrids~\citep{yang2026rope,puvvada2025swan} to enhance the length extrapolation of GLA-NoPE and GDN-NoPE hybrids?

\subsection{Length Extrapolation of LA Hybrid}

In RoPE-NoPE hybrids, we achieve length extrapolation beyond traditional NTK interpolation by treating RoPE attention as a sliding-window attention module with a 4k window, while keeping NoPE attention globally aware and strengthening it in a log-scale correction~\citep{yang2026rope,puvvada2025swan}. For LA-NoPE hybrids, we analogously propose \textbf{Sliding-Window Linear Attention}, which applies a sliding window to linear attention with gates. This may sound counterintuitive, because it appears to conflict with the recurrent nature of linear attention, but it can still be implemented. Without modifying FLA kernels~\citep{yang2024fla}, we use an approximation similar to sliding-window perplexity evaluation~\citep{presstrain}. Suppose the window size is 4096 and the stride is 1024. We can use the original GLA or GDN operator to compute attention on chunks $[0,4095], [1024,5119], [2048,6143], \cdots$ in parallel, and then extract the outputs at $[0,4095], [4096,5119], [5120,6143], \cdots$ before stacking them back together. This implementation is independent of the internal details of a specific LA variant.

\begin{figure}[!tb]
  \noindent
  \centering
  \begin{subfigure}[b]{0.48\linewidth}
    \centering
    \includegraphics[width=\linewidth]{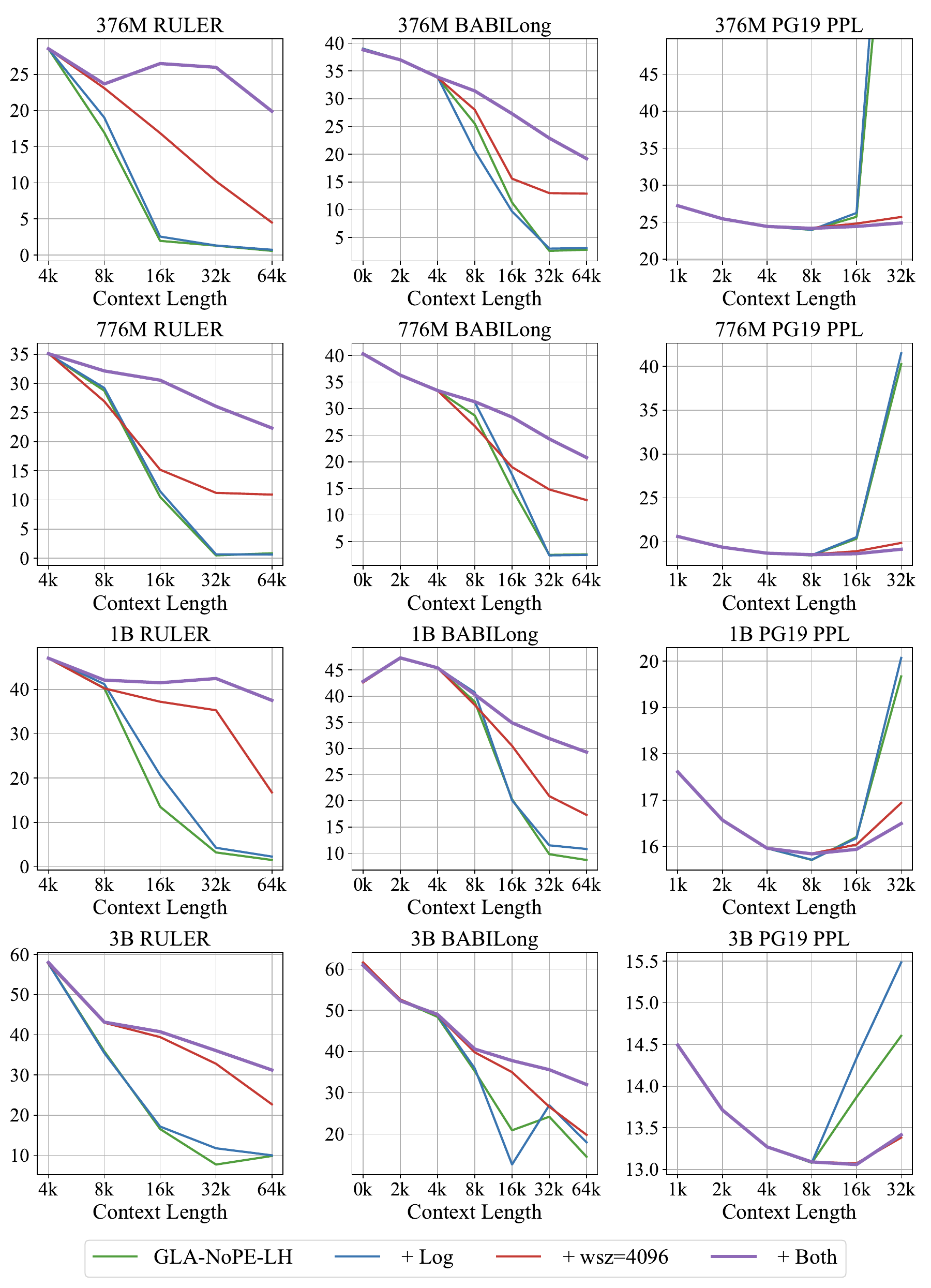}
    \caption{Results on GLA-NoPE-LH.}
  \end{subfigure}
  \hfill
  \begin{subfigure}[b]{0.48\linewidth}
    \centering
    \includegraphics[width=\linewidth]{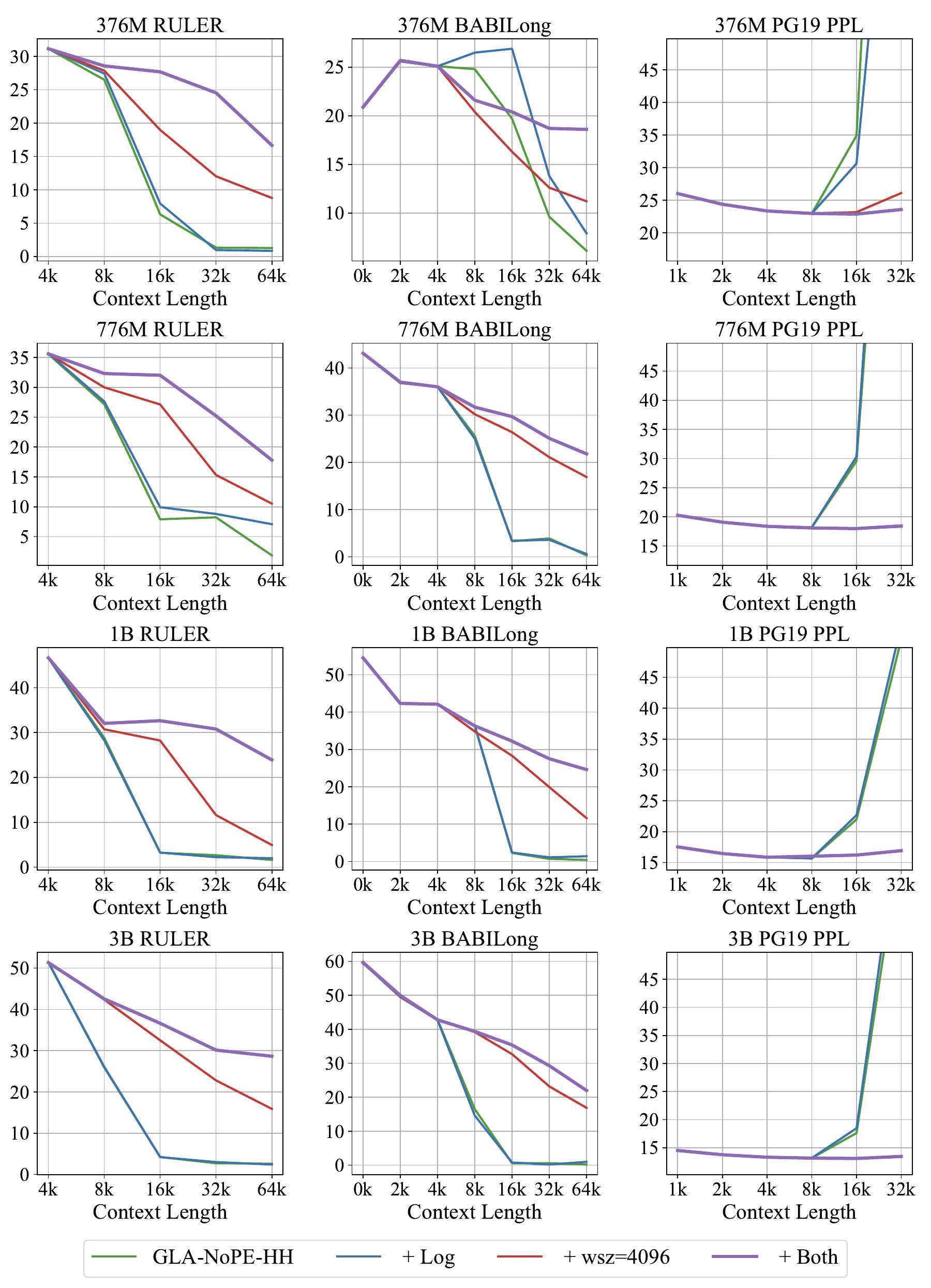}
    \caption{Results on GLA-NoPE-HH.}
  \end{subfigure}
  \caption{Comparison of length extrapolation in GLA-NoPE hybrid models. Applying a sliding window for GLA and a log scale for NoPE attention achieves the best length extrapolation performance.}
  \label{fig-matthew-gla}
\end{figure}

\begin{figure}[!tb]
  \noindent
  \centering
  \begin{subfigure}[b]{0.48\linewidth}
    \centering
    \includegraphics[width=\linewidth]{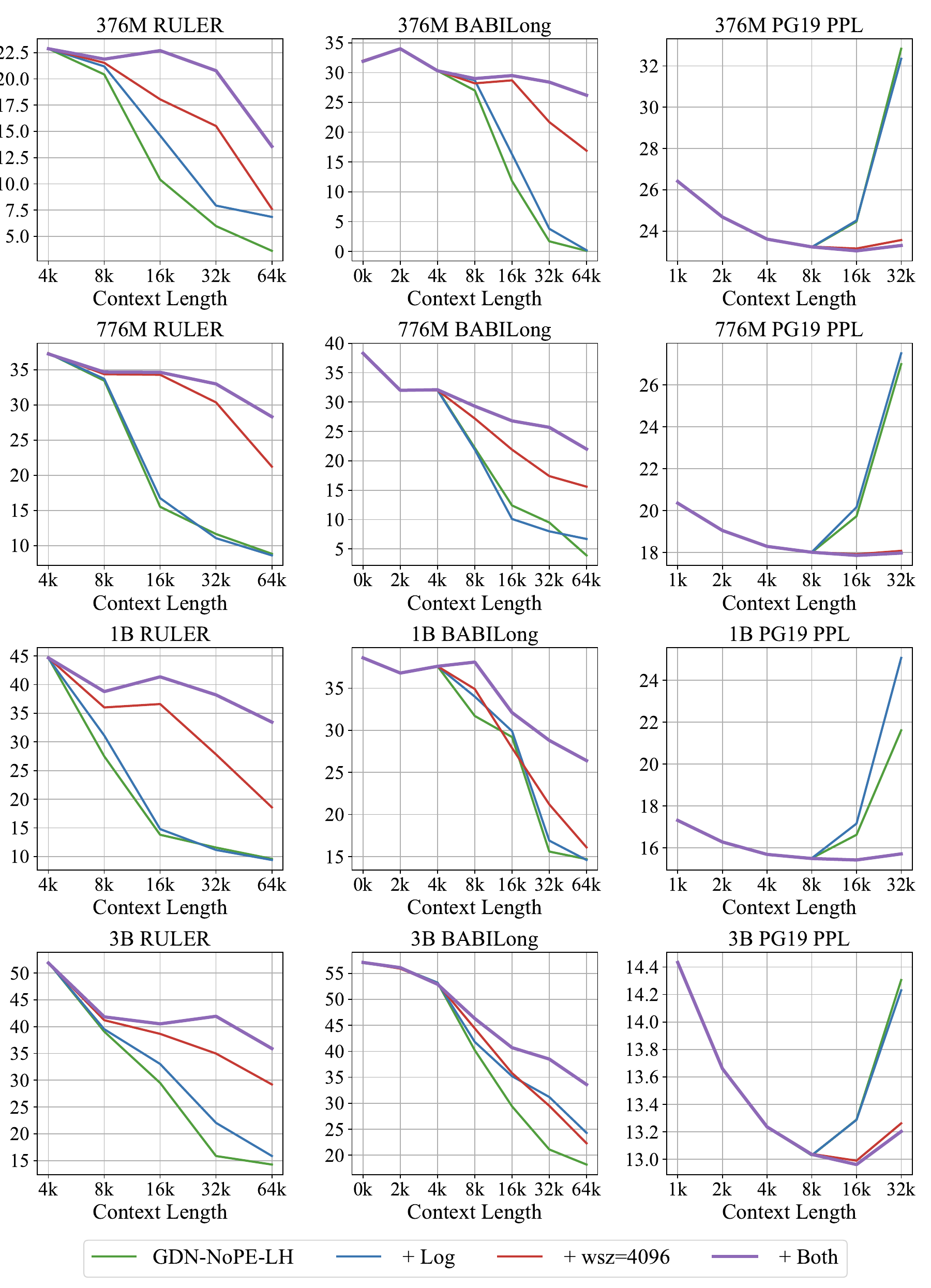}
    \caption{Results on Layer-wise Hybrid Model (LH).}
  \end{subfigure}
  \hfill
  \begin{subfigure}[b]{0.48\linewidth}
    \centering
    \includegraphics[width=\linewidth]{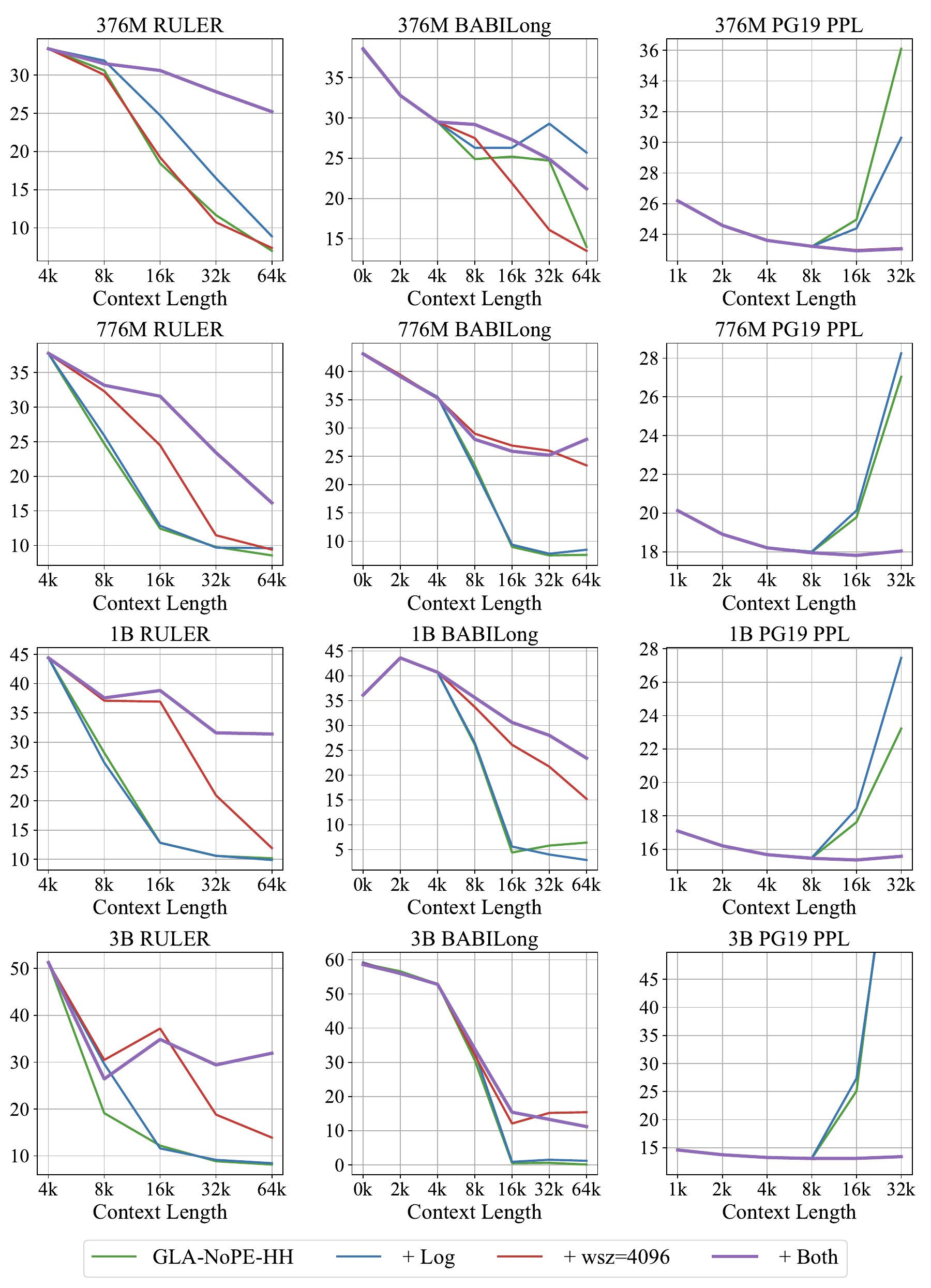}
    \caption{Results on Head-wise Hybrid Model (HH).}
  \end{subfigure}
  \caption{Comparison of length extrapolation in GDN-NoPE hybrid models. Still, applying a sliding window for GDN and a log scale for NoPE attention achieves the best length extrapolation performance in most cases.}
  \label{fig-matthew-gdn}
\end{figure}

From the internal logic of linear attention, using GLA as an example, the original interaction between $\bm{q}_t$ and $\bm{k}_s$ is multiplied by $\prod_{i=s+1}^t{\mathrm{Diag}{(\bm{\alpha}_i)}}$. After applying Sliding-Window Linear Attention, the number of factors accumulated in this product is bounded by the window size, implying that the relative-position signal is also restricted to the training length. As shown in Figures~\ref{fig-matthew-gla} and \ref{fig-matthew-gdn}, GLA-NoPE and GDN-NoPE hybrids with Sliding-Window Linear Attention achieve clear improvements in length extrapolation. Moreover, directly adding log-scale correction to the NoPE attention in LA-NoPE hybrids brings only limited improvement, but after applying SWLA, log-scale correction becomes much more effective. This strategy effectively relieves the no-free-lunch effect in length extrapolation of LA hybrids and also implies that \textbf{even for linear attention with implicit positional bias, extrapolation should still restrict the relative-position range}. This gives \textbf{Takeaway 6} of this paper: in length extrapolation of hybrid-position models, position-biased attention should use a sliding window to restrict its relative-position information, while NoPE attention should strengthen its global perception. We call this phenomenon the \textbf{Matthew Effect of Hybrid Position Extrapolation}, 
\begin{tcolorbox}[float,colback=blue!5!white,colframe=blue!75!black,title=\paragraph{Takeaway 6: Matthew Effect of Hybrid Position Extrapolation}\ ]
Applying a sliding window in position-biased attention (i.e., SWA for RoPE attention and SWLA for linear attention with gates) and enhanced global aggregation in NoPE attention by increasing the attention-logit scale can lead to strong length extrapolation for hybrid models.
\end{tcolorbox}
and abbreviate the \textbf{Extrapolation based on Matthew effect} as EME. This gives a simple yet effective extrapolation strategy, and on RoPE-NoPE hybrids it outperforms classical interpolation methods.

\begin{figure}[!tb]
  \noindent
  \centering
  \begin{subfigure}[b]{0.48\linewidth}
    \centering
    \includegraphics[width=\linewidth]{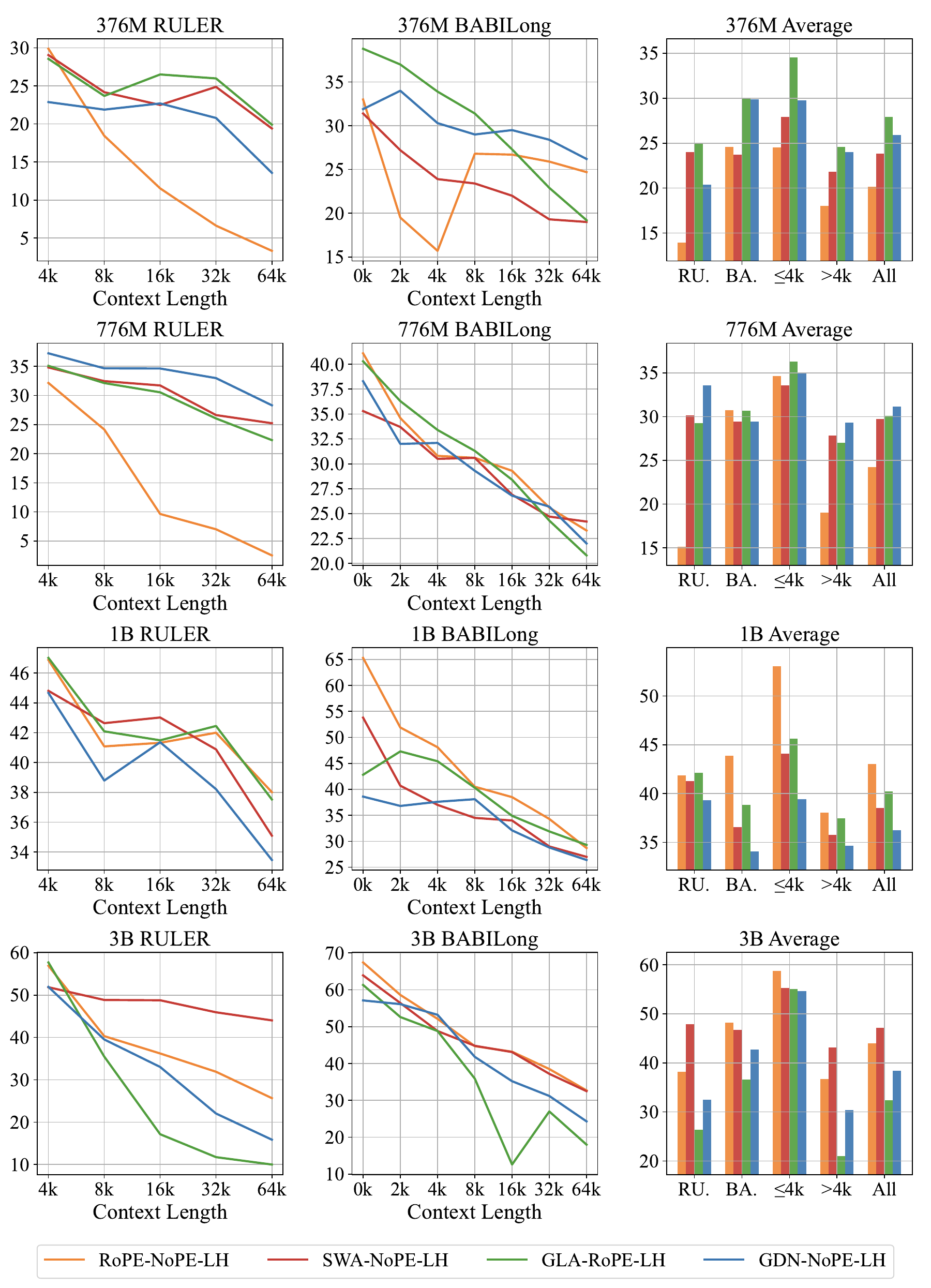}
    \caption{Results on Layer-wise Hybrid Model (LH).}
  \end{subfigure}
  \hfill
  \begin{subfigure}[b]{0.48\linewidth}
    \centering
    \includegraphics[width=\linewidth]{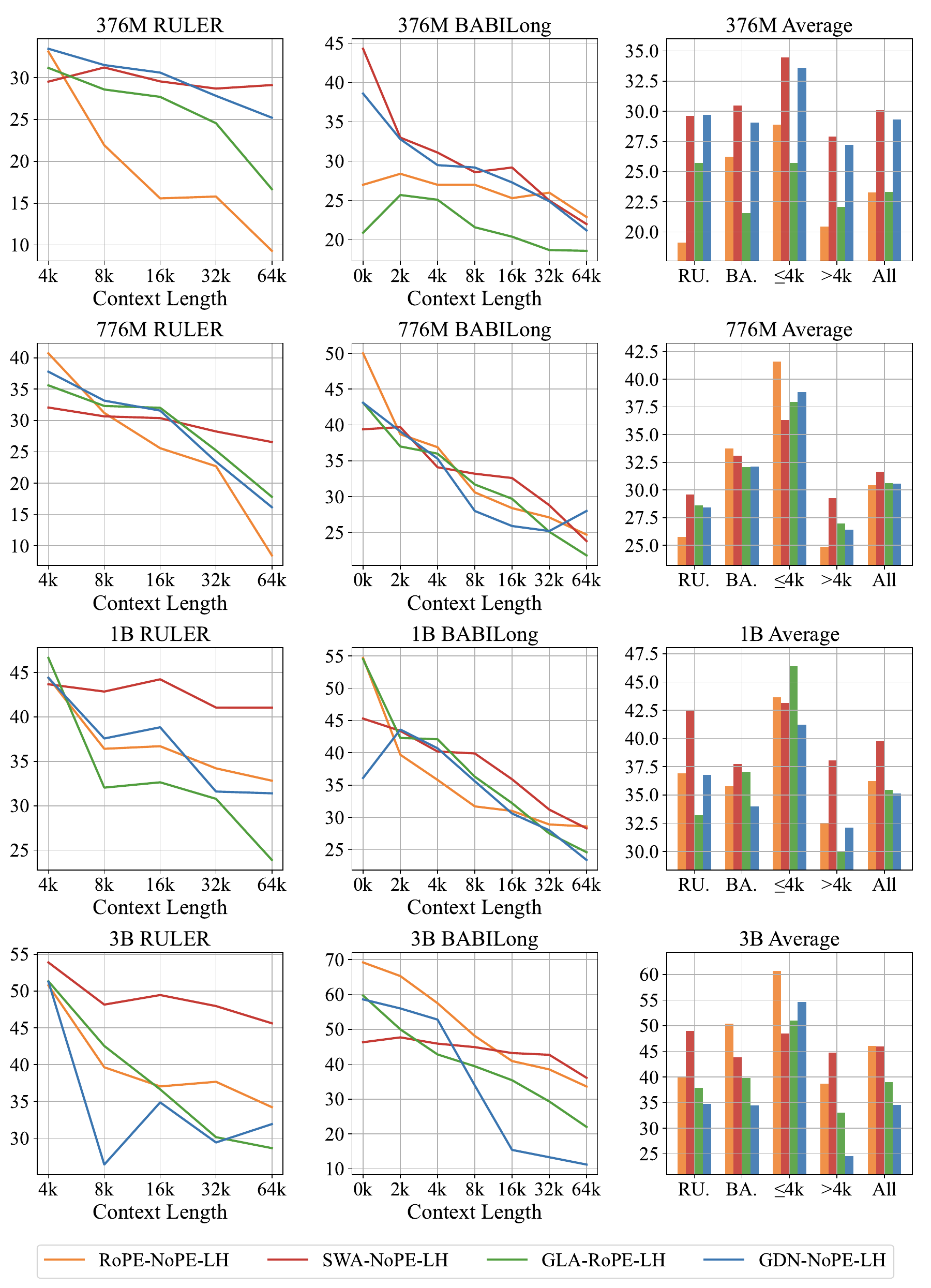}
    \caption{Results on Head-wise Hybrid Model (HH).}
  \end{subfigure}
  \caption{Comparison of EME in RoPE-NoPE, GLA-NoPE, and GDN-NoPE hybrids as well as SWA-NoPE hybrids.}
  \label{fig-matthew-all}
\end{figure}

\begin{table}[!ht]
\small
\tabcolsep=0.12cm
\centering
\begin{tabular}{l ccccc ccccc ccccc}
\toprule 
& \multicolumn{5}{c}{\textbf{NIAH-SK1}} & \multicolumn{5}{c}{\textbf{NIAH-SK2}} & \multicolumn{5}{c}{\textbf{NIAH-SK3}} \\ 
\cmidrule(lr){2-6} \cmidrule(lr){7-11} \cmidrule(lr){12-16}
 & 4k & 8k & 16k & 32k & 64k & 4k & 8k & 16k & 32k & 64k & 4k & 8k & 16k & 32k & 64k \\ 
\midrule
\multicolumn{16}{l}{\textbf{\textit{776M Short}}} \\[0.2ex]
GLA-NoPE-LH & 99.0 & 95.0 & 80.0 & 0.0 & 0.0 & 99.0 & 69.0 & 0.0 & 0.0 & 0.0 & 90.0 & 88.0 & 0.0 & 0.0 & 0.0 \\
+ Log & 99.0 & 99.0 & 98.0 & 0.0 & 0.0 & 99.0 & 68.0 & 1.0 & 0.0 & 0.0 & 90.0 & 91.0 & 0.0 & 0.0 & 0.0 \\
+ SWLA & 99.0 & \textbf{100.0} & \textbf{100.0} & 97.0 & 97.0 & 99.0 & 74.0 & 5.0 & 0.0 & 0.0 & 90.0 & 56.0 & 21.0 & 8.0 & 3.0 \\
+ SWLA \& Log (EME) & 99.0 & \textbf{100.0} & \textbf{100.0} & 96.0 & \textbf{100.0} & 99.0 & 92.0 & 87.0 & 72.0 & 59.0 & 90.0 & \textbf{79.0} & \textbf{75.0} & \textbf{76.0} & \textbf{64.0} \\
\hdashline
GLA-NoPE-HH & \textbf{100.0} & \textbf{100.0} & 68.0 & 82.0 & 0.0 & \textbf{100.0} & 71.0 & 0.0 & 0.0 & 0.0 & 65.0 & 45.0 & 0.0 & 0.0 & 0.0 \\
+ Log & \textbf{100.0} & \textbf{100.0} & 98.0 & 95.0 & 72.0 & \textbf{100.0} & 68.0 & 0.0 & 0.0 & 0.0 & 65.0 & 49.0 & 0.0 & 0.0 & 0.0 \\
+ SWLA & \textbf{100.0} & \textbf{100.0} & \textbf{100.0} & 92.0 & 75.0 & \textbf{100.0} & \textbf{100.0} & 82.0 & 32.0 & 1.0 & 65.0 & 62.0 & 44.0 & 9.0 & 1.0 \\
+ SWLA \& Log (EME) & \textbf{100.0} & \textbf{100.0} & \textbf{100.0} & \textbf{100.0} & \textbf{100.0} & \textbf{100.0} & \textbf{100.0} & 97.0 & 66.0 & 31.0 & 65.0 & 68.0 & 65.0 & 62.0 & 34.0 \\
\hdashline
GDN-NoPE-LH & \textbf{100.0} & \textbf{100.0} & \textbf{100.0} & \textbf{100.0} & 99.0 & 99.0 & 98.0 & 2.0 & 0.0 & 0.0 & 62.0 & 71.0 & 5.0 & 0.0 & 0.0 \\
+ Log & \textbf{100.0} & \textbf{100.0} & \textbf{100.0} & \textbf{100.0} & \textbf{100.0} & 99.0 & 98.0 & 9.0 & 1.0 & 0.0 & 62.0 & 70.0 & 9.0 & 0.0 & 0.0 \\
+ SWLA & \textbf{100.0} & \textbf{100.0} & \textbf{100.0} & \textbf{100.0} & 99.0 & 99.0 & 97.0 & 96.0 & 92.0 & 53.0 & 62.0 & 64.0 & 61.0 & 56.0 & 31.0 \\
+ SWLA \& Log (EME) & \textbf{100.0} & \textbf{100.0} & \textbf{100.0} & \textbf{100.0} & \textbf{100.0} & 99.0 & 97.0 & \textbf{99.0} & \textbf{99.0} & \textbf{90.0} & 62.0 & 67.0 & 64.0 & 66.0 & 53.0 \\
\hdashline
GDN-NoPE-HH & \textbf{100.0} & \textbf{100.0} & \textbf{100.0} & 99.0 & 92.0 & \textbf{100.0} & 95.0 & 2.0 & 0.0 & 0.0 & \textbf{94.0} & 38.0 & 0.0 & 0.0 & 0.0 \\
+ Log & \textbf{100.0} & \textbf{100.0} & \textbf{100.0} & \textbf{100.0} & \textbf{100.0} & \textbf{100.0} & 93.0 & 3.0 & 0.0 & 0.0 & \textbf{94.0} & 43.0 & 0.0 & 0.0 & 0.0 \\
+ SWLA & \textbf{100.0} & 99.0 & \textbf{100.0} & 95.0 & 92.0 & \textbf{100.0} & 99.0 & 64.0 & 4.0 & 0.0 & \textbf{94.0} & 68.0 & 31.0 & 5.0 & 2.0 \\
+ SWLA \& Log (EME) & \textbf{100.0} & \textbf{100.0} & \textbf{100.0} & \textbf{100.0} & \textbf{100.0} & \textbf{100.0} & \textbf{100.0} & 97.0 & 72.0 & 35.0 & \textbf{94.0} & 71.0 & 65.0 & 47.0 & 16.0 \\
\bottomrule
\end{tabular}
\caption{Extrapolation comparison of the 776M GLA/GDN-NoPE hybrid model on the NIAH-SK1, SK2, and SK3 tasks. Applying a sliding-window linear attention and a log scale for NoPE attention achieves the best length extrapolation. \label{tab-niah-matthew-la}}
\end{table}

In this paper, we verify the effectiveness of EME on RoPE-NoPE, GLA-NoPE, and GDN-NoPE hybrids, and further show that, for RoPE-NoPE hybrids, EME outperforms traditional interpolation methods. Figure~\ref{fig-matthew-all} summarizes the comparison among three hybrid models extrapolated with EME and SWA-NoPE extrapolation. The four models show roughly comparable performance. We also highlight the results on the NIAH-SK1, SK2, and SK3 tasks in RULER, which are the most frequently reported tasks in hybrid model research. GLA-NoPE and GDN-NoPE with EME achieve 16$\times$ training-free length extrapolation from a 4k training length to 64k inference length, while maintaining 100\% accuracy on SK1. The results at the 776M scale are shown in Table~\ref{tab-niah-matthew-la}, with more reported in Table~\ref{tab-niah-matthew-la-all}. In fact, our finding is also consistent with reports from other work. For example, in Figure 5 of \citet{chen2026hybrid}, the best extrapolating linear-attention hybrid is HypeNet-Lightning, and long-range decay is a key feature of Lightning Attention~\citep{qin2024various}. \textbf{Our approach based on SWLA further amplifies this property through the sliding window, breaks the traditional boundaries between SWA and LA, and is potentially compatible with arbitrary LA variants.}

In addition to the aforementioned discussion of length extrapolation and context extension phenomena in hybrid models, we also include further validation experiments and open discussions in Appendix~\ref{app-more}. 
\begin{itemize}
\item In Appendix~\ref{app-tasks}, we also add evaluation results on LongBench~\citep{bai2024longbench}, a more realistic and comprehensive benchmark for long-context tasks. Although small-scale pre-trained models perform relatively poorly, the results on In-Context Learning (ICL) tasks~\citep{brown2020language} still validate these takeaways.
\item In Appendix~\ref{app-verify}, we first demonstrate that the Seesaw Effect persists under different long-context learning rates and YOCO-style layer-wise hybrid~\citep{sun2024you}. Then, we also validate the effectiveness of the enhanced recipe for context extension in SWA hybrids under different layer-wise hybrid layouts.
\item In Appendix~\ref{app-open}, we also include discussions on topics where conclusions are not yet definitive, such as the selection of the rotary base for SWA in SWA hybrids, performance differences within LA-based hybrids, transfer training experiments and more analysis of hybrid position.
\end{itemize}
Furthermore, Appendix~\ref{app-detail} provides detailed evaluation scores, along with results from additional standard short-context tasks~\citep{openllm}. We believe our work contributes to a better understanding of long-context phenomena in hybrid models and fosters the development of the hybrid model research community.

\subsection{Efficiency Comparison}

\begin{figure}[!tb]
  \noindent
  \centering
  \begin{subfigure}[b]{0.49\linewidth}
    \centering
    \includegraphics[width=\linewidth]{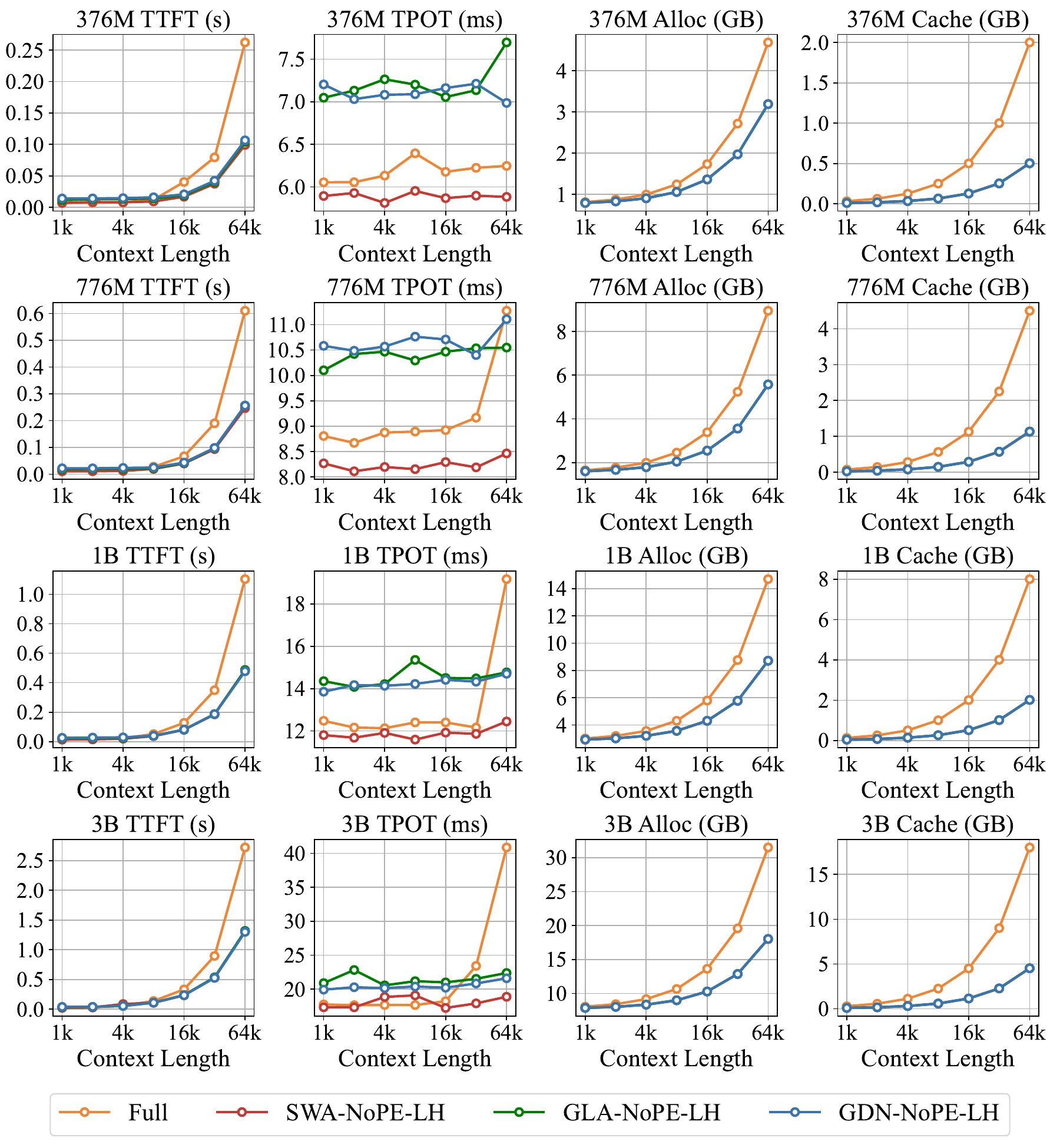}
    \caption{Results on Layer-wise Hybrid Model (LH).}
  \end{subfigure}
  \hfill
  \begin{subfigure}[b]{0.49\linewidth}
    \centering
    \includegraphics[width=\linewidth]{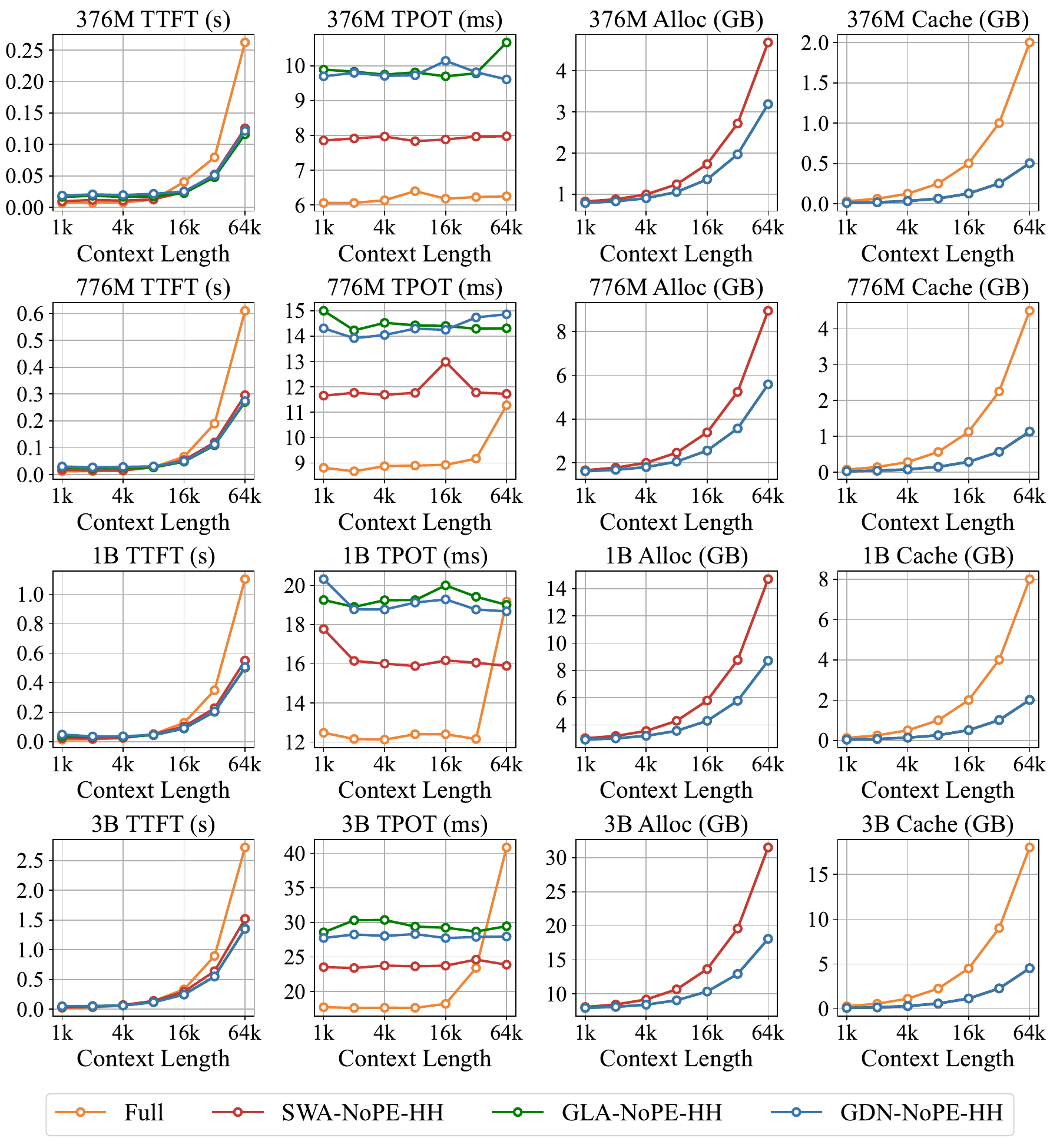}
    \caption{Results on Head-wise Hybrid Model (HH).}
  \end{subfigure}
  \caption{A brief comparison between the traditional full attention model and hybrid models on inference efficiency.}
  \label{fig-scatter-effi_infer}
\end{figure}

\begin{figure}[!tb]
  \noindent
  \centering
  \begin{subfigure}[b]{0.49\linewidth}
    \centering
    \includegraphics[width=\linewidth]{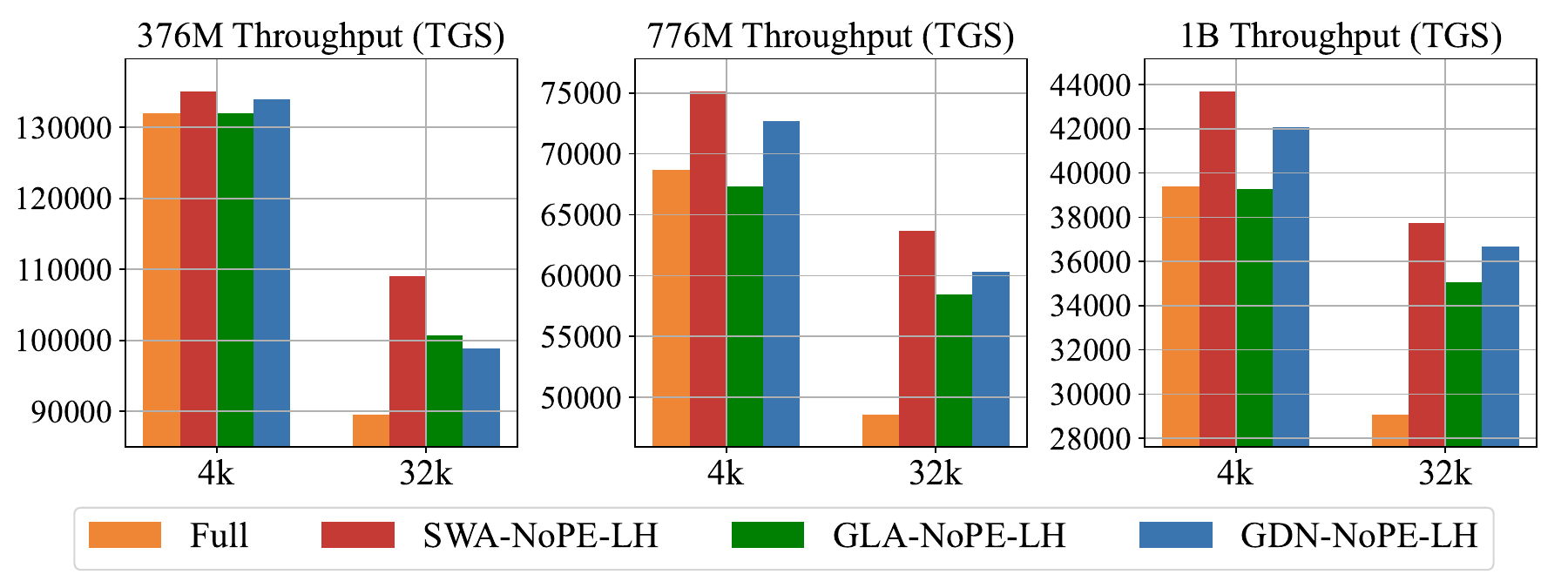}
    \caption{Results on Layer-wise Hybrid Model (LH).}
  \end{subfigure}
  \hfill
  \begin{subfigure}[b]{0.49\linewidth}
    \centering
    \includegraphics[width=\linewidth]{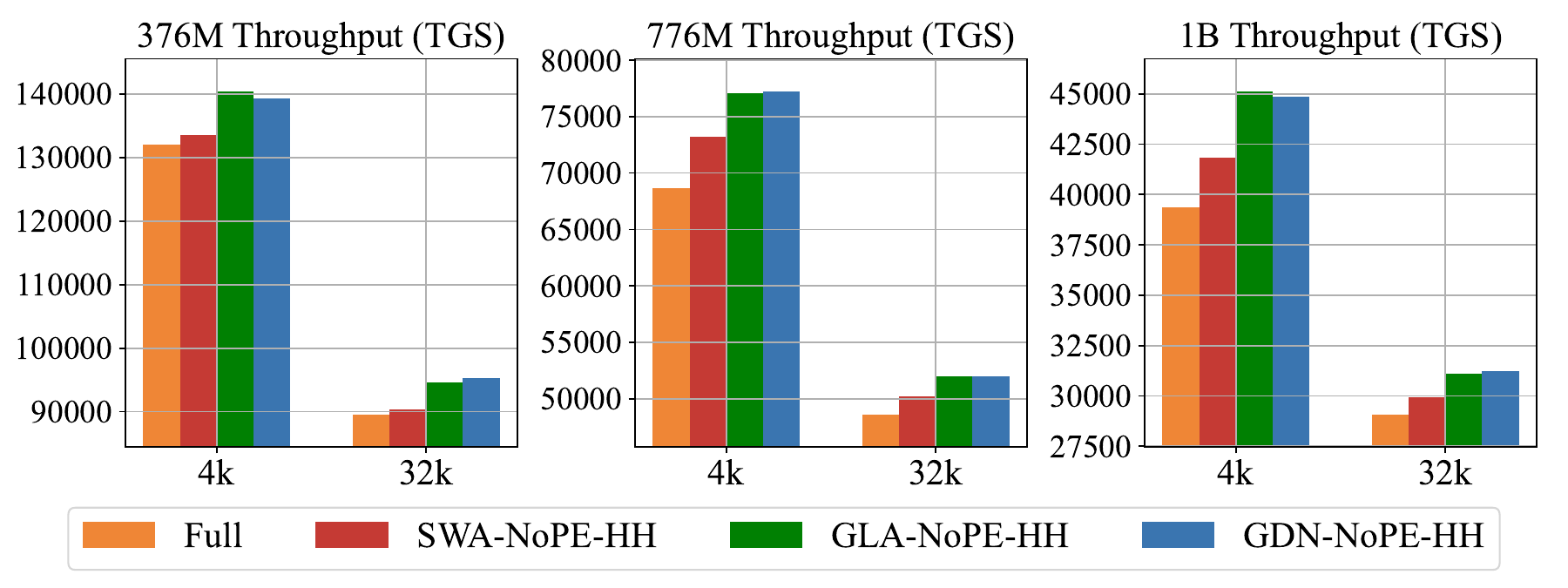}
    \caption{Results on Head-wise Hybrid Model (HH).}
  \end{subfigure}
  \caption{A brief comparison between the traditional full attention model and hybrid models on training efficiency.}
  \label{fig-scatter-effi_train}
\end{figure}

Additionally, we include a brief comparison of inference and training efficiency, as shown in Figure~\ref{fig-scatter-effi_infer} and \ref{fig-scatter-effi_train}. Regarding inference efficiency, we compare traditional full-attention-only models with 3:1 layer-wise or head-wise hybrid models that combine SWA, GLA, or GDN with NoPE full attention, across model sizes of 376M, 776M, 1B, and 3B. We measure prefilling efficiency with Time to First Token (TTFT) and decoding efficiency with Time per Output Token (TPOT), both in microseconds. We evaluate memory efficiency based on peak allocated GPU memory and cache size, both measured in GB. As shown in Figure~\ref{fig-scatter-effi_infer}, hybrid models offer clear advantages as the model size and context length increase. Notably, although head-wise hybrids compute attention twice in every layer, the advantage in larger model sizes and longer context lengths is still remarkable. Similarly, regarding training efficiency, we compare training throughput on 8 H200 GPUs, measured by TGS (tokens per GPU per second) in 4k and 32k context lengths. We still compare full-attention-only models with 3:1 layer-wise or head-wise hybrid models on 376M, 776M, and 1B, without 3B, since training 3B models requires 16 H200 GPUs. As shown in Figure~\ref{fig-scatter-effi_train}, hybrid models still offer a larger throughput, especially for the 1B model size and 32k context length.

Moreover, because the GLA and GDN variants store their recurrent states in higher FP32 precision, their resulting cache sizes are comparable to the cache overhead of the SWA hybrid with a window size of 128 and FP16 precision, so the corresponding curves overlap in the memory-efficiency comparison. In addition, because the QKV matrices of head-wise SWA hybrids are concatenated rather than split, as in head-wise LA hybrids, the intermediate memory footprint is larger, so head-wise SWA hybrids share the same peak inference memory as full attention and offer lower training throughput than LA hybrids.

\section{Conclusion}

In this paper, we conduct extensive experiments to validate the effectiveness of the hybrid model in length extrapolation and context extension. Regarding hybrid position in NoPE attention versus other position-biased attention, we propose the Tidal Effect in collaboration and the Matthew Effect in extrapolation. We also discover the Seesaw Effect in context extension of the sliding-window attention hybrid models and the No-Free-Lunch Effect in length extrapolation of linear attention hybrid models. For SWA hybrids, we identify short-window weariness as well as long-window laziness and enlarge the window during extended training to overcome its Short-Context Learning Trap. For LA hybrids, we propose the Sliding-Window Linear Attention, achieving 16$\times$ training-free extrapolation with 100\% accuracy on NIAH-SK1 in 64k context length.

\section*{Limitations}

This paper provides a mechanistic analysis of long-context hybrid models, with a broad training-validation-observation-induction loop. Even so, several directions remain open.
\begin{itemize}[leftmargin=*]
    \item \textbf{Validation scope}. Our current validation pipeline is focused on context extension and length extrapolation in the pretraining stage. We still need to extend it to post-training~\citep{zeng2024scaling,agiteam2025nexn1} and multimodal, especially vision-language~\citep{yang2025qwen3,wang2026moss,wei2025videorope} models. On that basis, we should also extend our evaluation to complex reasoning, agentic tasks, multimodal retrieval, and multimodal understanding.
    \item \textbf{Architecture scope}. Our analysis focuses on RoPE/NoPE hybrids, SWA hybrids, and LA hybrids. We still need to apply the same mechanistic lens to other efficient hybrid families, such as sparse-attention~\citep{liu2024deepseek,lai2026minimax,qwen3.8flashnext} and compressed-attention~\citep{xu2026deepseek,chu2026kwai} models. We leave this broader comparison for future work.
    \item \textbf{Mechanism scope}. Even within the hybrid families studied here, we only probe a subset of position and attention design choices. We will also further our study by analyzing techniques such as sink bias~\citep{xiaoefficient,agarwal2025gpt,mimo2026v25pro}, partial RoPE~\citep{zeng2026glm,qwen35blog}, gated attention variants~\citep{qiu2026gated,qwen35blog}, and ablation of short convolution~\citep{yang2025gated,yang2025path,inkling}.
\end{itemize}

\section*{Acknowledgement}

I am grateful to all advisors who have helped and encouraged me, and to all friends who still value and trust me. In particular, I want to thank Ms. Chunying Wang, a counselor at the Shanghai Innovation Institute, for her invaluable support, and Dr. Yutao Sun and Dr. Qipeng Guo for the academic exchanges we shared.

\bibliographystyle{unsrtnat}
\bibliography{main}

\beginappendix

\section{Extended Verification}\label{app-more}

We conduct experiments at 376M, 776M, 1B, and 3B model sizes, with the configuration detailed in Table~\ref{tab-config}. Our models use the same tokenizer as the Llama 3 Series~\citep{meta2024llama,meta2024introducing,dubey2024llama}. 

\begin{table}[!tb]
\centering\small
\begin{tabular}{lcccc}
\toprule
 & 376M & 776M & 1B & 3B \\
\midrule
Hidden Size & 1024 & 1536 & 2048 & 3072 \\
Intermediate Size & 3584 & 5376 & 7168 & 10752 \\
Num Layer & 8 & 12 & 16 & 24 \\
Num Attn Head & 8 & 12 & 16 & 24 \\
Num KV Head & 8 & 12 & 16 & 24 \\
Vocab Size & 128256 & 128256 & 128256 & 128256 \\
\bottomrule
\end{tabular}
\caption{The hyperparameters of different model sizes.\label{tab-config}}
\end{table}

All models are first pretrained on the DCLM-Baseline-1.0 corpus~\citep{li2024datacomp} in BFloat16 with a 4k context length using DeepSpeed~\citep {rasley2020deepspeed}, FlashAttention2~\citep{daoflashattention} and FlashLinearAttention~\citep{yang2024fla}. The 3B models use 16 NVIDIA H200 GPUs; all other models use 8 GPUs. For each size, we use a batch size of 1M tokens and pretrain for 50B tokens. We use the AdamW~\citep{loshchilov2017fixing} optimizer with weight decay 0.1, a maximum learning rate of 1e-3, and a cosine annealing scheduler. We use the first 0.5B tokens for warm-up, and the learning rate ends at 0. 

In long-context continual pretraining, we use holistic long documents from the high-quality long-context corpus LongWanjuan~\citep{lv2024longwanjuan}, and continue training at a 32k context length. We follow the data mixture ratios used in \citet{touvron2023llama} and \citet{lv2024longwanjuan}, and increase the rotary base from 10000 to 500000 for RoPE full-attention to support contexts of around 64k tokens based on the scaling laws of RoPE-based extrapolation~\citep{liuscaling,liu2025longllada}. We use a 0.5M batch size and continue training for 5B tokens. We use the AdamW~\citep{loshchilov2017fixing} optimizer with 0 weight decay, a peak learning rate of 1e-3, and a cosine annealing scheduler. We use the first 1B tokens for warm-up, and finally decay the learning rate to 0.

\subsection{More Verification on LongBench Tasks}\label{app-tasks}

\begin{figure}[!t]
  \noindent
  \centering
  \begin{subfigure}[b]{0.48\linewidth}
    \centering
    \includegraphics[width=\linewidth]{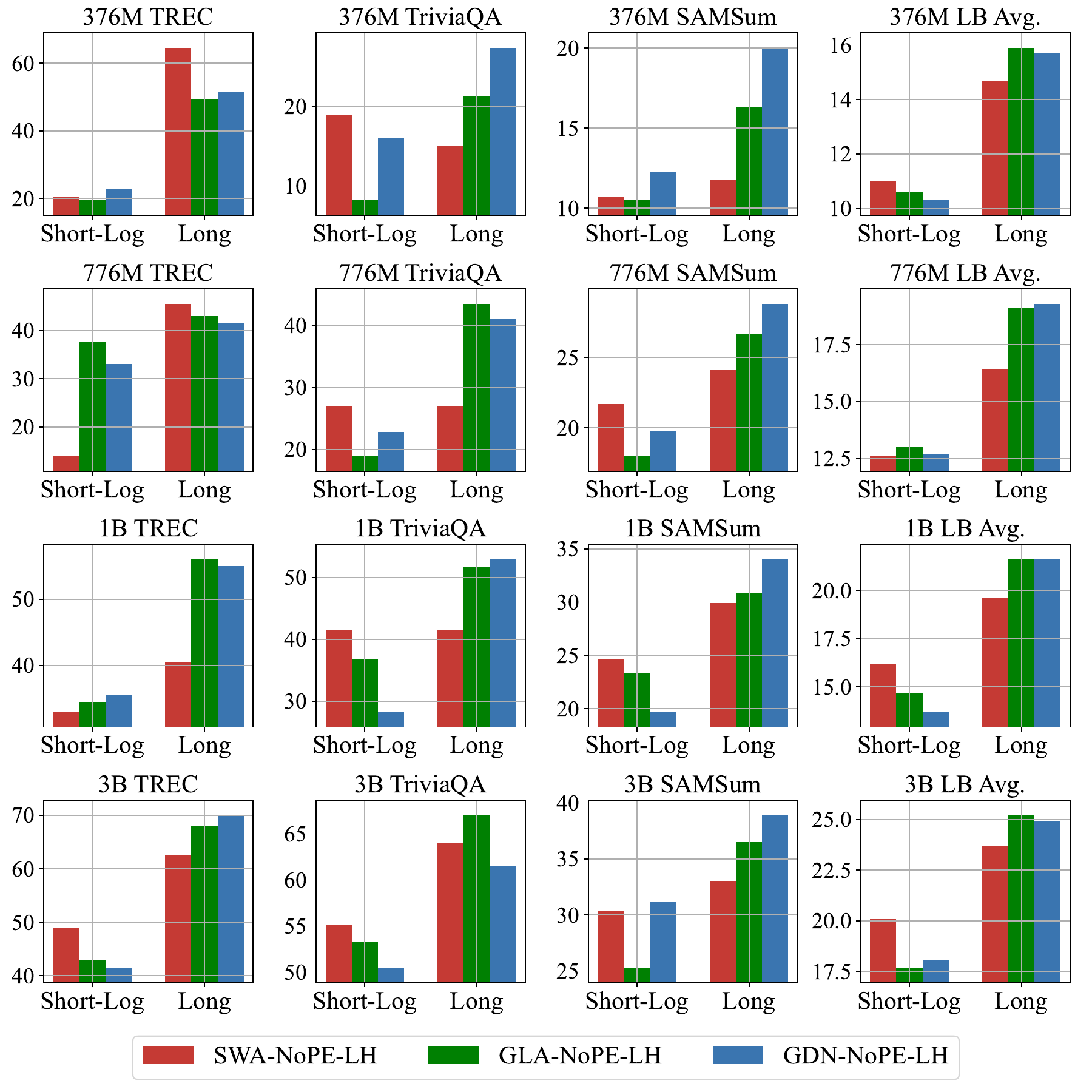}
    \caption{Results of layer-wise hybrid models.}
  \end{subfigure}
  \hfill
  \begin{subfigure}[b]{0.48\linewidth}
    \centering
    \includegraphics[width=\linewidth]{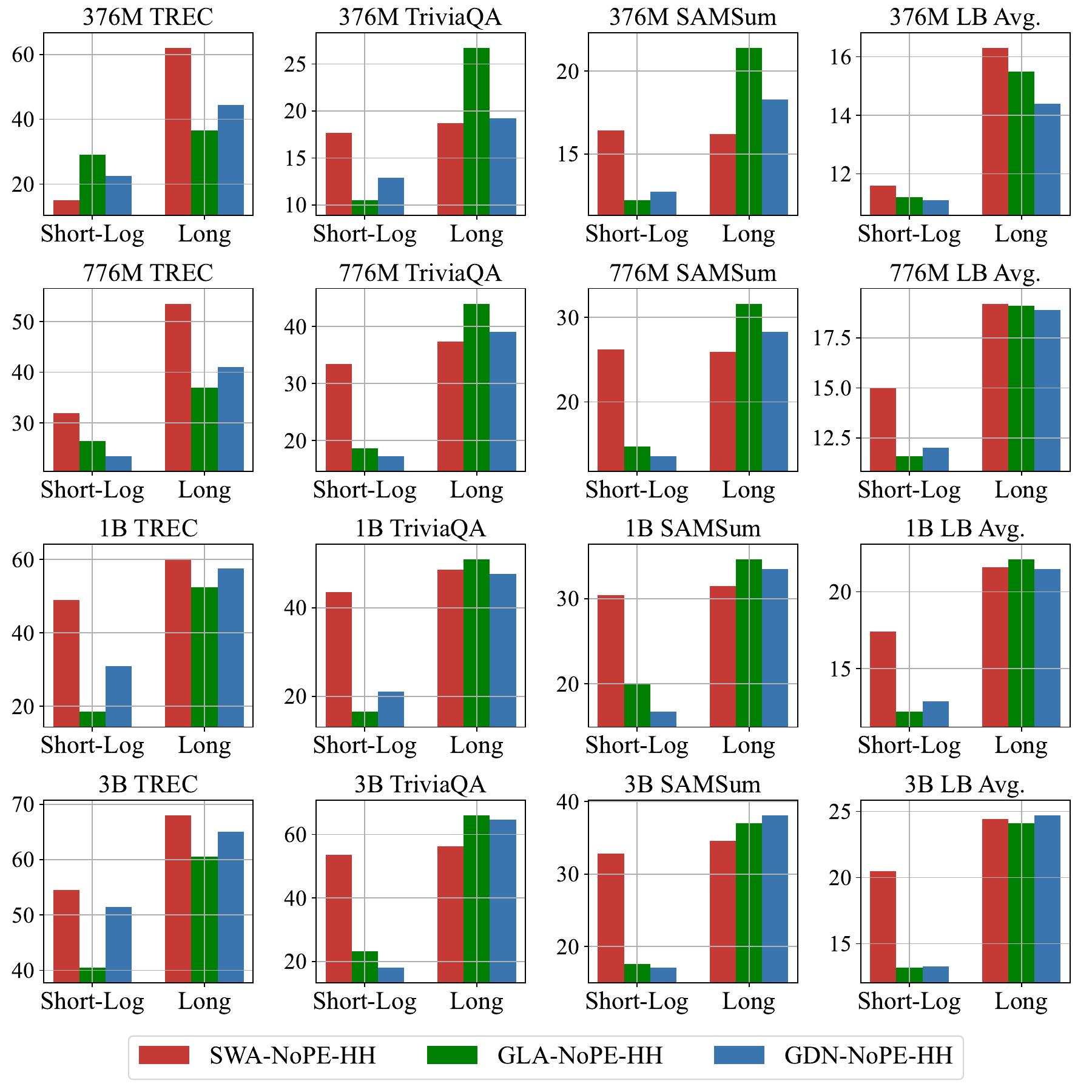}
    \caption{Results of head-wise hybrid models.}
  \end{subfigure}
  \caption{Comparison of the LongBench performance of the SWA/GLA/GDN-NoPE hybrid models under log-scale NoPE extrapolation (Short-Log) and context extension (Long). The Seesaw Effect still exists.}
  \label{fig-longbench-seesaw}
\end{figure}

\begin{figure}[!t]
  \noindent
  \centering
  \begin{subfigure}[b]{0.48\linewidth}
    \centering
    \includegraphics[width=\linewidth]{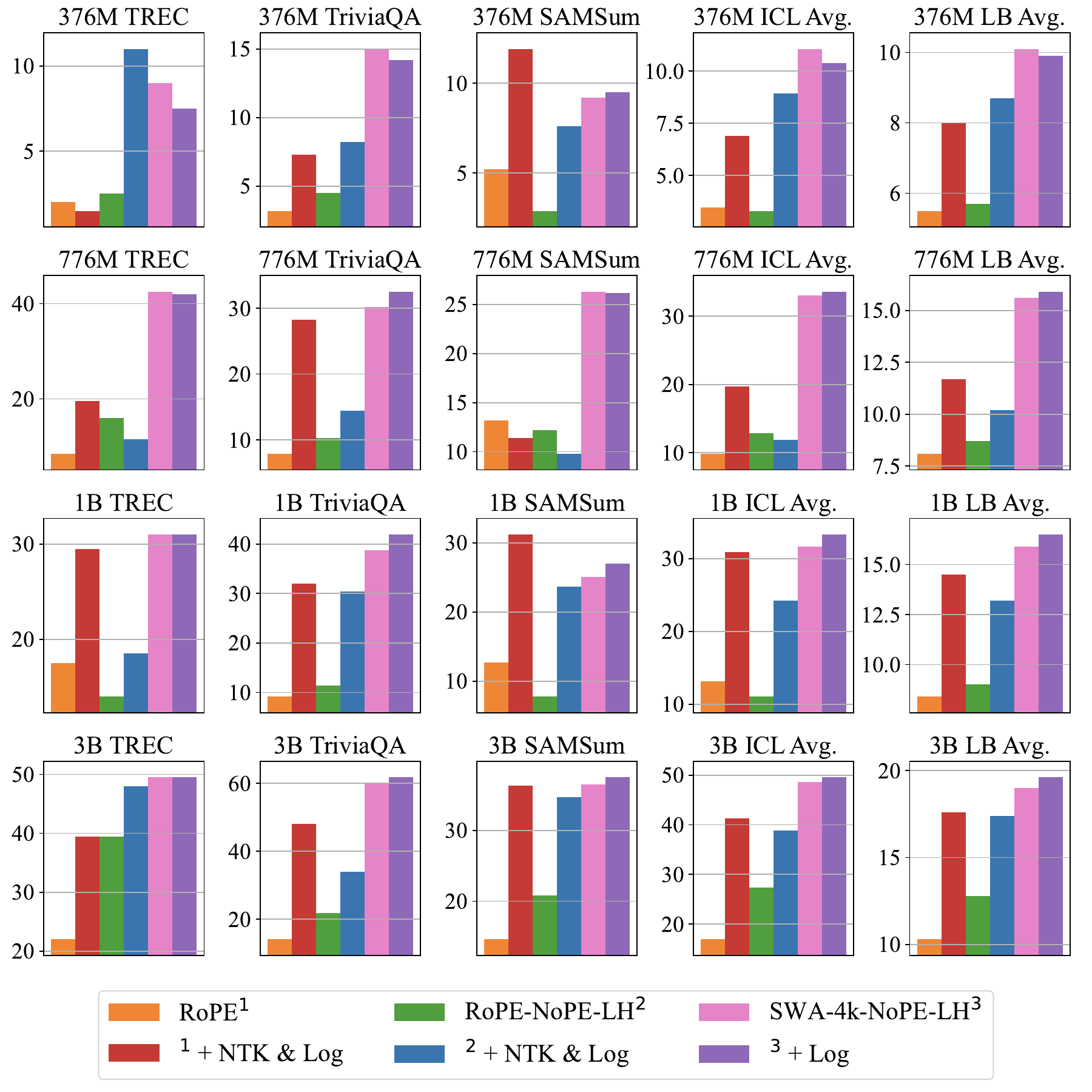}
    \caption{Results of layer-wise hybrid models.}
  \end{subfigure}
  \hfill
  \begin{subfigure}[b]{0.48\linewidth}
    \centering
    \includegraphics[width=\linewidth]{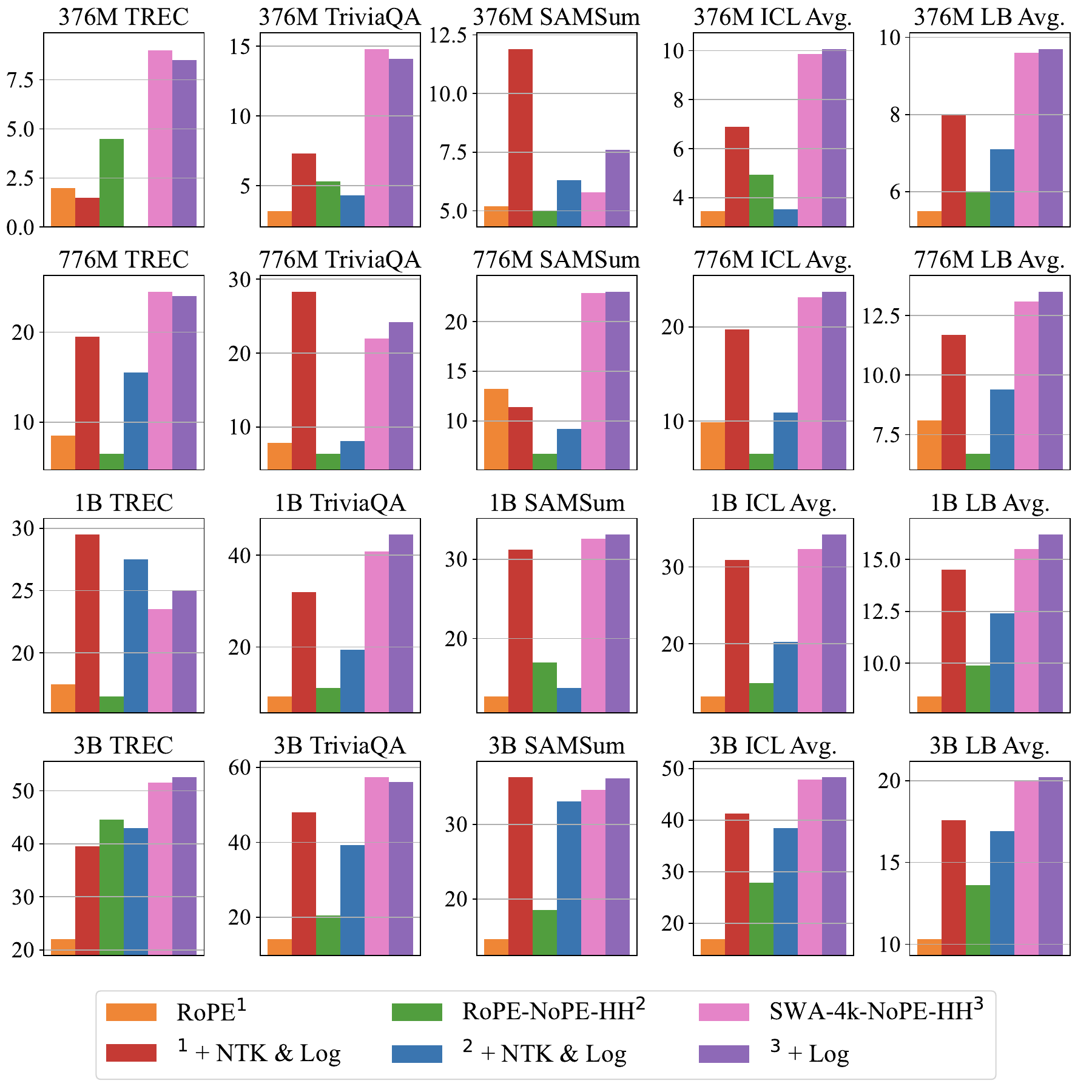}
    \caption{Results of head-wise hybrid models.}
  \end{subfigure}
  \caption{Comparison of LongBench performance in RoPE-NoPE hybrid models. Applying a sliding-window RoPE attention and a log-scale NoPE attention achieves the best length extrapolation. The Matthew Effect still works.}
  \label{fig-longbench-matthew-fa}
\end{figure}

\begin{figure}[!t]
  \noindent
  \centering
  \begin{subfigure}[b]{0.48\linewidth}
    \centering
    \includegraphics[width=\linewidth]{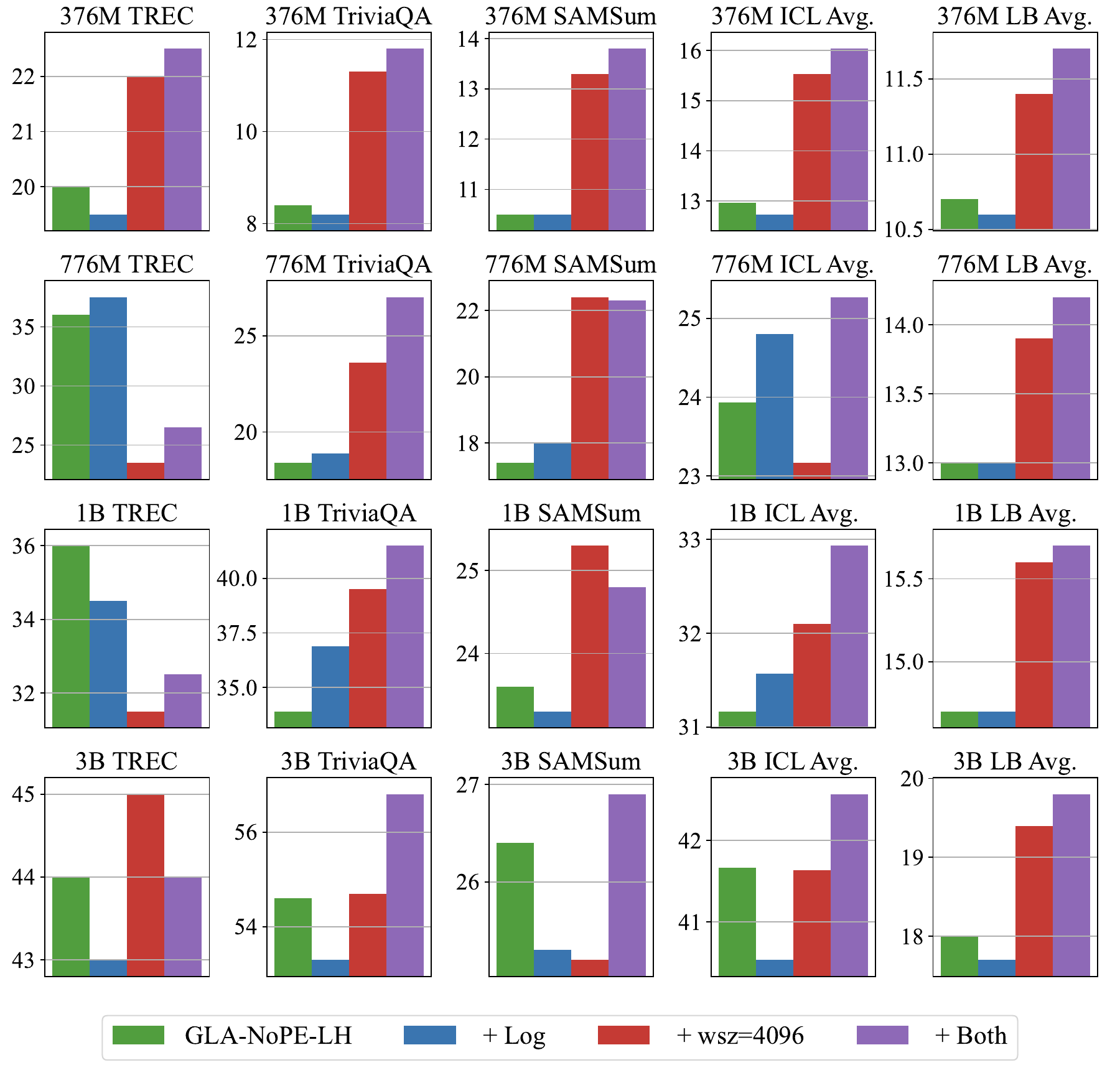}
    \caption{Results of layer-wise hybrid models.}
  \end{subfigure}
  \hfill
  \begin{subfigure}[b]{0.48\linewidth}
    \centering
    \includegraphics[width=\linewidth]{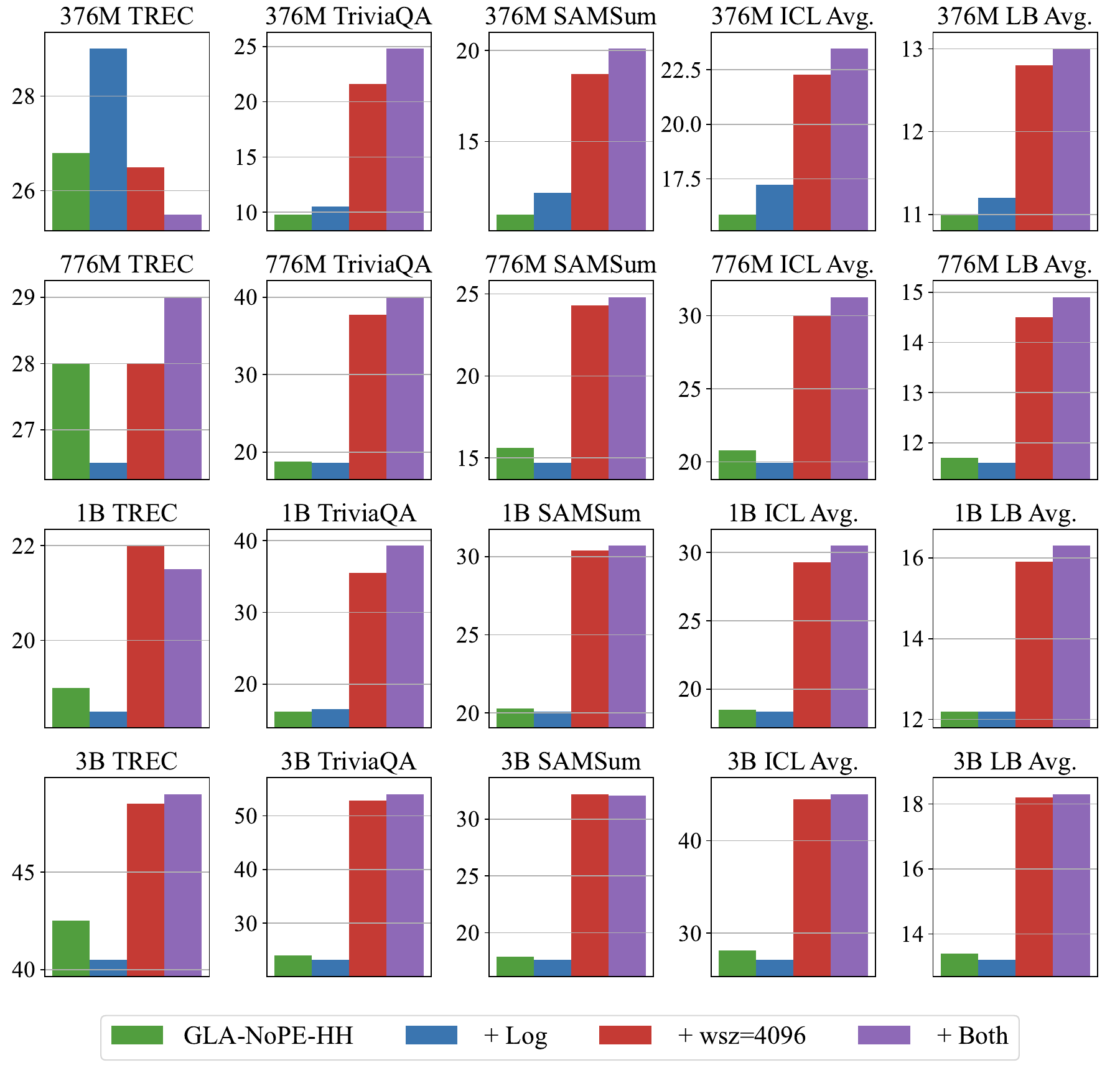}
    \caption{Results of head-wise hybrid models.}
  \end{subfigure}
  \caption{Comparison of LongBench performance in GLA-NoPE hybrid models. Applying a sliding-window GLA and a log-scale NoPE attention achieves the best length extrapolation. The Matthew Effect still works.}
  \label{fig-longbench-matthew-gla}
\end{figure}

\begin{figure}[!t]
  \noindent
  \centering
  \begin{subfigure}[b]{0.48\linewidth}
    \centering
    \includegraphics[width=\linewidth]{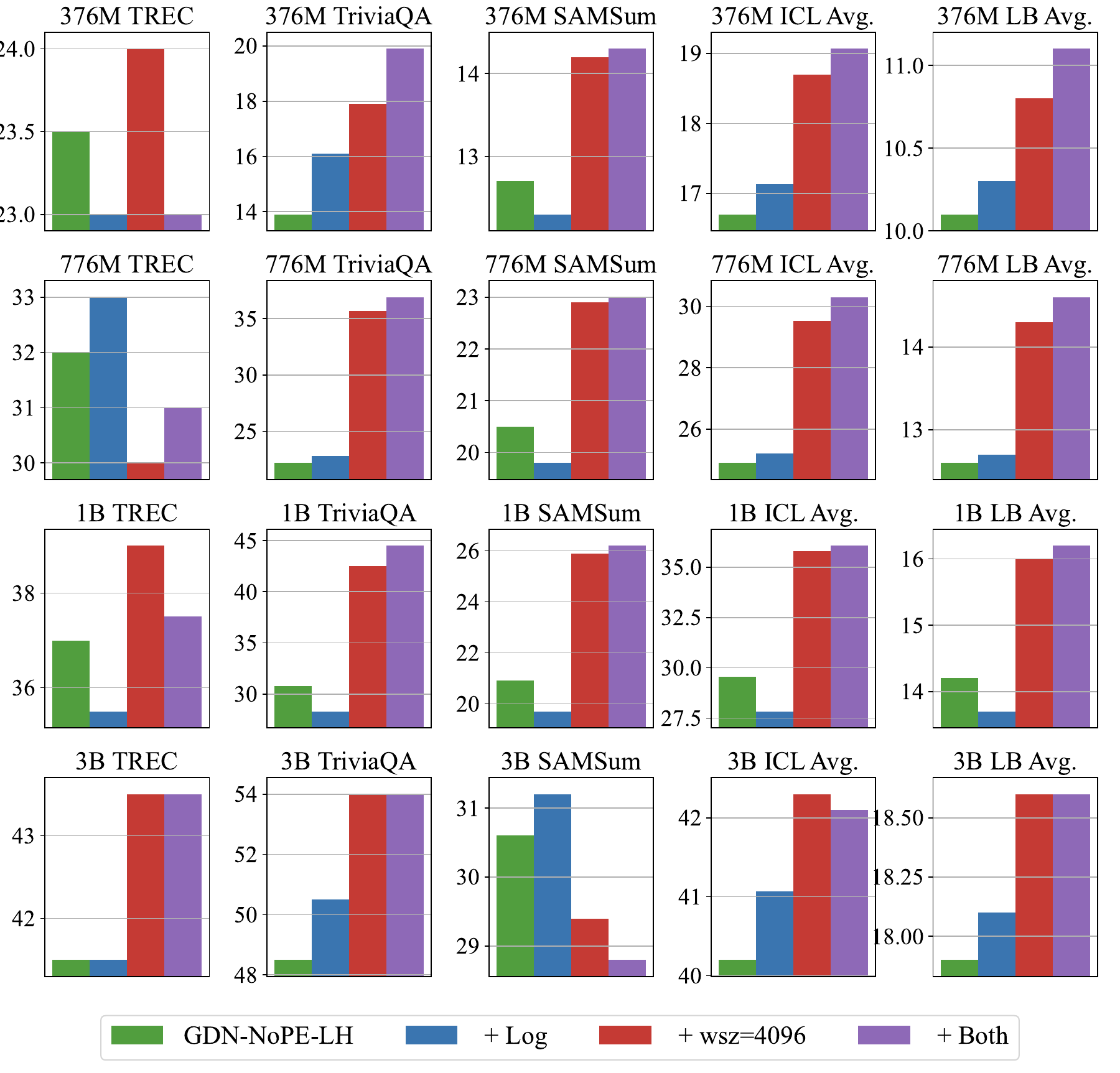}
    \caption{Results of layer-wise hybrid models.}
  \end{subfigure}
  \hfill
  \begin{subfigure}[b]{0.48\linewidth}
    \centering
    \includegraphics[width=\linewidth]{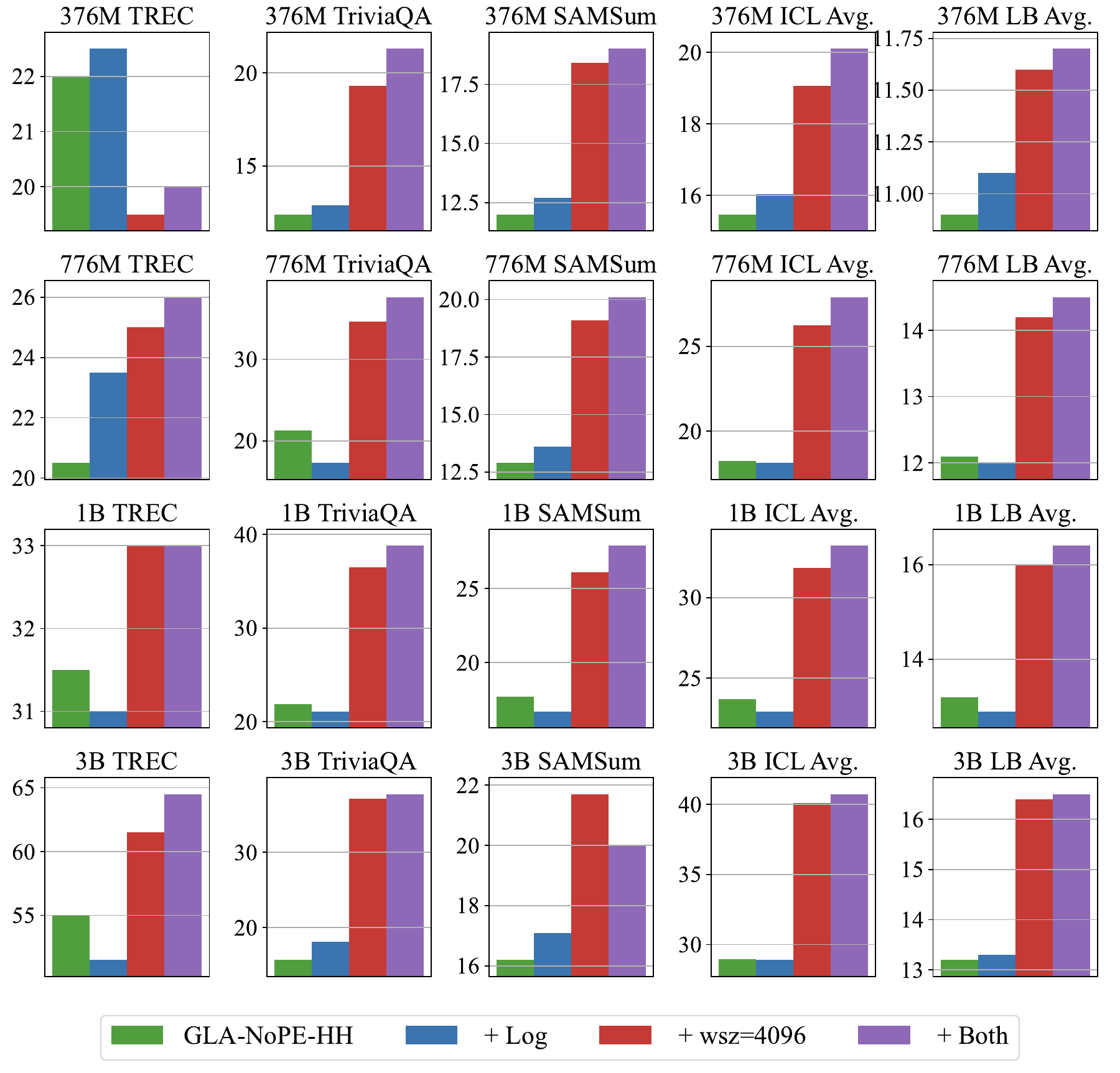}
    \caption{Results of head-wise hybrid models.}
  \end{subfigure}
  \caption{Comparison of LongBench performance in GDN-NoPE hybrid models. Applying a sliding-window GDN and a log-scale NoPE attention achieves the best length extrapolation. The Matthew Effect still works.}
  \label{fig-longbench-matthew-gdn}
\end{figure}

To validate the takeaways in this paper, we add evaluations on LongBench~\citep{bai2024longbench}, a more realistic and comprehensive benchmark for long-context tasks. The evaluation is conducted with a maximum input length of 31500 and a maximum output length of 500. Although small-scale pre-trained models perform relatively poorly, the results on In-Context Learning (ICL) tasks~\citep{brown2020language} still confirm our findings. Figure~\ref{fig-longbench-seesaw} compares SWA/GLA/GDN-NoPE hybrid models under log-scale NoPE extrapolation (Short-Log) and context extension (Long). The SWA-NoPE hybrids outperform in short-context extrapolation; however, after long-context training, layer-mixed LA-NoPE outperforms layer-mixed SWA-NoPE, while head-mixed LA-NoPE performs comparably to head-mixed SWA-NoPE, confirming the Seesaw Effect in LongBench evaluations. 

We also validate the Matthew Effect of hybrid position extrapolation on the LongBench benchmark; Figures~\ref{fig-longbench-matthew-fa}, \ref{fig-longbench-matthew-gla}, and \ref{fig-longbench-matthew-gdn} illustrate the extrapolation performance of RoPE-NoPE, GLA-NoPE, and GDN-NoPE, respectively, when employing this effect. We find that applying a sliding window for position-biased attention and a log scale for NoPE attention yields the best length extrapolation results on average. We also include detailed LongBench scores in Tables~\ref{tab-longbench-lh-short-small}, \ref{tab-longbench-lh-short-large}, \ref{tab-longbench-hh-short-small}, \ref{tab-longbench-hh-short-large}, \ref{tab-longbench-lh-long} and \ref{tab-longbench-hh-long}.

\subsection{More Verification of Seesaw Effect}\label{app-verify}

\begin{figure}[!t]
  \noindent
  \centering
  \begin{subfigure}[b]{0.48\linewidth}
    \centering
    \includegraphics[width=\linewidth]{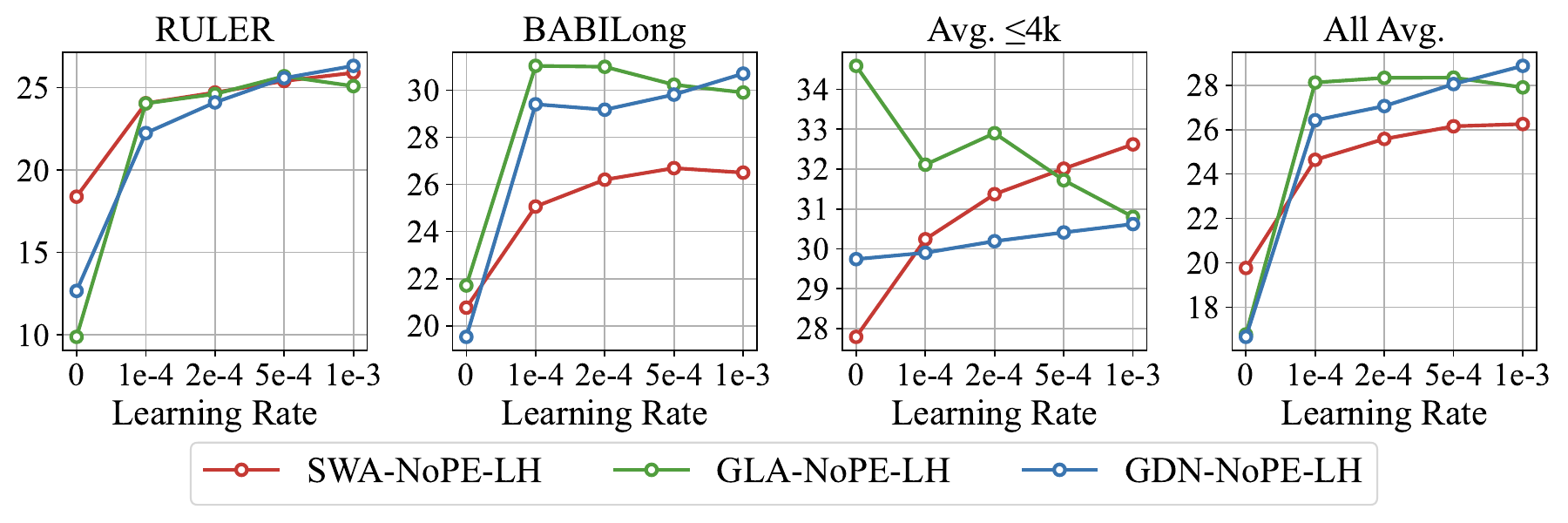}
    \caption{Results of 376M models.}
  \end{subfigure}
  \hfill
  \begin{subfigure}[b]{0.48\linewidth}
    \centering
    \includegraphics[width=\linewidth]{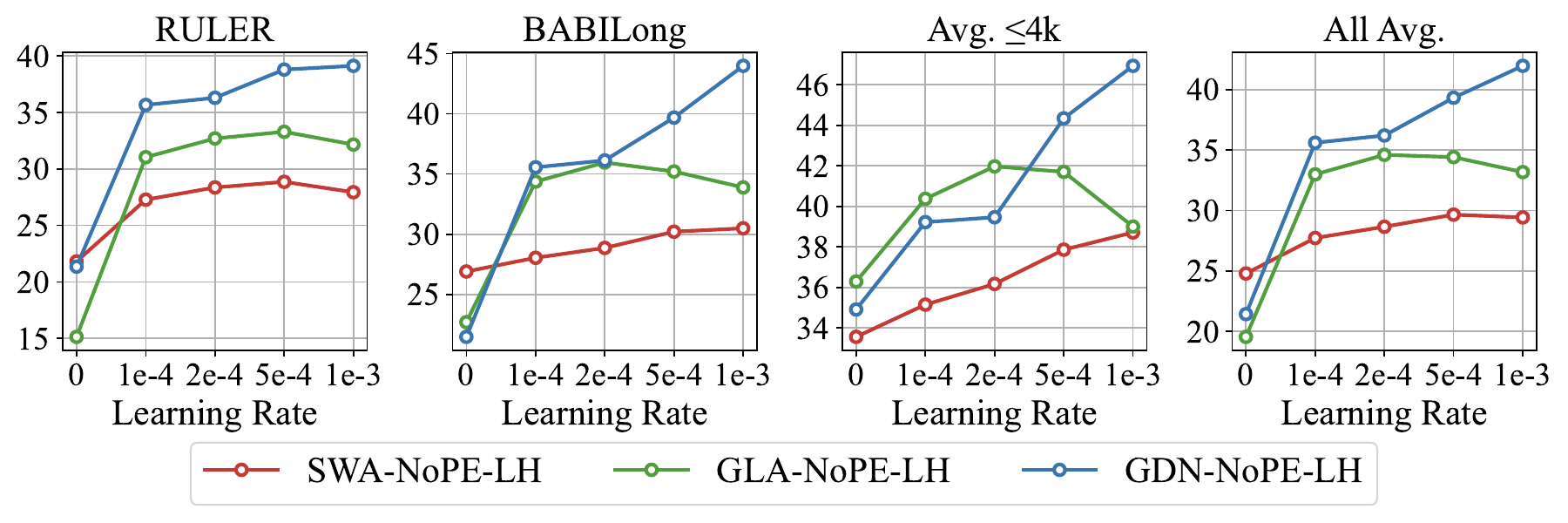}
    \caption{Results of 776M models.}
  \end{subfigure}
  \caption{Comparison of long-context training performance between SWA-NoPE-LH, GLA-NoPE-LH, and GDN-NoPE-LH under different learning rates, where 0 learning rate means the performance before long-context continual pretraining.}
  \label{fig-seesaw-verify}
\end{figure}

\begin{figure}[!t]
  \noindent
  \centering
  \begin{subfigure}[b]{0.48\linewidth}
    \centering
    \includegraphics[width=\linewidth]{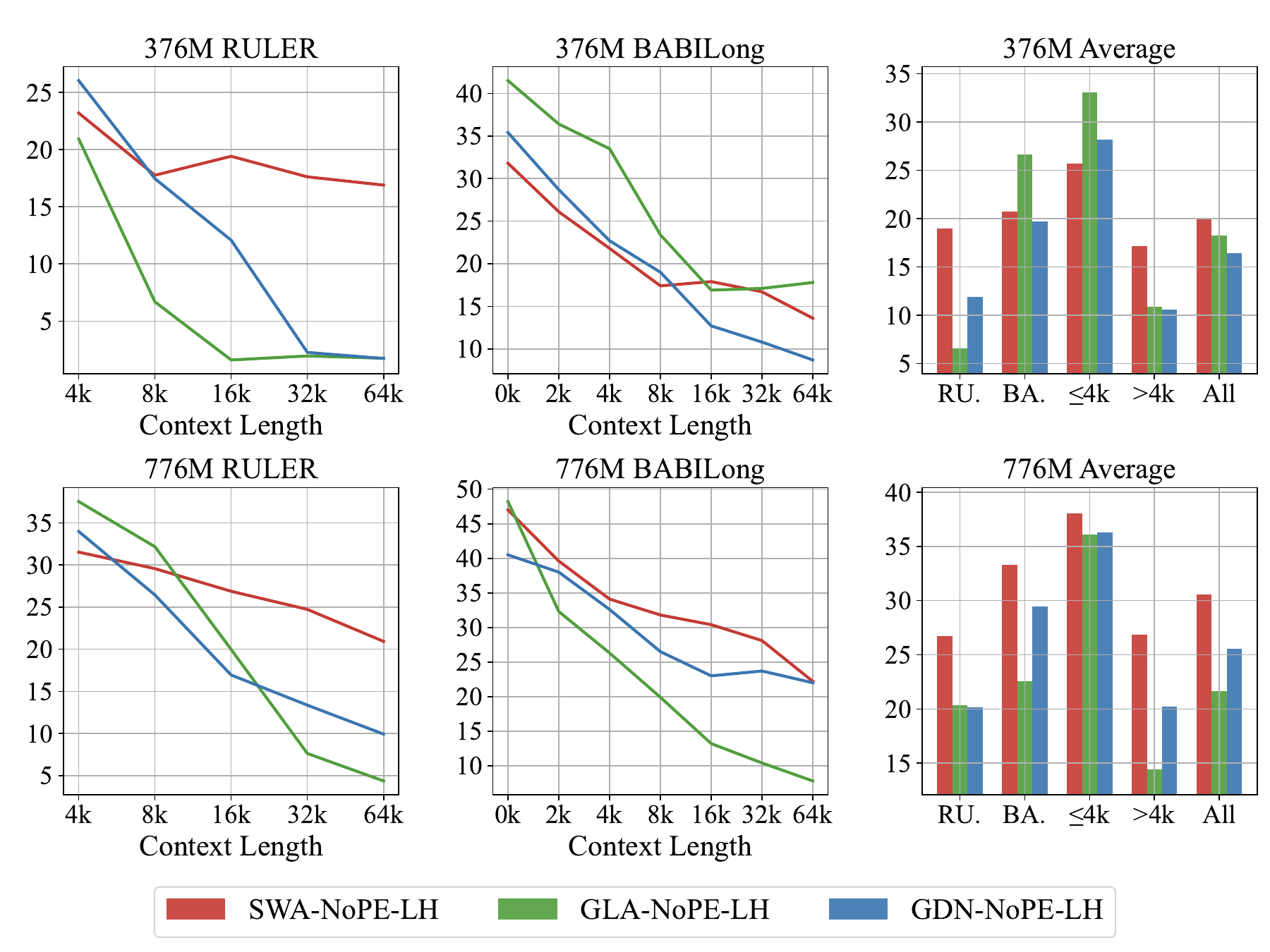}
    \caption{Performance of log-scale NoPE extrapolation.}
  \end{subfigure}
  \hfill
  \begin{subfigure}[b]{0.48\linewidth}
    \centering
    \includegraphics[width=\linewidth]{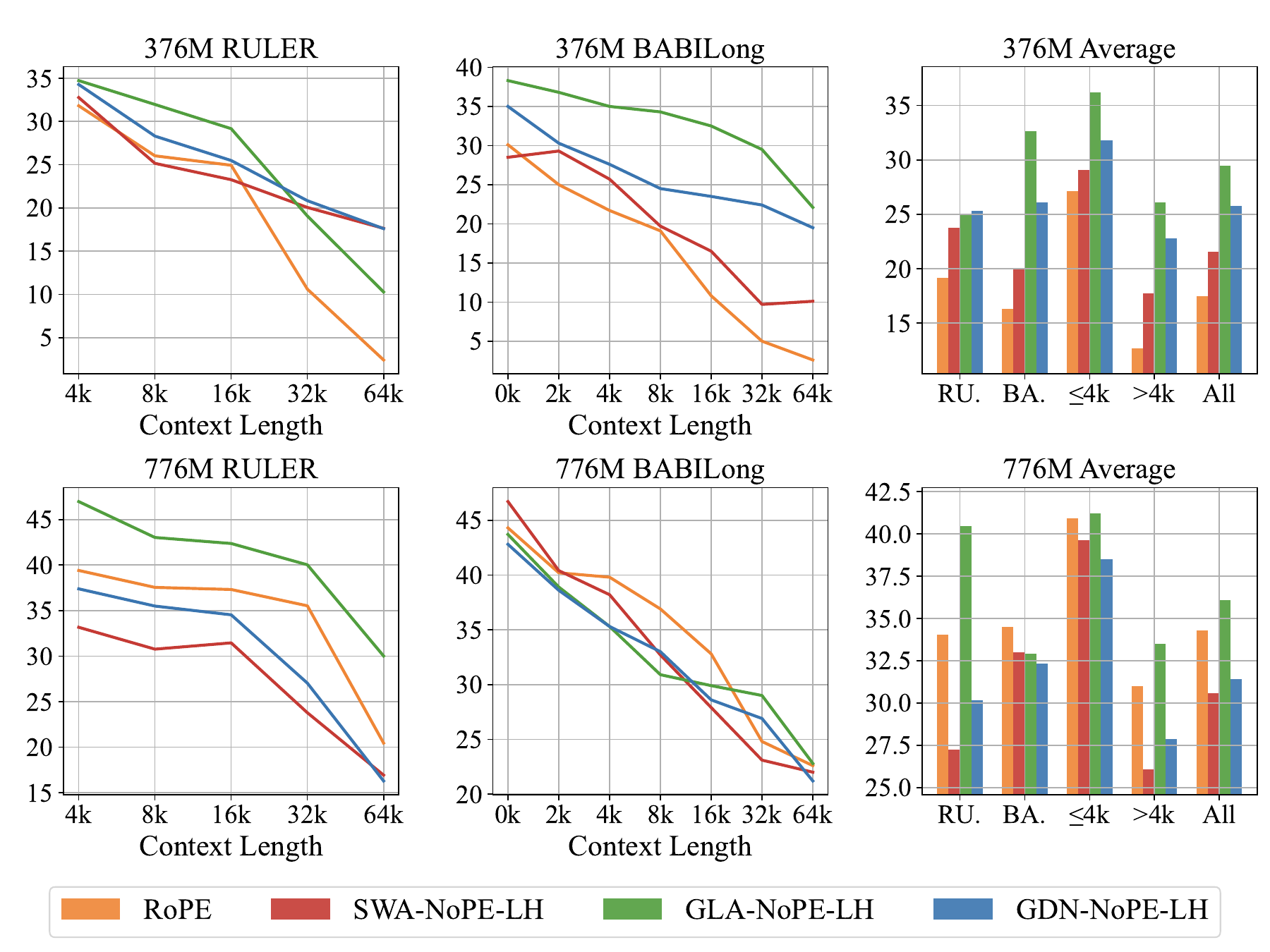}
    \caption{Performance after long-context pretraining.}
  \end{subfigure}
  \caption{Comparison of the long-context performance of the SWA/GLA/GDN-NoPE layer-wise hybrid models in YOCO-like layout under log-scale NoPE extrapolation and context extension. The Seesaw Effect still exists.}
  \label{fig-seesaw-yoco}
\end{figure}

To verify that the seesaw effect is stable, we first compare long-context continual pretraining results for SWA and LA hybrids at different peak learning rates, including 1e-4, 2e-4, 5e-4, and 1e-3, with 1e-3 as the default learning rate. We conduct this comparison at both 376M and 776M scales. We use the point with a zero learning rate to denote the direct length extrapolation result before long-context pretraining. As shown in Figure~\ref{fig-seesaw-verify}, the seesaw effect remains stable under different learning rates: after context extension, SWA-NoPE hybrids, especially on long-context tasks, consistently lag behind LA-NoPE hybrids. 

\begin{figure}[!t]
  \noindent
  \centering
  \begin{subfigure}[b]{0.48\linewidth}
    \centering
    \includegraphics[width=\linewidth]{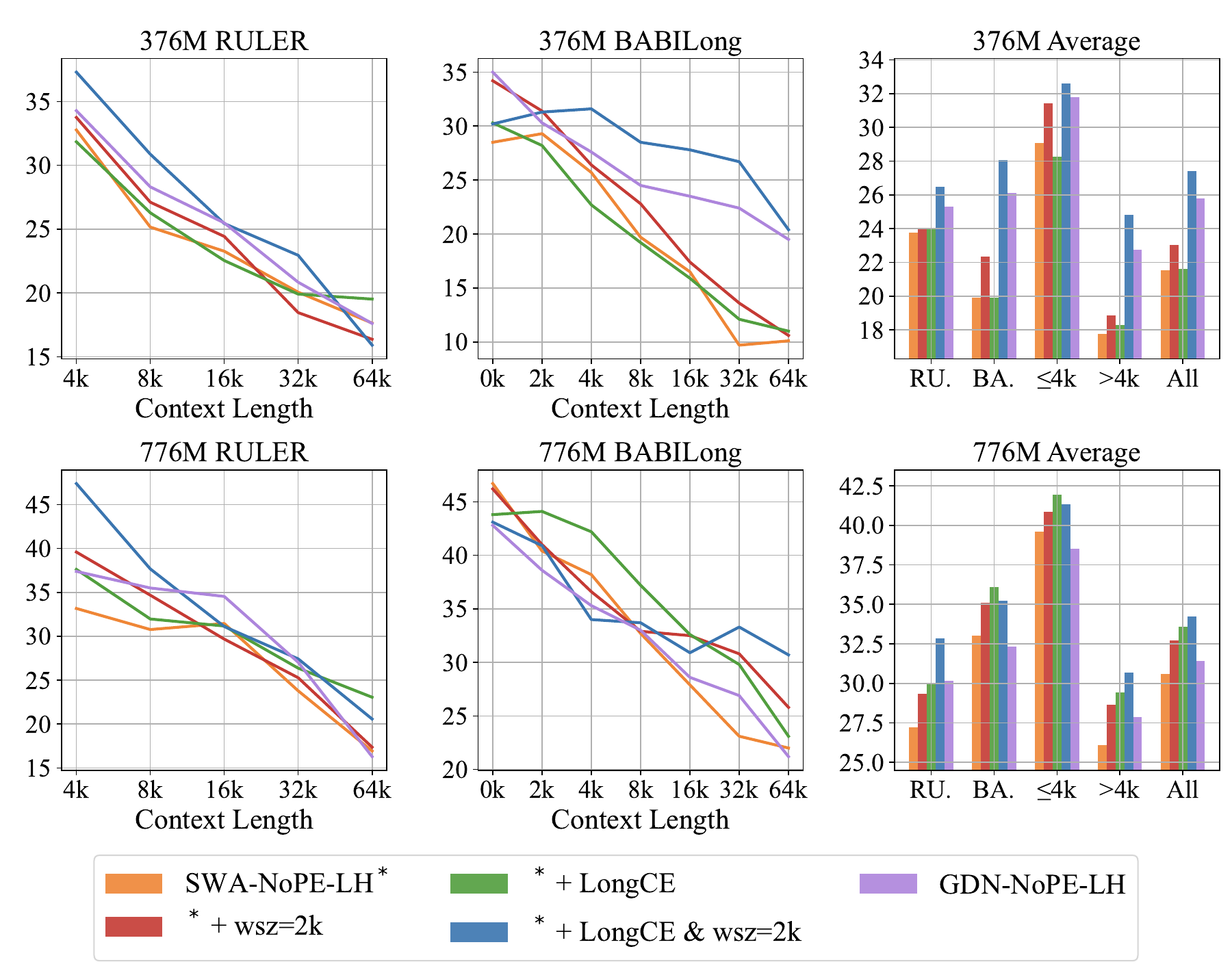}
    \caption{Performance of SWA-NoPE-LH context extension.}
  \end{subfigure}
  \hfill
  \begin{subfigure}[b]{0.48\linewidth}
    \centering
    \includegraphics[width=\linewidth]{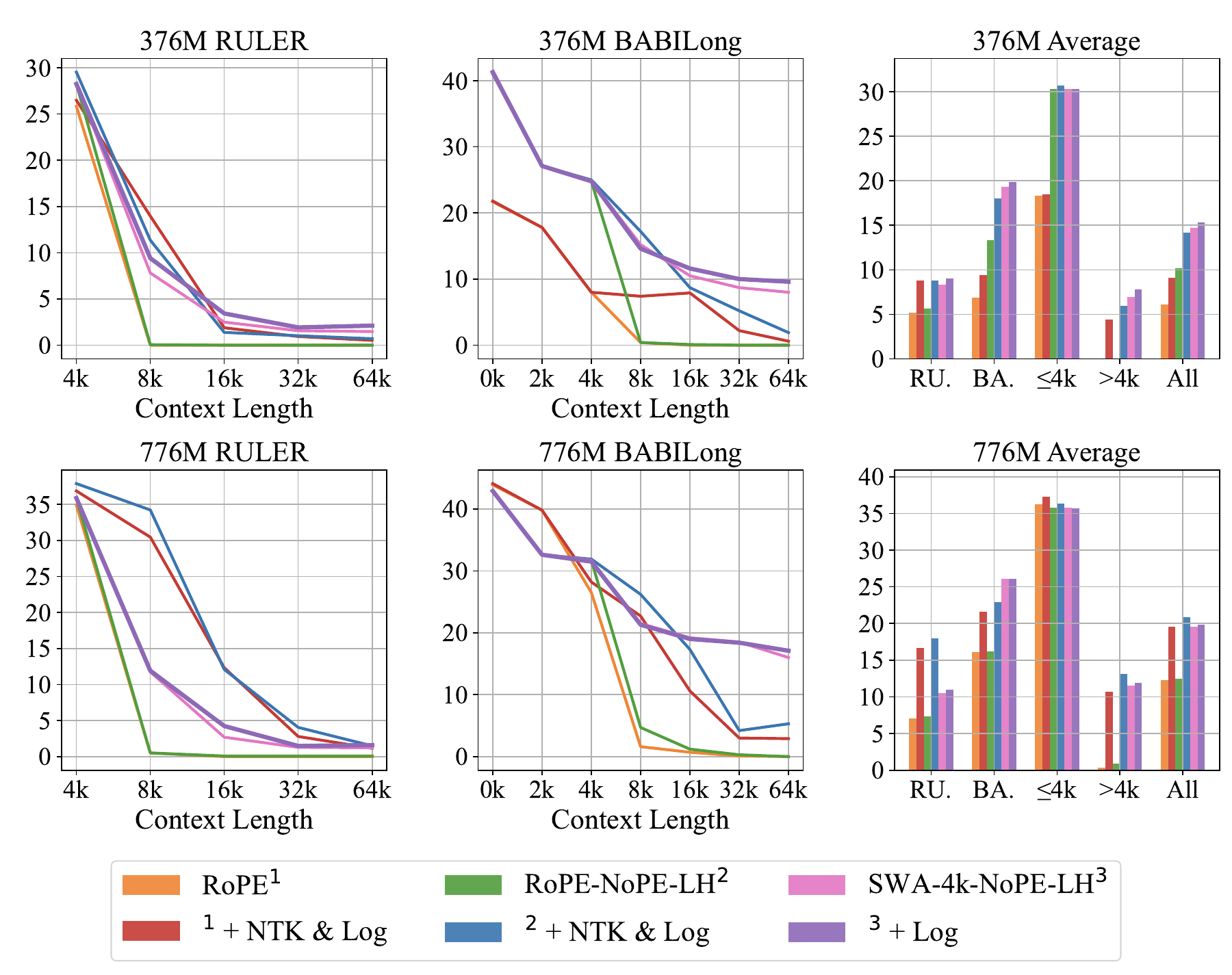}
    \caption{Performance of RoPE-NoPE-LH extrapolation.}
  \end{subfigure}
  \vskip0.6\baselineskip
  \begin{subfigure}[b]{0.48\linewidth}
    \centering
    \includegraphics[width=\linewidth]{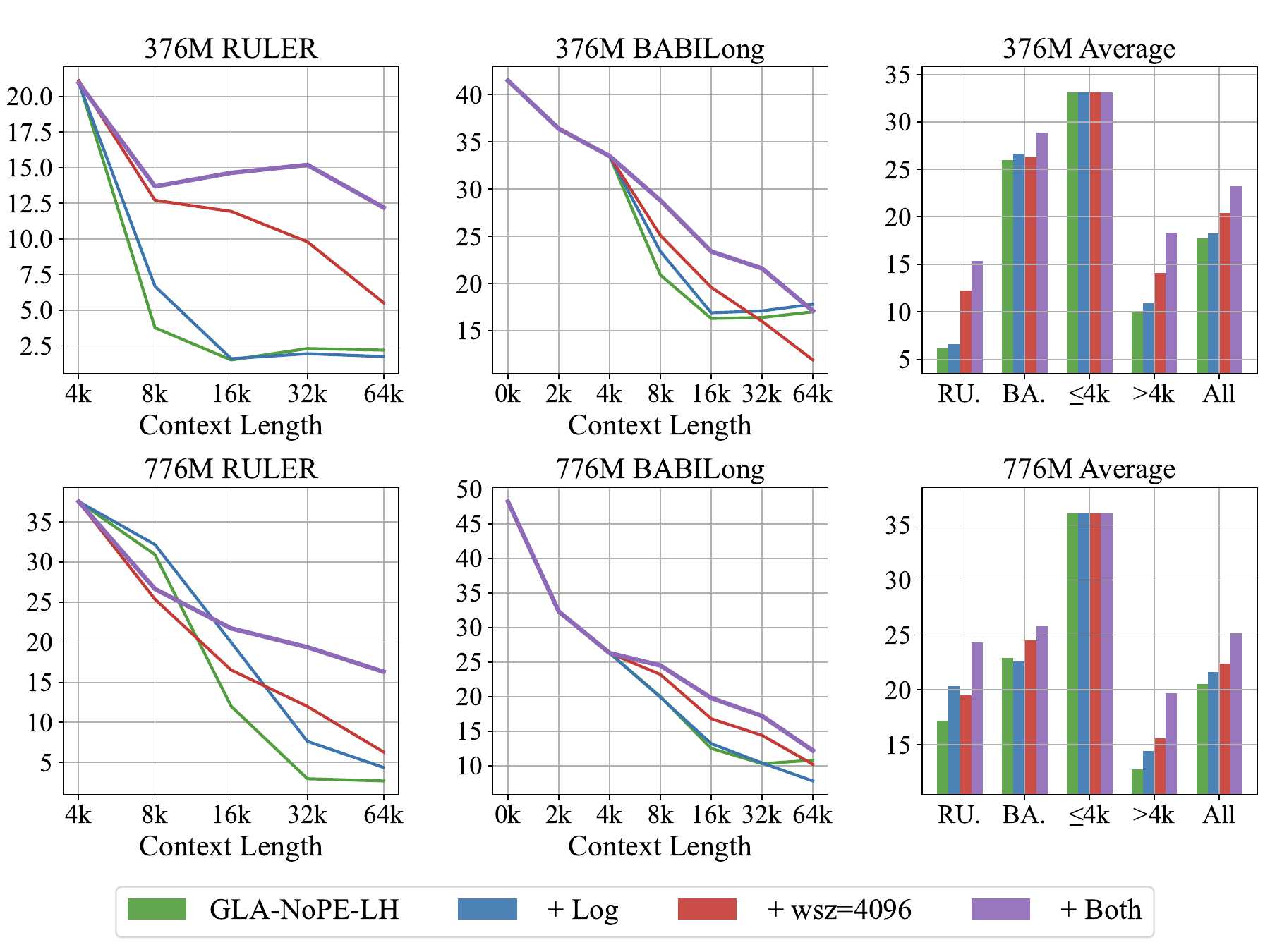}
    \caption{Performance of GLA-NoPE-LH extrapolation.}
  \end{subfigure}
  \hfill
  \begin{subfigure}[b]{0.48\linewidth}
    \centering
    \includegraphics[width=\linewidth]{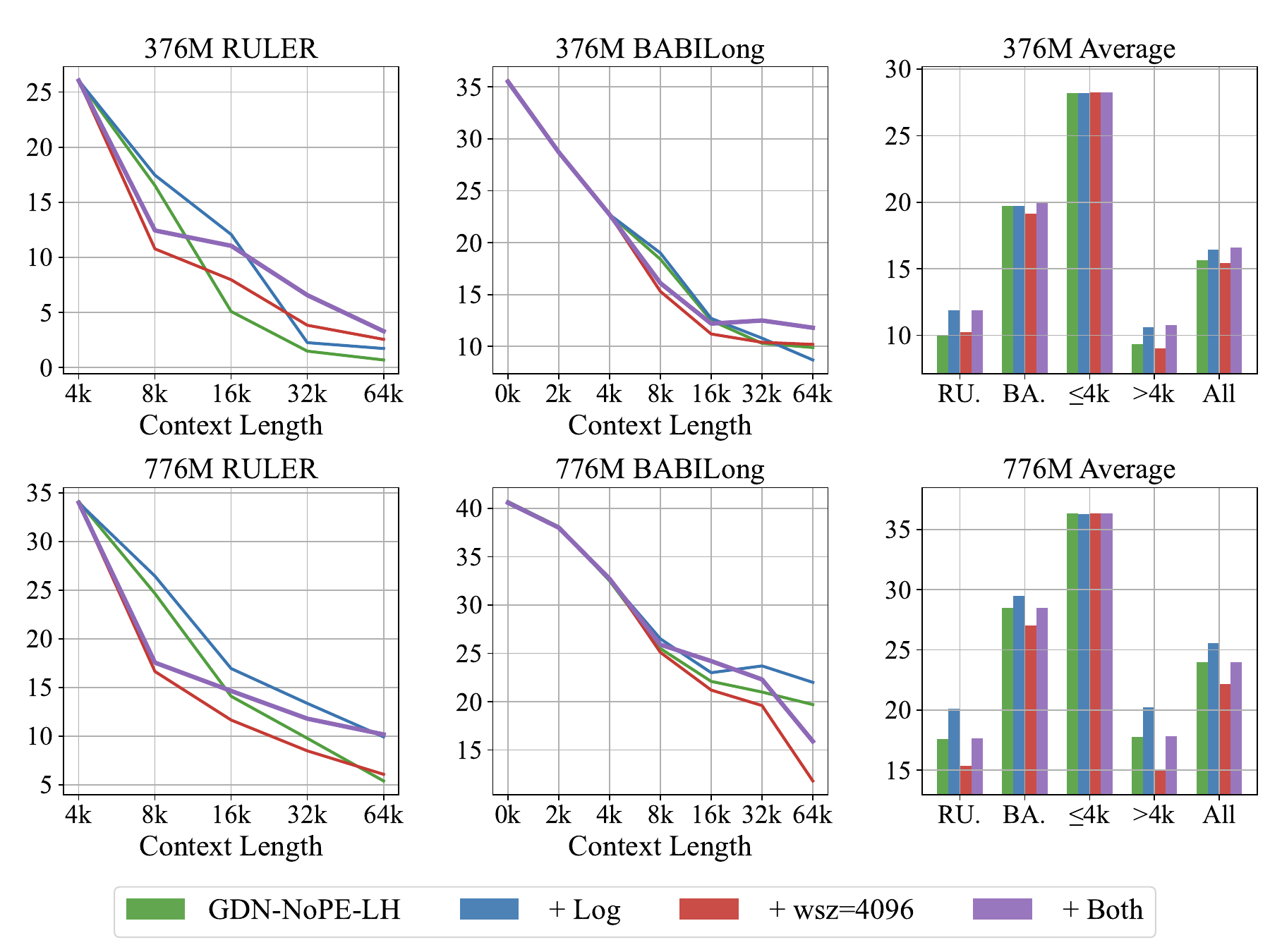}
    \caption{Performance of GDN-NoPE-LH extrapolation.}
  \end{subfigure}
  \caption{Performance of enhanced SWA-NoPE-LH context extension and RoPE/GLA/GDN-NoPE-LH extrapolation based on the Matthew Effect in layer-wise hybrid models in YOCO-like layout.}
  \label{fig-matthew-yoco}
\end{figure}

We also verify the existence of the Seesaw Effect in YOCO-like layer-wise hybrid models, in which all efficient-attention layers are placed at the lower layers. As shown in Figure~\ref{fig-seesaw-yoco}, the Seesaw Effect is a general phenomenon in SWA hybrids across different layer layouts. We further examine the enhanced context extension strategy based on enlarged window sizes, as well as LongCE and RoPE/GLA/GDN-NoPE extrapolation based on the Matthew Effect in layer-wise hybrid models in a YOCO-like layout, as shown in Figure~\ref{fig-matthew-yoco}. However, although the enhanced SWA-NoPE context extension still works, the RoPE/GLA/GDN-NoPE extrapolation based on the Matthew Effect becomes less effective. The model behavior may still differ across layer-wise hybrid layouts, requiring in-depth observation, analysis, and verification.

\subsection{Broader Discussion}\label{app-open}


\paragraph{Smaller Rotary Base in SWA Hybrids} We have compared short-context training results for SWA hybrids with different window sizes in short-context pretraining and long-context pretraining above. During window-size extension, the RoPE base must be adjusted accordingly. When the window size is increased from 128 to 256, 512, 1024, 2048, and 4096, the RoPE bases, calculated according to the scaling laws for RoPE extrapolation~\citep{liuscaling} and rounded up, become 90000, 700000, 6000000, 50000000, and 400000000, respectively. 

These values are clearly very large. This is because the initial window size is only 128, meaning that only a very limited number of dimensions have observed the complete positional information range; consequently, substantial interpolation is required. However, an excessively large RoPE base can reduce the discriminability of the position embedding. This motivated us to investigate the use of smaller RoPE bases in SWA. We kept the window size fixed at 128 and experimented with RoPE bases ranging from 5000 to 100 on both the 376M and 776M models. The results confirm that smaller bases can lead to better length extrapolation and context extension performance, as shown in Table~\ref{tab-ruler-swa-base}, but we do not observe a consistent trend across all settings. So we do not emphasize this finding in the main text. We will further investigate the underlying mechanism and try to develop an optimal scheduler for the training length, window size, and RoPE base in SWA hybrids.

\paragraph{Performance Difference within LA Hybrids} In the main text, we analyze LA hybrids by treating GLA and GDN as a whole. Nevertheless, there still exist differences within LA hybrids. For example, in short-context training, GLA hybrids are better at fitting but weaker at extrapolation than GDN hybrids, as shown in Figures~\ref{fig-main_ruler-lh}, \ref{fig-main_ruler-lh-long}, \ref{fig-main_ruler-hh} and \ref{fig-main_ruler-hh-long}. In contrast, GDN's fitting and extrapolation performance are more moderate, falling between GLA and SWA. However, when combined with SWLA, GLA hybrids show better length extrapolation performance than GDN hybrids, as shown in Figures~\ref{fig-matthew-gla} and \ref{fig-matthew-gdn}. Moreover, in the entropy-hit-rate diagram, GDN also occupies some part of the high-entropy hit-rate region while GLA does not, as shown in Figure~\ref{fig-scatter-all} and Figure~\ref{fig-scatter-la_hh}. These differences imply that there might exist a mechanism related to the introduction of the delta rule~\citep{yang2024parallelizing}, and we intend to conduct additional analysis in the future.

\paragraph{Transfer Training Experiment} Inspired by the effectiveness of hybrid position, we also attempt to transfer SWA/GLA/GDN-RoPE hybrids trained in 4k short contexts to corresponding SWA/GLA/GDN-NoPE in the long-context training phase, namely dropping RoPE in long-context training like DRoPE~\citep{gelberg2026extending,merrill2026olmo}. We report both short- and long-context evaluation results in Table~\ref{tab-short-main-lh-long}, \ref{tab-short-main-hh-long}, \ref{tab-ruler-lh-long}, \ref{tab-ruler-hh-long} and \ref{tab-ruler-yoco}, and find no consistent pattern across models or sizes. This area may still need further exploration.

\begin{figure}[!tb]
  \noindent
  \centering
  \begin{subfigure}[b]{0.49\linewidth}
    \centering
    \includegraphics[width=\linewidth]{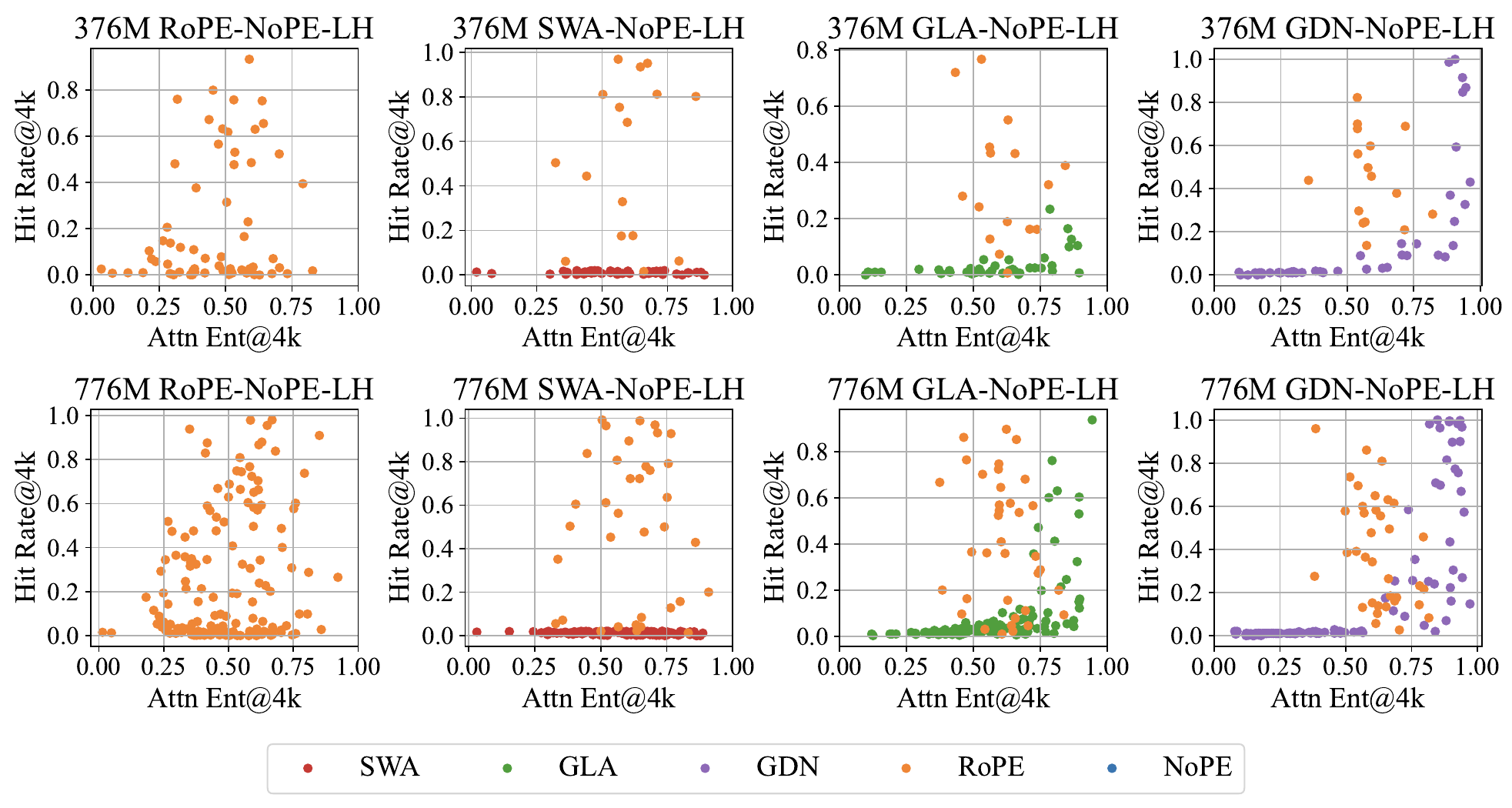}
    \caption{Comparison among RoPE-LH}
  \end{subfigure}
  \hfill
  \begin{subfigure}[b]{0.49\linewidth}
    \centering
    \includegraphics[width=\linewidth]{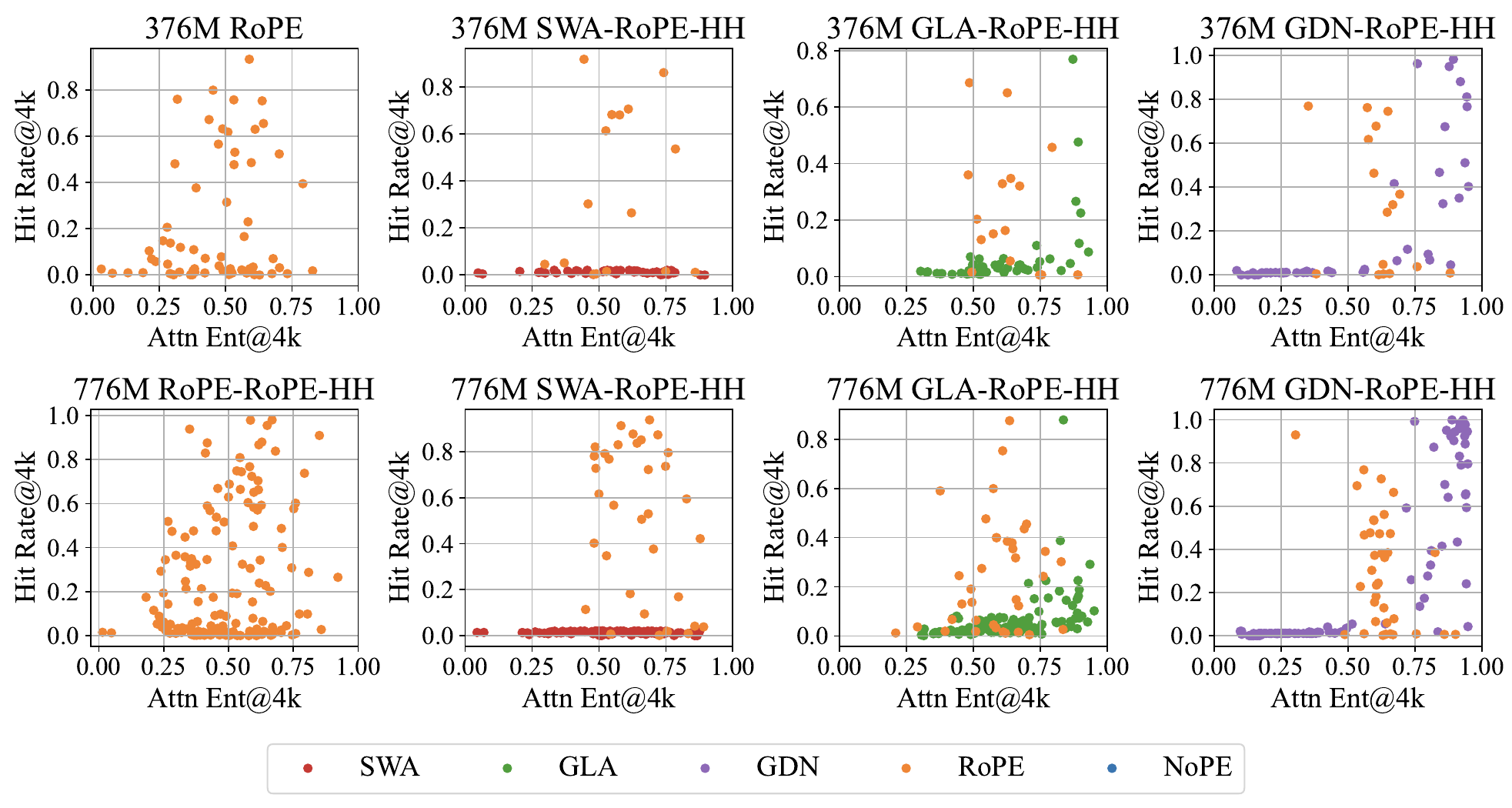}
    \caption{Comparison among RoPE-HH}
  \end{subfigure}
  \caption{The entropy-hit-rate diagram of RoPE-based full attention models and SWA/GLA/GDN-RoPE hybrid models.}
  \label{fig-scatter-rope}
\end{figure}

\paragraph{More Analysis of Hybrid Position} We further analyze the hybrid position in hybrid models that combine RoPE with other attention mechanisms. As shown in Figure~\ref{fig-scatter-rope}, in SWA/GLA/GDN-RoPE hybrid models, RoPE attention loses the lower-left region of the entropy–hit-rate diagram and instead retains the outward sector region. The low-entropy region at the bottom is mainly occupied by SWA or LA, resulting in a distribution pattern similar to that of NoPE attention. This observation motivates us to rethink the characteristic distribution of NoPE attention: is the band-like distribution extending from the upper-middle to the lower-right a property of softmax attention in hybrid models, or is it intrinsic to NoPE attention itself?

\begin{figure}[!tb]
  \noindent
  \centering
  \includegraphics[width=\linewidth]{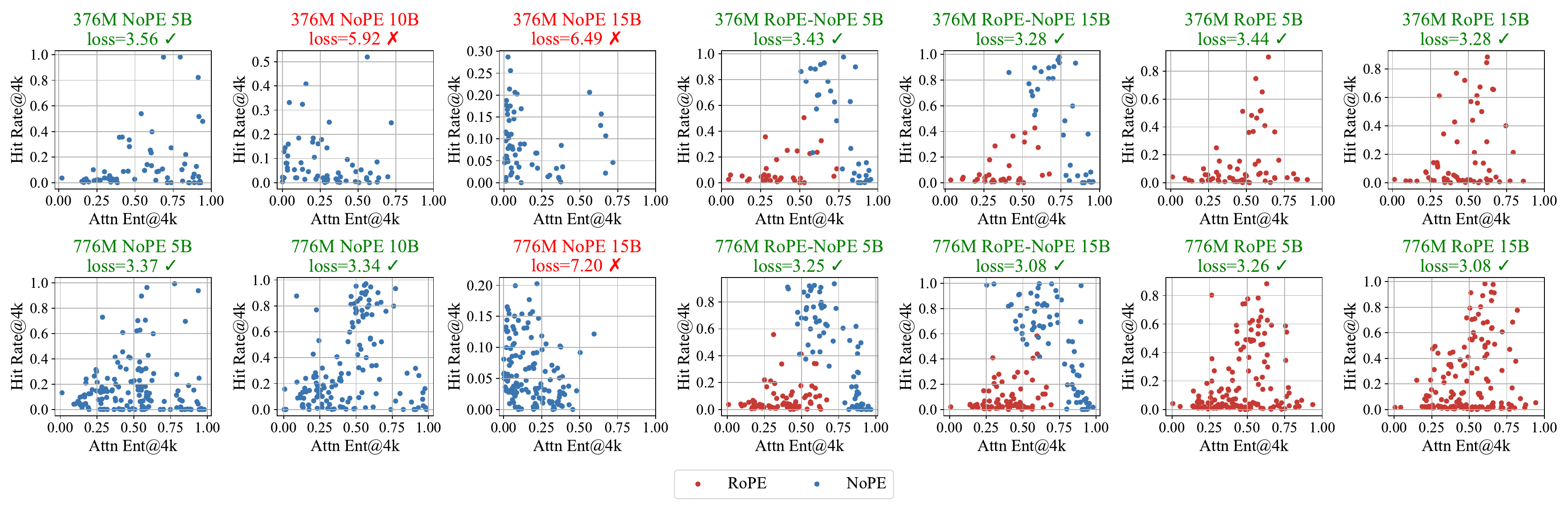}
  \caption{The entropy-hit-rate diagram of NoPE/RoPE-based full attention models and RoPE-NoPE-HH models.}
  \label{fig-scatter-nope}
\end{figure}

Unfortunately, because NoPE-only models have inherent limitations, we are currently unable to answer this question. As shown in Figure~\ref{fig-scatter-nope}, at the beginning of training, NoPE attentions try to develop a trend that expands outward from the lower-left region of the entropy-hit-rate diagram, similar to RoPE attentions. However, as the loss fluctuates remarkably during training, this distribution pattern also collapses. In contrast, in standard RoPE models and RoPE-NoPE hybrid models, the features analyzed in the main text emerge after only a relatively small number of training tokens. Given the observed behavior of the NoPE-RoPE hybrid model and the instability of the NoPE-only model, we still regard the band-like distribution as a property of NoPE-based attention in the hybrid settings, rather than as a universal property of all softmax attention.

\section{Detailed Evaluation Results}\label{app-detail}

In this section, we first report the results of short-context evaluation. We evaluate both short-context and long-context models on downstream tasks mainly in Open LLM Leaderboard~\citep{openllm}, including TruthfulQA~\citep{DBLP:conf/acl/LinHE22}, LAMBADA\citep{DBLP:conf/acl/PapernoKLPBPBBF16}, PIQA~\citep{DBLP:conf/aaai/BiskZLGC20}, HellaSwag~\citep{DBLP:conf/acl/ZellersHBFC19}, WinoGrande~\citep{DBLP:conf/aaai/SakaguchiBBC20}, ARC~\citep{DBLP:journals/corr/abs-1803-05457}, GPQA~\citep{DBLP:journals/corr/abs-2311-12022}, SocialIQA~\citep{DBLP:conf/emnlp/SapRCBC19}, OpenBookQA~\citep{DBLP:conf/emnlp/MihaylovCKS18}, SuperGLUE~\citep{DBLP:conf/nips/WangPNSMHLB19}, and MMLU~\citep{DBLP:conf/iclr/HendrycksBBZMSS21}. All models are tested within a 2k context length. The results are reported in Tables~\ref{tab-short-main-lh-short}, \ref{tab-short-main-lh-long}, \ref{tab-short-main-hh-short} and \ref{tab-short-main-hh-long}. Hybrid models achieve performance comparable to the traditional RoPE-based full-attention-only model, before and after long-context continual pretraining.

Regarding the experiments listed in the main text, we report the long-context performance of hybrid models after short-context pretraining in Tables~\ref{tab-ruler-lh-short-small}, \ref{tab-ruler-lh-short-large}, \ref{tab-ruler-hh-short-small} and \ref{tab-ruler-hh-short-large}. The length extrapolation performance verifies the effectiveness of the Matthew Effect in extrapolation and the sliding-window linear attention. We also report long-context performance of hybrid models after long-context continual pretraining in Tables~\ref{tab-ruler-lh-long} and \ref{tab-ruler-hh-long}. The context extension performance validates the Seesaw Effect of SWA Hybrids, while the length extrapolation performance validates the No-Free-Lunch Effect of LA Hybrids. We further detail the long-context performance of SWA layer-wise hybrid models with different window sizes in short-context pretraining and long-context continual pretraining in Tables~\ref{tab-ruler-swa-sww} and \ref{tab-ruler-swa-lwl}, validating the Short-Window Weariness and Long-Window Laziness. Besides, in Table~\ref{tab-ruler-swa-longce-lh}, we also report the long-context performance of SWA-LH after long-context continual pretraining with extended window sizes (wsz) and LongCE enhancement, both of which effectively relieve the short-context learning trap. We detail more NIAH results in the RULER evaluation in Table~\ref{tab-niah-matthew-la-all} to verify the effectiveness of sliding-window linear attention, and report the long-context performance of RoPE/SWA/GLA/GDN-NoPE-HH with different hybrid ratios in Table~\ref{tab-ruler-hybrid-hh-short}.

Regarding the experiments listed in the Appendix, we also include detailed LongBench scores in Tables~\ref{tab-longbench-lh-short-small}, \ref{tab-longbench-lh-short-large}, \ref{tab-longbench-hh-short-small}, \ref{tab-longbench-hh-short-large}, \ref{tab-longbench-lh-long} and \ref{tab-longbench-hh-long}, verifying the Seesaw Effect and Matthew Effect in different evaluation benchmarks. Additionally, in Table~\ref{tab-ruler-yoco}, we verify the Seesaw Effect under different layer-wise hybrid layouts. Finally, in Table~\ref{tab-ruler-swa-base}, we report the long-context performance of SWA-NoPE-LH with different rotary bases in SWA. 

We provide these detailed results at the end of the paper to convey the excitement of discovering patterns through careful observation. Although the tables may appear tedious, reliable conclusions emerge only from momentary observations followed by repeated verification, just like Oersted’s accidental discovery during a classroom demonstration that an electric current in a straight wire could deflect a magnetic needle. Through subsequent verification and systematic induction, this observation became an important starting point for the development of electromagnetism. Although our mechanics of long-context hybrid models are still at an early stage, we believe that continued observation will reveal more of their underlying principles. We also hope that readers may discover, from these results, conclusions that we have not yet recognized.

\begin{table}[!ht]
\small
\tabcolsep=0.12cm
\centering
\begin{tabular}{l ccccccccccccc}
\toprule 
& \textbf{TQA} & \textbf{LMD} & \textbf{PIQA} & \textbf{Hella} & \textbf{Wino} & \textbf{ARC-e} & \textbf{ARC-c} & \textbf{GPQA} & \textbf{SIQA} & \textbf{OBQA} & \textbf{SG} & \textbf{MMLU} & \textbf{Avg.} \\
\midrule
\multicolumn{14}{l}{\textbf{\textit{376M Short}}} \\[0.2ex]
RoPE & 36.8 & 40.2 & 65.9 & 33.8 & 52.2 & \textbf{39.7} & 26.4 & 26.3 & 39.2 & 27.4 & 42.1 & 24.6 & 37.9 \\ 
RoPE-NoPE-LH & 36.4 & 41.7 & 65.9 & 33.8 & 51.1 & 39.2 & \textbf{27.5} & \textbf{28.8} & 40.3 & 25.8 & \textbf{45.2} & \textbf{26.9} & \textbf{38.5} \\ 
\hdashline
SWA-RoPE-LH & 35.8 & 42.6 & 65.8 & 33.4 & 52.3 & 37.0 & 25.4 & 24.8 & 39.9 & \textbf{27.6} & 44.8 & 25.9 & 37.9 \\ 
SWA-NoPE-LH & 36.4 & 42.0 & 66.4 & 33.4 & 50.2 & 37.2 & 24.1 & 25.8 & 39.7 & 21.6 & 43.6 & 25.6 & 37.2 \\ 
\hdashline
GLA-RoPE-LH & \textbf{37.0} & 42.7 & 66.4 & 33.8 & 52.3 & 38.8 & 25.1 & 24.2 & 39.0 & 26.6 & 44.7 & 26.3 & 38.1 \\ 
GLA-NoPE-LH & 36.0 & 44.2 & 65.9 & 34.1 & 52.4 & \textbf{39.7} & 23.4 & 23.7 & 38.9 & 27.4 & 44.9 & 24.9 & 38.0 \\
\bottomrule
\end{tabular}
\caption{Short-context performance of layer-wise hybrid models after short-context pretraining.\label{tab-short-main-lh-short}}
\end{table}

\begin{table}[!ht]
\small
\tabcolsep=0.12cm
\centering
\begin{tabular}{l ccccccccccccc}
\toprule 
& \textbf{TQA} & \textbf{LMD} & \textbf{PIQA} & \textbf{Hella} & \textbf{Wino} & \textbf{ARC-e} & \textbf{ARC-c} & \textbf{GPQA} & \textbf{SIQA} & \textbf{OBQA} & \textbf{SG} & \textbf{MMLU} & \textbf{Avg.} \\
\midrule
\multicolumn{14}{l}{\textbf{\textit{376M Short}}} \\[0.2ex]
RoPE & 36.8 & 40.2 & 65.9 & 33.8 & 52.2 & 39.7 & \textbf{26.4} & 26.3 & 39.2 & 27.4 & 42.1 & 24.6 & 37.9 \\ 
RoPE-NoPE-HH & \textbf{37.1} & 40.1 & 66.0 & 33.3 & 50.4 & 40.0 & 24.8 & 27.8 & 39.9 & 27.8 & 43.3 & 25.8 & 38.0 \\ 
\hdashline
SWA-RoPE-HH & 35.1 & 41.3 & 65.4 & 33.6 & 51.3 & 39.7 & 23.7 & 22.7 & 38.8 & \textbf{28.0} & 45.0 & 26.3 & 37.6 \\ 
SWA-NoPE-HH & 35.8 & 41.7 & 65.5 & 33.8 & 51.7 & 39.0 & \textbf{26.4} & 27.8 & 38.7 & 23.4 & 42.2 & 25.2 & 37.6 \\ 
\hdashline
GLA-RoPE-HH & 36.0 & \textbf{46.0} & 66.0 & \textbf{35.0} & 52.4 & \textbf{40.2} & 24.8 & 25.8 & 39.1 & 23.6 & 43.5 & 25.6 & 38.2 \\ 
GLA-NoPE-HH & 36.0 & 45.2 & 65.7 & 34.6 & 51.5 & 37.7 & 25.8 & 27.8 & 39.6 & 27.8 & 44.7 & 26.2 & 38.5 \\
\bottomrule
\end{tabular}
\caption{Short-context performance of head-wise hybrid models after short-context pretraining.\label{tab-short-main-hh-short}}
\end{table}

\begin{table}[!ht]
\small
\tabcolsep=0.12cm
\centering
\begin{tabular}{l ccccccccccccc}
\toprule 
& \textbf{TQA} & \textbf{LMD} & \textbf{PIQA} & \textbf{Hella} & \textbf{Wino} & \textbf{ARC-e} & \textbf{ARC-c} & \textbf{GPQA} & \textbf{SIQA} & \textbf{OBQA} & \textbf{SG} & \textbf{MMLU} & \textbf{Avg.} \\
\midrule
\multicolumn{14}{l}{\textbf{\textit{376M Long}}} \\[0.2ex]
RoPE & 35.2 & 35.8 & 64.3 & 31.2 & 51.2 & 36.7 & 25.8 & 24.8 & 38.5 & 28.0 & 43.6 & 24.2 & 36.6 \\ 
RoPE-NoPE-LH & 36.0 & 38.7 & 64.2 & 31.7 & 52.5 & 37.6 & \textbf{28.8} & 25.8 & \textbf{39.2} & \textbf{28.2} & 42.9 & 24.7 & 37.5 \\ 
\hdashline
SWA-RoPE-LH & 35.4 & 39.3 & 64.2 & 31.0 & 52.4 & 39.2 & 25.8 & 23.7 & 38.3 & 28.0 & \textbf{45.3} & 25.2 & 37.3 \\ 
+ DRoPE & 35.5 & 39.1 & \textbf{65.0} & 30.8 & 51.6 & 39.0 & 23.4 & 24.8 & 38.1 & 27.6 & 44.7 & 25.7 & 37.1 \\ 
SWA-NoPE-LH & 35.8 & 39.6 & 64.3 & 31.4 & 49.0 & 36.7 & 25.4 & 25.3 & 37.6 & 26.4 & 44.4 & 25.0 & 36.7 \\ 
\hdashline
GLA-RoPE-LH & 37.1 & 41.6 & 64.4 & 32.0 & 52.2 & 37.0 & 24.1 & 25.3 & 39.1 & 27.2 & 44.9 & \textbf{26.2} & 37.6 \\
\bottomrule
\end{tabular}
\caption{Short-context performance of layer-wise hybrid models after long-context continual pretraining.\label{tab-short-main-lh-long}}
\end{table}

\begin{table}[!ht]
\small
\tabcolsep=0.12cm
\centering
\begin{tabular}{l ccccccccccccc}
\toprule 
& \textbf{TQA} & \textbf{LMD} & \textbf{PIQA} & \textbf{Hella} & \textbf{Wino} & \textbf{ARC-e} & \textbf{ARC-c} & \textbf{GPQA} & \textbf{SIQA} & \textbf{OBQA} & \textbf{SG} & \textbf{MMLU} & \textbf{Avg.} \\
\midrule
\multicolumn{14}{l}{\textbf{\textit{376M Long}}} \\[0.2ex]
RoPE & 35.2 & 35.8 & 64.3 & 31.2 & 51.2 & 36.7 & 25.8 & 24.8 & 38.5 & 28.0 & 43.6 & 24.2 & 36.6 \\ 
RoPE-NoPE-HH & \textbf{37.9} & 36.4 & 63.6 & 31.3 & 50.2 & 38.5 & \textbf{27.1} & \textbf{26.3} & 37.8 & 28.0 & 44.1 & \textbf{26.1} & 37.3 \\ 
\hdashline
SWA-RoPE-HH & 35.2 & 37.6 & 63.3 & 31.4 & 51.0 & 37.0 & 22.4 & 24.8 & 37.9 & 28.0 & \textbf{45.1} & 25.1 & 36.6 \\ 
+ DRoPE & 35.1 & 37.1 & 63.3 & 31.3 & 52.0 & 38.1 & 23.4 & 24.8 & 37.2 & 27.8 & \textbf{45.1} & 25.2 & 36.7 \\ 
SWA-NoPE-HH & 35.5 & 36.9 & 63.8 & 31.5 & 51.7 & 37.2 & 26.1 & 25.8 & 37.9 & 27.8 & 43.8 & 23.5 & 36.8 \\ 
\hdashline
GLA-RoPE-HH & 37.0 & \textbf{43.0} & \textbf{65.4} & 32.4 & 51.5 & \textbf{39.2} & 24.4 & 25.8 & 38.2 & 24.6 & 43.9 & 25.4 & \textbf{37.6} \\
\bottomrule
\end{tabular}
\caption{Short-context performance of head-wise hybrid models after long-context continual pretraining.\label{tab-short-main-hh-long}}
\end{table}

\begin{table}[!ht]
\small
\tabcolsep=0.12cm
\centering
\begin{tabular}{l ccccc ccccccc ccccc}
\toprule 
& \multicolumn{5}{c}{\textbf{RULER}} & \multicolumn{7}{c}{\textbf{BABILong}} & \multicolumn{5}{c}{\textbf{Average}} \\
\cmidrule(lr){2-6} \cmidrule(lr){7-13} \cmidrule(lr){14-18}
 & 4k & 8k & 16k & 32k & 64k & 0k & 2k & 4k & 8k & 16k & 32k & 64k & RU. & BA & $\leq$4k & >4k & All \\
\midrule
\multicolumn{18}{l}{\textbf{\textit{376M Short}}} \\[0.2ex]
RoPE & 25.8 & 0.0 & 0.0 & 0.0 & 0.0 & 21.7 & 17.8 & 8.0 & 0.4 & 0.0 & 0.0 & 0.0 & 5.2 & 6.8 & 18.3 & 0.1 & 6.1 \\ 
+ NTK & 26.5 & 9.9 & 0.6 & 0.1 & 0.0 & 21.8 & 17.8 & 7.8 & 8.9 & 2.2 & 0.6 & 0.0 & 7.4 & 8.4 & 18.5 & 2.8 & 8.0 \\ 
+ NTK + Log & 26.5 & 14.0 & 1.9 & 0.9 & 0.5 & 21.8 & 17.8 & 8.0 & 7.4 & 7.9 & 2.2 & 0.6 & 8.8 & 9.4 & 18.5 & 4.4 & 9.1 \\ 
RoPE-NoPE-LH & 29.9 & 0.1 & 0.0 & 0.0 & 0.0 & 32.9 & 19.5 & 15.7 & 0.6 & 0.2 & 0.0 & 0.0 & 6.0 & 9.8 & 24.5 & 0.1 & 8.2 \\ 
+ NTK & \textbf{30.6} & 12.1 & 1.3 & 0.7 & 0.4 & 33.0 & 19.5 & 15.9 & 23.2 & 4.9 & 0.7 & 0.0 & 9.0 & 13.9 & 24.7 & 5.4 & 11.9 \\
\bottomrule
\end{tabular}
\caption{Long-context performance of 376M and 776M layer-wise hybrid models after short-context pretraining.\label{tab-ruler-lh-short-small}}
\end{table}

\begin{table}[!ht]
\small
\tabcolsep=0.12cm
\centering
\begin{tabular}{l ccccc ccccccc ccccc}
\toprule 
& \multicolumn{5}{c}{\textbf{RULER}} & \multicolumn{7}{c}{\textbf{BABILong}} & \multicolumn{5}{c}{\textbf{Average}} \\
\cmidrule(lr){2-6} \cmidrule(lr){7-13} \cmidrule(lr){14-18}
 & 4k & 8k & 16k & 32k & 64k & 0k & 2k & 4k & 8k & 16k & 32k & 64k & RU. & BA & $\leq$4k & >4k & All \\
\midrule
\multicolumn{18}{l}{\textbf{\textit{1B Short}}} \\[0.2ex]
RoPE & 37.3 & 1.5 & 0.0 & 0.0 & 0.0 & 47.4 & 41.7 & 28.8 & 1.7 & 0.3 & 0.2 & 0.1 & 7.8 & 17.2 & 38.8 & 0.5 & 13.3 \\ 
+ NTK & 39.3 & 29.7 & 6.9 & 0.5 & 0.7 & 47.5 & 41.6 & 28.8 & 21.5 & 9.0 & 7.0 & 2.0 & 15.4 & 22.5 & 39.3 & 9.7 & 19.5 \\ 
+ NTK + Log & 39.3 & 33.4 & 18.7 & 8.6 & 5.8 & 47.4 & 41.7 & 28.8 & 24.7 & 20.0 & 13.4 & 9.5 & 21.2 & 26.5 & 39.3 & 16.8 & 24.3 \\ 
RoPE-NoPE-LH & 46.9 & 0.1 & 0.2 & 0.0 & 0.0 & 65.2 & \textbf{51.9} & \textbf{48.1} & 2.5 & 0.1 & 0.0 & 0.0 & 9.4 & 24.0 & 53.0 & 0.4 & 17.9 \\ 
+ NTK & 45.8 & 35.9 & 6.9 & 0.3 & 0.2 & 65.1 & 51.7 & 47.0 & 39.3 & 13.3 & 4.2 & 0.2 & 17.8 & 31.5 & 52.4 & 12.5 & 25.8 \\
\bottomrule
\end{tabular}
\caption{Long-context performance of 1B and 3B layer-wise hybrid models after short-context pretraining.\label{tab-ruler-lh-short-large}}
\end{table}

\begin{table}[!ht]
\small
\tabcolsep=0.12cm
\centering
\begin{tabular}{l ccccc ccccccc ccccc}
\toprule 
& \multicolumn{5}{c}{\textbf{RULER}} & \multicolumn{7}{c}{\textbf{BABILong}} & \multicolumn{5}{c}{\textbf{Average}} \\
\cmidrule(lr){2-6} \cmidrule(lr){7-13} \cmidrule(lr){14-18}
 & 4k & 8k & 16k & 32k & 64k & 0k & 2k & 4k & 8k & 16k & 32k & 64k & RU. & BA & $\leq$4k & >4k & All \\
\midrule
\multicolumn{18}{l}{\textbf{\textit{376M Short}}} \\[0.2ex]
RoPE & 25.8 & 0.0 & 0.0 & 0.0 & 0.0 & 21.7 & 17.8 & 8.0 & 0.4 & 0.0 & 0.0 & 0.0 & 5.2 & 6.8 & 18.3 & 0.1 & 6.1 \\ 
+ NTK & 26.5 & 9.9 & 0.6 & 0.1 & 0.0 & 21.8 & 17.8 & 7.8 & 8.9 & 2.2 & 0.6 & 0.0 & 7.4 & 8.4 & 18.5 & 2.8 & 8.0 \\ 
+ NTK + Log & 26.5 & 14.0 & 1.9 & 0.9 & 0.5 & 21.8 & 17.8 & 8.0 & 7.4 & 7.9 & 2.2 & 0.6 & 8.8 & 9.4 & 18.5 & 4.4 & 9.1 \\ 
RoPE-NoPE-HH & 33.1 & 0.3 & 0.0 & 0.0 & 0.0 & 27.1 & 28.4 & 27.0 & 0.3 & 0.0 & 0.0 & 0.0 & 6.7 & 11.8 & 28.9 & 0.1 & 9.7 \\ 
+ NTK & 33.2 & 14.3 & 0.6 & 0.2 & 0.1 & 27.3 & 28.4 & 27.0 & 24.6 & 2.9 & 1.4 & 0.1 & 9.7 & 16.0 & 29.0 & 5.5 & 13.3 \\
\bottomrule
\end{tabular}
\caption{Long-context performance of 376M and 776M head-wise hybrid models after short-context pretraining.\label{tab-ruler-hh-short-small}}
\end{table}

\begin{table}[!ht]
\small
\tabcolsep=0.12cm
\centering
\begin{tabular}{l ccccc ccccccc ccccc}
\toprule 
& \multicolumn{5}{c}{\textbf{RULER}} & \multicolumn{7}{c}{\textbf{BABILong}} & \multicolumn{5}{c}{\textbf{Average}} \\
\cmidrule(lr){2-6} \cmidrule(lr){7-13} \cmidrule(lr){14-18}
 & 4k & 8k & 16k & 32k & 64k & 0k & 2k & 4k & 8k & 16k & 32k & 64k & RU. & BA & $\leq$4k & >4k & All \\
\midrule
\multicolumn{18}{l}{\textbf{\textit{1B Short}}} \\[0.2ex]
RoPE & 37.3 & 1.5 & 0.0 & 0.0 & 0.0 & 47.4 & 41.7 & 28.8 & 1.7 & 0.3 & 0.2 & 0.1 & 7.8 & 17.2 & 38.8 & 0.5 & 13.3 \\ 
+ NTK & 39.3 & 29.7 & 6.9 & 0.5 & 0.7 & 47.5 & 41.6 & 28.8 & 21.5 & 9.0 & 7.0 & 2.0 & 15.4 & 22.5 & 39.3 & 9.7 & 19.5 \\ 
+ NTK + Log & 39.3 & 33.4 & 18.7 & 8.6 & 5.8 & 47.4 & 41.7 & 28.8 & 24.7 & 20.0 & 13.4 & 9.5 & 21.2 & 26.5 & 39.3 & 16.8 & 24.3 \\ 
RoPE-NoPE-HH & 44.4 & 1.0 & 0.1 & 0.0 & 0.0 & \textbf{54.9} & 39.7 & 35.9 & 11.4 & 0.0 & 0.0 & 0.0 & 9.1 & 20.3 & 43.7 & 1.6 & 15.6 \\ 
+ NTK & 43.9 & 33.7 & 10.1 & 1.5 & 0.8 & 54.8 & 39.6 & 35.6 & 32.4 & 13.2 & 3.7 & 0.0 & 18.0 & 25.6 & 43.5 & 11.9 & 22.5 \\
\bottomrule
\end{tabular}
\caption{Long-context performance of 1B and 3B head-wise hybrid models after short-context pretraining.\label{tab-ruler-hh-short-large}}
\end{table}

\begin{table}[!ht]
\small
\tabcolsep=0.12cm
\centering
\begin{tabular}{l ccccc ccccccc ccccc}
\toprule 
& \multicolumn{5}{c}{\textbf{RULER}} & \multicolumn{7}{c}{\textbf{BABILong}} & \multicolumn{5}{c}{\textbf{Average}} \\
\cmidrule(lr){2-6} \cmidrule(lr){7-13} \cmidrule(lr){14-18}
 & 4k & 8k & 16k & 32k & 64k & 0k & 2k & 4k & 8k & 16k & 32k & 64k & RU. & BA & $\leq$4k & >4k & All \\
\midrule
\multicolumn{18}{l}{\textbf{\textit{376M Long}}} \\[0.2ex]
RoPE & 31.8 & 26.0 & 24.9 & 10.6 & 2.4 & 30.1 & 25.0 & 21.7 & 19.1 & 10.8 & 5.0 & 2.6 & 19.2 & 16.3 & 27.2 & 12.7 & 17.5 \\ 
RoPE-NoPE-LH & 35.0 & 28.0 & 27.1 & 14.0 & 3.7 & \textbf{39.7} & 30.1 & 26.2 & 27.5 & 30.5 & 28.2 & 19.2 & 21.6 & 28.8 & 32.8 & 22.3 & 25.8 \\ 
\hdashline
SWA-RoPE-LH & 30.4 & 28.6 & 25.8 & 12.8 & 6.1 & 31.1 & 27.9 & 24.6 & 23.3 & 25.6 & 15.5 & 12.6 & 20.7 & 22.9 & 28.5 & 18.8 & 22.0 \\ 
+ DRoPE & 28.2 & 23.6 & 23.9 & 15.0 & 11.0 & 30.4 & 25.6 & 21.8 & 20.9 & 20.1 & 18.2 & 16.2 & 20.3 & 21.9 & 26.5 & 18.6 & 21.2 \\ 
SWA-NoPE-LH & \textbf{35.4} & 28.1 & 26.8 & 22.2 & 17.1 & 35.3 & 30.6 & 29.2 & 28.0 & 24.4 & 20.7 & 17.3 & 25.9 & 26.5 & 32.6 & 23.1 & 26.3 \\
\bottomrule
\end{tabular}
\caption{Long-context performance of layer-wise hybrid models after long-context continual pretraining.\label{tab-ruler-lh-long}}
\end{table}

\begin{table}[!ht]
\small
\tabcolsep=0.12cm
\centering
\begin{tabular}{l ccccc ccccccc ccccc}
\toprule 
& \multicolumn{5}{c}{\textbf{RULER}} & \multicolumn{7}{c}{\textbf{BABILong}} & \multicolumn{5}{c}{\textbf{Average}} \\
\cmidrule(lr){2-6} \cmidrule(lr){7-13} \cmidrule(lr){14-18}
 & 4k & 8k & 16k & 32k & 64k & 0k & 2k & 4k & 8k & 16k & 32k & 64k & RU. & BA & $\leq$4k & >4k & All \\
\midrule
\multicolumn{18}{l}{\textbf{\textit{376M Long}}} \\[0.2ex]
RoPE & 31.8 & 26.0 & 24.9 & 10.6 & 2.4 & 30.1 & 25.0 & 21.7 & 19.1 & 10.8 & 5.0 & 2.6 & 19.2 & 16.3 & 27.2 & 12.7 & 17.5 \\ 
RoPE-NoPE-HH & 39.0 & 33.3 & 26.3 & 9.8 & 7.3 & 35.9 & 25.7 & 27.1 & 25.3 & 24.2 & 14.1 & 5.9 & 23.1 & 22.6 & 31.9 & 18.3 & 22.8 \\ 
\hdashline
SWA-RoPE-HH & 24.3 & 19.3 & 16.4 & 9.7 & 6.3 & 39.9 & 33.8 & 27.9 & 27.0 & 25.0 & 17.6 & 13.5 & 15.2 & 26.4 & 31.5 & 16.8 & 21.7 \\ 
+ DRoPE & 22.8 & 20.8 & 16.6 & 16.4 & 14.3 & \textbf{42.6} & 30.9 & 29.1 & 29.5 & 27.2 & 26.2 & 23.8 & 18.2 & 29.9 & 31.3 & 21.9 & 25.0 \\ 
SWA-NoPE-HH & 35.9 & \textbf{35.7} & \textbf{35.1} & 33.7 & \textbf{32.3} & 33.3 & \textbf{34.5} & \textbf{34.5} & \textbf{32.9} & \textbf{31.5} & \textbf{30.5} & \textbf{26.7} & \textbf{34.5} & \textbf{32.0} & \textbf{34.6} & \textbf{32.3} & \textbf{33.1} \\
\bottomrule
\end{tabular}
\caption{Long-context performance of head-wise hybrid models after long-context continual pretraining.\label{tab-ruler-hh-long}}
\end{table}

\begin{table}[!ht]
\small
\tabcolsep=0.12cm
\centering
\begin{tabular}{l ccccc ccccccc ccccc}
\toprule 
& \multicolumn{5}{c}{\textbf{RULER}} & \multicolumn{7}{c}{\textbf{BABILong}} & \multicolumn{5}{c}{\textbf{Average}} \\
\cmidrule(lr){2-6} \cmidrule(lr){7-13} \cmidrule(lr){14-18}
 & 4k & 8k & 16k & 32k & 64k & 0k & 2k & 4k & 8k & 16k & 32k & 64k & RU. & BA & $\leq$4k & >4k & All \\
\midrule
\multicolumn{18}{l}{\textbf{\textit{376M Long}}} \\[0.2ex]
SWA-NoPE-LH & 35.4 & \textbf{28.1} & 26.8 & 22.2 & 17.1 & 35.3 & 30.6 & 29.2 & 28.0 & 24.4 & 20.7 & 17.3 & 25.9 & 26.5 & 32.6 & 23.1 & 26.3 \\ 
+ wsz=256 & 35.1 & 26.4 & 24.8 & 21.7 & 17.7 & 35.7 & 31.0 & 28.7 & 28.8 & 25.5 & 23.5 & 19.2 & 25.2 & 27.5 & 32.6 & 23.5 & 26.5 \\ 
+ wsz=512 & \textbf{37.5} & 26.5 & \textbf{27.4} & 24.4 & 20.9 & \textbf{39.1} & 33.4 & 29.1 & 30.4 & 28.0 & 23.9 & 20.1 & \textbf{27.4} & 29.1 & \textbf{34.8} & 25.2 & 28.4 \\ 
+ wsz=1k & 36.8 & 26.0 & 26.5 & 23.3 & 22.1 & 37.6 & 33.7 & 30.0 & 32.1 & 30.7 & 29.3 & 21.7 & 27.0 & 30.7 & 34.5 & 26.5 & 29.2 \\ 
+ wsz=2k & 36.2 & 24.4 & \textbf{27.4} & \textbf{25.8} & \textbf{22.8} & 34.6 & \textbf{33.8} & \textbf{31.3} & \textbf{32.8} & \textbf{32.2} & \textbf{29.9} & \textbf{24.7} & 27.3 & \textbf{31.3} & 34.0 & \textbf{27.5} & \textbf{29.7} \\
\bottomrule
\end{tabular}
\caption{Long-context performance of SWA-NoPE-LH under long-context continual pretraining with larger window sizes (wsz). Larger window sizes exhibit better long-context performance, implying the Short-Window Weariness.\label{tab-ruler-swa-sww}}
\end{table}

\begin{table}[!ht]
\small
\tabcolsep=0.12cm
\centering
\begin{tabular}{l ccccc ccccccc ccccc}
\toprule 
& \multicolumn{5}{c}{\textbf{RULER}} & \multicolumn{7}{c}{\textbf{BABILong}} & \multicolumn{5}{c}{\textbf{Average}} \\
\cmidrule(lr){2-6} \cmidrule(lr){7-13} \cmidrule(lr){14-18}
 & 4k & 8k & 16k & 32k & 64k & 0k & 2k & 4k & 8k & 16k & 32k & 64k & RU. & BA & $\leq$4k & >4k & All \\
\midrule
\multicolumn{18}{l}{\textbf{\textit{376M Short}}} \\[0.2ex]
SWA-NoPE-LH & 29.1 & 23.5 & 19.3 & 11.0 & \textbf{9.2} & 31.0 & 27.2 & 23.9 & 20.7 & 17.1 & 12.9 & 12.6 & 18.4 & 20.8 & 27.8 & 15.8 & 19.8 \\ 
+ wsz=256 & \textbf{33.2} & \textbf{30.2} & 20.8 & 11.3 & 8.5 & \textbf{42.3} & 35.2 & 29.6 & 24.9 & 23.1 & 17.2 & 15.3 & \textbf{20.8} & 26.8 & \textbf{35.1} & 18.9 & \textbf{24.3} \\ 
+ wsz=512 & 30.6 & 26.1 & 20.6 & 11.6 & 8.3 & 40.5 & \textbf{35.8} & 32.4 & 28.3 & 24.5 & 17.8 & 11.2 & 19.5 & 27.2 & 34.8 & 18.6 & 24.0 \\ 
+ wsz=1k & 32.9 & 25.9 & \textbf{22.8} & \textbf{12.4} & 4.8 & 26.6 & 33.8 & 30.7 & 28.8 & \textbf{25.7} & \textbf{23.0} & \textbf{21.4} & 19.8 & 27.1 & 31.0 & \textbf{20.6} & 24.1 \\ 
+ wsz=2k & 27.1 & 18.2 & 10.2 & 2.4 & 1.6 & 39.9 & 34.1 & \textbf{34.1} & \textbf{30.6} & 25.4 & 19.7 & 15.2 & 11.9 & \textbf{28.4} & 33.8 & 15.4 & 21.6 \\
\bottomrule
\end{tabular}
\caption{Long-context performance of SWA-NoPE-LH under short-context pretraining with larger window sizes (wsz). Shorter window sizes exhibit better length extrapolation, implying the Long-Window Laziness.\label{tab-ruler-swa-lwl}}
\end{table}

\begin{table}[!ht]
\small
\tabcolsep=0.12cm
\centering
\begin{tabular}{l ccccc ccccccc ccccc}
\toprule 
& \multicolumn{5}{c}{\textbf{RULER}} & \multicolumn{7}{c}{\textbf{BABILong}} & \multicolumn{5}{c}{\textbf{Average}} \\
\cmidrule(lr){2-6} \cmidrule(lr){7-13} \cmidrule(lr){14-18}
 & 4k & 8k & 16k & 32k & 64k & 0k & 2k & 4k & 8k & 16k & 32k & 64k & RU. & BA & $\leq$4k & >4k & All \\
\midrule
\multicolumn{18}{l}{\textbf{\textit{376M Long}}} \\[0.2ex]
SWA-RoPE-LH & 30.4 & 28.6 & 25.8 & 12.8 & 6.1 & 31.1 & 27.9 & 24.6 & 23.3 & 25.6 & 15.5 & 12.6 & 20.7 & 22.9 & 28.5 & 18.8 & 22.0 \\ 
+ wsz=2k & 30.9 & 30.9 & 28.0 & 16.9 & 7.6 & 26.1 & 25.9 & 24.7 & 22.3 & 21.8 & 17.8 & 14.1 & 22.9 & 21.8 & 26.9 & 19.9 & 22.3 \\ 
+ LongCE & 32.1 & 29.3 & 29.2 & 28.7 & 13.5 & 37.7 & 31.7 & 29.0 & 31.8 & 32.5 & 18.8 & 15.3 & 26.5 & 28.1 & 32.6 & 24.9 & 27.5 \\ 
+ Both & 33.7 & 34.2 & 32.3 & 26.8 & 12.8 & \textbf{38.7} & 31.5 & 28.3 & 31.2 & \textbf{35.2} & 21.0 & 18.4 & 28.0 & 29.2 & 33.1 & 26.5 & 28.7 \\ 
\hdashline
SWA-NoPE-LH & 35.4 & 28.1 & 26.8 & 22.2 & 17.1 & 35.3 & 30.6 & 29.2 & 28.0 & 24.4 & 20.7 & 17.3 & 25.9 & 26.5 & 32.6 & 23.1 & 26.3 \\
\bottomrule
\end{tabular}
\caption{Long-context performance of SWA layer-wise hybrid models after long-context continual pretraining with extended window sizes (wsz) and LongCE enhancement, both relieving the short-context learning trap effectively.\label{tab-ruler-swa-longce-lh}}
\end{table}

\begin{table}[!ht]
\small
\tabcolsep=0.12cm
\centering
\begin{tabular}{l ccccc ccccc ccccc}
\toprule 
& \multicolumn{5}{c}{\textbf{NIAH-SK1}} & \multicolumn{5}{c}{\textbf{NIAH-SK2}} & \multicolumn{5}{c}{\textbf{NIAH-SK3}} \\ 
\cmidrule(lr){2-6} \cmidrule(lr){7-11} \cmidrule(lr){12-16}
 & 4k & 8k & 16k & 32k & 64k & 4k & 8k & 16k & 32k & 64k & 4k & 8k & 16k & 32k & 64k \\ 
\midrule
\multicolumn{16}{l}{\textbf{\textit{376M Short}}} \\[0.2ex]
GLA-NoPE-LH & 100.0 & 0.0 & 0.0 & 0.0 & 0.0 & 100.0 & 88.0 & 0.0 & 0.0 & 0.0 & 39.0 & 35.0 & 0.0 & 0.0 & 0.0 \\
+ Log & 100.0 & 0.0 & 0.0 & 0.0 & 0.0 & 100.0 & 94.0 & 3.0 & 0.0 & 0.0 & 39.0 & 51.0 & 0.0 & 0.0 & 0.0 \\
+ SWLA & 100.0 & 70.0 & 79.0 & 75.0 & 46.0 & 100.0 & 94.0 & 57.0 & 9.0 & 0.0 & 39.0 & 34.0 & 17.0 & 16.0 & 0.0 \\
+ SWLA \& Log & 100.0 & 70.0 & 81.0 & 84.0 & 84.0 & 100.0 & 99.0 & 97.0 & 92.0 & 68.0 & 39.0 & 34.0 & 52.0 & 70.0 & 37.0 \\
\hdashline
GLA-NoPE-HH & 100.0 & 100.0 & 75.0 & 0.0 & 0.0 & 100.0 & 98.0 & 0.0 & 0.0 & 0.0 & 96.0 & 75.0 & 0.0 & 0.0 & 0.0 \\
\bottomrule
\end{tabular}
\caption{Extrapolation comparison of the GLA/GDN-NoPE hybrid model on the NIAH-SK1, SK2, and SK3 tasks. Applying a sliding-window linear attention and a log scale for NoPE attention achieves the best length extrapolation. \label{tab-niah-matthew-la-all}}
\end{table}

\begin{table}[!ht]
\small
\tabcolsep=0.12cm
\centering
\begin{tabular}{l ccccc ccccccc ccccc}
\toprule 
& \multicolumn{5}{c}{\textbf{RULER}} & \multicolumn{7}{c}{\textbf{BABILong}} & \multicolumn{5}{c}{\textbf{Average}} \\
\cmidrule(lr){2-6} \cmidrule(lr){7-13} \cmidrule(lr){14-18}
 & 4k & 8k & 16k & 32k & 64k & 0k & 2k & 4k & 8k & 16k & 32k & 64k & RU. & BA & $\leq$4k & >4k & All \\
\midrule
\multicolumn{18}{l}{\textbf{\textit{376M Short}}} \\[0.2ex]
RoPE & 25.8 & 0.0 & 0.0 & 0.0 & 0.0 & 21.7 & 17.8 & 8.0 & 0.4 & 0.0 & 0.0 & 0.0 & 5.2 & 6.8 & 18.3 & 0.1 & 6.1 \\ 
RoPE-NoPE-3:1 & 33.1 & 0.3 & 0.0 & 0.0 & 0.0 & 27.1 & 28.4 & 27.0 & 0.3 & 0.0 & 0.0 & 0.0 & 6.7 & 11.8 & 28.9 & 0.1 & 9.7 \\ 
+ EME & 33.1 & 21.9 & 15.6 & 15.8 & 9.3 & 27.0 & 28.4 & 27.0 & 27.0 & 25.3 & 26.0 & 22.9 & 19.1 & 26.2 & 28.9 & 20.5 & 23.3 \\ 
RoPE-NoPE-1:1 & 27.6 & 0.2 & 0.0 & 0.0 & 0.0 & 29.8 & 26.7 & 21.9 & 0.1 & 0.5 & 0.1 & 0.1 & 5.6 & 11.3 & 26.5 & 0.1 & 8.9 \\ 
+ EME & 27.7 & 22.3 & 11.6 & 7.6 & 6.2 & 29.9 & 26.7 & 21.9 & 24.0 & 22.2 & 20.5 & 16.4 & 15.1 & 23.1 & 26.6 & 16.4 & 19.8 \\
\bottomrule
\end{tabular}
\caption{Long-context performance of head-wise hybrids with different hybrid ratios after short-context pre-training.\label{tab-ruler-hybrid-hh-short}}
\end{table}

\begin{table}[!ht]
\small
\tabcolsep=0.12cm
\centering
\begin{tabular}{l ccc ccc ccc ccc cc cc c}
\toprule
& \multicolumn{3}{c}{\textbf{SD}} & \multicolumn{3}{c}{\textbf{MD}} & \multicolumn{3}{c}{\textbf{Sum}} & \multicolumn{3}{c}{\textbf{ICL}} & \multicolumn{2}{c}{\textbf{Syn}} & \multicolumn{2}{c}{\textbf{Code}} & \multirow{2}{*}{\textbf{Avg.}} \\
\cmidrule(lr){2-4} \cmidrule(lr){5-7} \cmidrule(lr){8-10} \cmidrule(lr){11-13} \cmidrule(lr){14-15} \cmidrule(lr){16-17}
 & NQ & Qsp & MF & HG & WQ & Msq & GR & QS & MN & TR & TQ & SS & PC & PR & LCC & Re-P & \\
\midrule
\multicolumn{18}{l}{\textbf{\textit{376M Short}}} \\[0.2ex]
RoPE & 0.6 & 5.0 & 4.1 & 1.0 & 2.2 & 0.5 & 2.8 & 4.4 & 11.3 & 2.0 & 3.2 & 5.2 & 1.0 & \textbf{4.5} & 22.0 & 18.7 & 5.5 \\ 
+ NTK & 0.6 & 5.8 & 5.6 & 1.0 & 3.0 & 0.8 & 3.9 & 8.9 & 9.6 & 2.5 & 3.3 & 8.8 & 0.2 & 2.8 & 27.2 & 30.8 & 7.2 \\ 
+ NTK + Log & 1.4 & 6.1 & 6.8 & 1.6 & 3.9 & 1.3 & 4.3 & 9.3 & 9.8 & 1.5 & 7.3 & 11.9 & 0.5 & 3.6 & 26.9 & 31.3 & 8.0 \\ 
RoPE-NoPE-LH & 0.4 & 6.8 & 4.9 & 1.3 & 2.7 & 0.8 & 3.8 & 4.7 & 12.5 & 2.5 & 4.5 & 2.9 & 1.6 & 1.1 & 21.9 & 19.0 & 5.7 \\ 
+ NTK & 0.6 & 5.3 & 5.2 & 1.1 & 2.2 & 1.4 & 3.3 & 4.5 & 11.3 & 11.0 & 4.9 & 8.8 & 0.8 & 3.0 & \textbf{31.9} & 33.1 & 8.0 \\
\bottomrule
\end{tabular}
\caption{LongBench performance of 376M and 776M layer-wise hybrid models after short-context pretraining.\label{tab-longbench-lh-short-small}}
\end{table}

\begin{table}[!ht]
\small
\tabcolsep=0.12cm
\centering
\begin{tabular}{l ccc ccc ccc ccc cc cc c}
\toprule
& \multicolumn{3}{c}{\textbf{SD}} & \multicolumn{3}{c}{\textbf{MD}} & \multicolumn{3}{c}{\textbf{Sum}} & \multicolumn{3}{c}{\textbf{ICL}} & \multicolumn{2}{c}{\textbf{Syn}} & \multicolumn{2}{c}{\textbf{Code}} & \multirow{2}{*}{\textbf{Avg.}} \\
\cmidrule(lr){2-4} \cmidrule(lr){5-7} \cmidrule(lr){8-10} \cmidrule(lr){11-13} \cmidrule(lr){14-15} \cmidrule(lr){16-17}
 & NQ & Qsp & MF & HG & WQ & Msq & GR & QS & MN & TR & TQ & SS & PC & PR & LCC & Re-P & \\
\midrule
\multicolumn{18}{l}{\textbf{\textit{1B Short}}} \\[0.2ex]
RoPE & 0.4 & 9.5 & 7.1 & 1.3 & 5.2 & 0.4 & 4.2 & 5.8 & 10.7 & 17.5 & 9.2 & 12.7 & 0.3 & 4.2 & 18.5 & 26.9 & 8.4 \\ 
+ NTK & 0.9 & 10.3 & 9.4 & 4.4 & 8.7 & 1.9 & 6.7 & 7.1 & 8.3 & 25.0 & 16.4 & 24.7 & 2.7 & 3.1 & 24.4 & \textbf{38.4} & 12.0 \\ 
+ NTK + Log & 2.2 & 10.6 & 12.1 & 5.5 & 9.3 & 3.8 & 6.6 & 12.3 & 8.4 & 29.5 & 32.0 & \textbf{31.2} & 3.0 & 4.0 & 23.9 & 37.4 & 14.5 \\ 
RoPE-NoPE-LH & 0.9 & 10.9 & 8.9 & 2.3 & 6.4 & 1.0 & 6.6 & 4.8 & 14.5 & 14.0 & 11.4 & 7.8 & 2.0 & 1.1 & 23.9 & 28.3 & 9.0 \\ 
+ NTK & 0.8 & 10.2 & 10.4 & 3.4 & 7.1 & 2.1 & 6.0 & 4.5 & 9.5 & 18.0 & 17.3 & 14.6 & 2.9 & 3.7 & \textbf{29.1} & 36.3 & 11.0 \\
\bottomrule
\end{tabular}
\caption{LongBench performance of 1B and 3B layer-wise hybrid models after short-context pretraining.\label{tab-longbench-lh-short-large}}
\end{table}

\begin{table}[!ht]
\small
\tabcolsep=0.12cm
\centering
\begin{tabular}{l ccc ccc ccc ccc cc cc c}
\toprule
& \multicolumn{3}{c}{\textbf{SD}} & \multicolumn{3}{c}{\textbf{MD}} & \multicolumn{3}{c}{\textbf{Sum}} & \multicolumn{3}{c}{\textbf{ICL}} & \multicolumn{2}{c}{\textbf{Syn}} & \multicolumn{2}{c}{\textbf{Code}} & \multirow{2}{*}{\textbf{Avg.}} \\
\cmidrule(lr){2-4} \cmidrule(lr){5-7} \cmidrule(lr){8-10} \cmidrule(lr){11-13} \cmidrule(lr){14-15} \cmidrule(lr){16-17}
 & NQ & Qsp & MF & HG & WQ & Msq & GR & QS & MN & TR & TQ & SS & PC & PR & LCC & Re-P & \\
\midrule
\multicolumn{18}{l}{\textbf{\textit{376M Short}}} \\[0.2ex]
RoPE & 0.6 & 5.0 & 4.1 & 1.0 & 2.2 & 0.5 & 2.8 & 4.4 & 11.3 & 2.0 & 3.2 & 5.2 & 1.0 & 4.5 & 22.0 & 18.7 & 5.5 \\ 
+ NTK & 0.6 & 5.8 & 5.6 & 1.0 & 3.0 & 0.8 & 3.9 & 8.9 & 9.6 & 2.5 & 3.3 & 8.8 & 0.2 & 2.8 & 27.2 & 30.8 & 7.2 \\ 
+ NTK + Log & 1.4 & 6.1 & 6.8 & 1.6 & 3.9 & 1.3 & 4.3 & 9.3 & 9.8 & 1.5 & 7.3 & 11.9 & 0.5 & 3.6 & 26.9 & 31.3 & 8.0 \\ 
RoPE-NoPE-HH & 1.2 & 6.5 & 5.1 & 1.0 & 3.6 & 0.7 & 4.3 & 5.0 & 15.0 & 4.5 & 5.3 & 5.0 & 0.1 & 2.4 & 18.7 & 17.9 & 6.0 \\ 
+ NTK & 0.7 & 4.9 & 4.4 & 1.0 & 3.2 & 1.0 & 3.0 & 5.2 & 12.5 & 0.0 & 4.3 & 6.2 & 1.3 & 3.3 & 32.0 & 29.9 & 7.1 \\
\bottomrule
\end{tabular}
\caption{LongBench performance of 376M and 776M head-wise hybrid models after short-context pretraining.\label{tab-longbench-hh-short-small}}
\end{table}

\begin{table}[!ht]
\small
\tabcolsep=0.12cm
\centering
\begin{tabular}{l ccc ccc ccc ccc cc cc c}
\toprule
& \multicolumn{3}{c}{\textbf{SD}} & \multicolumn{3}{c}{\textbf{MD}} & \multicolumn{3}{c}{\textbf{Sum}} & \multicolumn{3}{c}{\textbf{ICL}} & \multicolumn{2}{c}{\textbf{Syn}} & \multicolumn{2}{c}{\textbf{Code}} & \multirow{2}{*}{\textbf{Avg.}} \\
\cmidrule(lr){2-4} \cmidrule(lr){5-7} \cmidrule(lr){8-10} \cmidrule(lr){11-13} \cmidrule(lr){14-15} \cmidrule(lr){16-17}
 & NQ & Qsp & MF & HG & WQ & Msq & GR & QS & MN & TR & TQ & SS & PC & PR & LCC & Re-P & \\
\midrule
\multicolumn{18}{l}{\textbf{\textit{1B Short}}} \\[0.2ex]
RoPE & 0.4 & 9.5 & 7.1 & 1.3 & 5.2 & 0.4 & 4.2 & 5.8 & 10.7 & 17.5 & 9.2 & 12.7 & 0.3 & 4.2 & 18.5 & 26.9 & 8.4 \\ 
+ NTK & 0.9 & 10.3 & 9.4 & 4.4 & 8.7 & 1.9 & 6.7 & 7.1 & 8.3 & 25.0 & 16.4 & 24.7 & 2.7 & 3.1 & 24.4 & \textbf{38.4} & 12.0 \\ 
+ NTK + Log & 2.2 & 10.6 & 12.1 & 5.5 & 9.3 & 3.8 & 6.6 & 12.3 & 8.4 & 29.5 & 32.0 & 31.2 & 3.0 & 4.0 & 23.9 & 37.4 & 14.5 \\ 
RoPE-NoPE-HH & 0.4 & 12.3 & 8.8 & 1.8 & 7.0 & 0.7 & 5.0 & 3.5 & 15.4 & 16.5 & 11.1 & 17.0 & 1.8 & 4.5 & 21.4 & 31.2 & 9.9 \\ 
+ NTK & 1.1 & 10.7 & 10.9 & 5.2 & 8.6 & 2.7 & 6.3 & 6.7 & 13.6 & 28.5 & 19.8 & 13.5 & 1.8 & 4.0 & 26.7 & 37.5 & 12.3 \\
\bottomrule
\end{tabular}
\caption{LongBench performance of 1B and 3B head-wise hybrid models after short-context pretraining.\label{tab-longbench-hh-short-large}}
\end{table}

\begin{table}[!ht]
\small
\tabcolsep=0.12cm
\centering
\begin{tabular}{l ccc ccc ccc ccc cc cc c}
\toprule
& \multicolumn{3}{c}{\textbf{SD}} & \multicolumn{3}{c}{\textbf{MD}} & \multicolumn{3}{c}{\textbf{Sum}} & \multicolumn{3}{c}{\textbf{ICL}} & \multicolumn{2}{c}{\textbf{Syn}} & \multicolumn{2}{c}{\textbf{Code}} & \multirow{2}{*}{\textbf{Avg.}} \\
\cmidrule(lr){2-4} \cmidrule(lr){5-7} \cmidrule(lr){8-10} \cmidrule(lr){11-13} \cmidrule(lr){14-15} \cmidrule(lr){16-17}
 & NQ & Qsp & MF & HG & WQ & Msq & GR & QS & MN & TR & TQ & SS & PC & PR & LCC & Re-P & \\
\midrule
\multicolumn{18}{l}{\textbf{\textit{376M Long}}} \\[0.2ex]
RoPE & 1.6 & 9.9 & 10.3 & 3.5 & 6.2 & 2.6 & 20.3 & 10.9 & 16.8 & 18.5 & 17.0 & 17.2 & 2.7 & 3.7 & 29.5 & 32.6 & 12.7 \\ 
RoPE-NoPE-LH & 1.5 & \textbf{12.7} & 12.4 & 3.8 & 7.4 & 3.0 & 19.5 & 10.7 & \textbf{18.9} & 38.0 & 19.8 & 9.6 & 3.2 & 3.7 & 27.8 & 31.8 & 14.0 \\ 
\hdashline
SWA-RoPE-LH & 1.6 & 9.4 & 11.2 & 4.6 & 7.6 & 3.1 & 14.5 & 8.1 & 15.3 & 40.0 & 17.3 & 16.7 & \textbf{3.4} & 3.2 & \textbf{34.9} & 36.8 & 14.2 \\ 
+ DRoPE & 1.4 & 9.0 & 11.5 & 3.8 & 7.1 & 2.9 & 12.7 & 9.7 & 14.2 & 43.0 & 18.5 & 12.5 & 3.1 & 3.9 & 32.1 & 34.9 & 13.8 \\ 
SWA-NoPE-LH & 1.6 & 10.1 & 12.1 & 3.9 & 7.2 & \textbf{3.6} & 11.0 & 10.4 & 17.2 & \textbf{64.5} & 15.0 & 11.8 & 2.8 & 3.8 & 29.5 & 31.4 & 14.7 \\
\bottomrule
\end{tabular}
\caption{LongBench performance of layer-wise hybrid models after long-context continual pretraining.\label{tab-longbench-lh-long}}
\end{table}

\begin{table}[!ht]
\small
\tabcolsep=0.12cm
\centering
\begin{tabular}{l ccc ccc ccc ccc cc cc c}
\toprule
& \multicolumn{3}{c}{\textbf{SD}} & \multicolumn{3}{c}{\textbf{MD}} & \multicolumn{3}{c}{\textbf{Sum}} & \multicolumn{3}{c}{\textbf{ICL}} & \multicolumn{2}{c}{\textbf{Syn}} & \multicolumn{2}{c}{\textbf{Code}} & \multirow{2}{*}{\textbf{Avg.}} \\
\cmidrule(lr){2-4} \cmidrule(lr){5-7} \cmidrule(lr){8-10} \cmidrule(lr){11-13} \cmidrule(lr){14-15} \cmidrule(lr){16-17}
 & NQ & Qsp & MF & HG & WQ & Msq & GR & QS & MN & TR & TQ & SS & PC & PR & LCC & Re-P & \\
\midrule
\multicolumn{18}{l}{\textbf{\textit{376M Long}}} \\[0.2ex]
RoPE & 1.6 & 9.9 & 10.3 & 3.5 & 6.2 & 2.6 & 20.3 & 10.9 & 16.8 & 18.5 & 17.0 & 17.2 & 2.7 & 3.7 & 29.5 & 32.6 & 12.7 \\ 
RoPE-NoPE-HH & \textbf{2.7} & 9.8 & 12.3 & 4.0 & 7.4 & 2.6 & \textbf{22.6} & 9.1 & \textbf{21.7} & 31.0 & 18.2 & 14.5 & 2.4 & 3.8 & 29.2 & 32.4 & 14.0 \\ 
\hdashline
SWA-RoPE-HH & 1.9 & \textbf{11.3} & 12.9 & 3.8 & 6.3 & \textbf{3.2} & 18.2 & \textbf{13.8} & 16.5 & 23.0 & 17.0 & 15.1 & 3.0 & 3.4 & 29.6 & 35.0 & 13.4 \\ 
+ DRoPE & 1.7 & 9.5 & 12.7 & 4.1 & 6.3 & 3.1 & 18.6 & 12.8 & 17.0 & 26.0 & 15.8 & 13.3 & 3.1 & 2.9 & 34.3 & \textbf{35.9} & 13.6 \\ 
SWA-NoPE-HH & 1.7 & 10.4 & \textbf{13.8} & 3.8 & 8.1 & 2.9 & 18.8 & 9.6 & 16.6 & \textbf{62.0} & 18.7 & 16.2 & \textbf{3.2} & \textbf{4.0} & \textbf{36.7} & 34.1 & \textbf{16.3} \\
\bottomrule
\end{tabular}
\caption{LongBench performance of head-wise hybrid models after long-context continual pretraining.\label{tab-longbench-hh-long}}
\end{table}

\begin{table}[!ht]
\small
\tabcolsep=0.12cm
\centering
\begin{tabular}{l ccccc ccccccc ccccc}
\toprule
& \multicolumn{5}{c}{\textbf{RULER}} & \multicolumn{7}{c}{\textbf{BABILong}} & \multicolumn{5}{c}{\textbf{Average}} \\
\cmidrule(lr){2-6} \cmidrule(lr){7-13} \cmidrule(lr){14-18}
 & 4k & 8k & 16k & 32k & 64k & 0k & 2k & 4k & 8k & 16k & 32k & 64k & RU. & BA & $\leq$4k & >4k & All \\
\midrule
\multicolumn{18}{l}{\textbf{\textit{376M Short}}} \\[0.2ex]
RoPE & 25.8 & 0.0 & 0.0 & 0.0 & 0.0 & 21.7 & 17.8 & 8.0 & 0.4 & 0.0 & 0.0 & 0.0 & 5.2 & 6.8 & 18.3 & 0.1 & 6.1 \\ 
RoPE-NoPE-LH & \textbf{28.3} & 0.1 & 0.0 & 0.0 & 0.0 & 41.2 & 27.1 & 24.8 & 0.4 & 0.1 & 0.0 & 0.0 & 5.7 & 13.4 & 30.3 & 0.1 & 10.2 \\ 
SWA-NoPE-LH & 23.2 & 16.0 & \textbf{14.0} & \textbf{13.2} & \textbf{10.5} & 31.8 & 26.1 & 21.8 & 15.8 & 14.7 & 12.0 & 10.6 & \textbf{15.4} & 19.0 & 25.7 & \textbf{13.4} & 17.5 \\ 
GLA-NoPE-LH & 21.0 & 3.8 & 1.5 & 2.3 & 2.2 & \textbf{41.5} & \textbf{36.4} & \textbf{33.5} & \textbf{20.9} & \textbf{16.3} & \textbf{16.4} & \textbf{17.0} & 6.2 & \textbf{26.0} & \textbf{33.1} & 10.1 & \textbf{17.7} \\ 
GDN-NoPE-LH & 26.0 & \textbf{16.5} & 5.1 & 1.5 & 0.7 & 35.4 & 28.7 & 22.7 & 18.4 & 12.5 & 10.3 & 9.9 & 10.0 & 19.7 & 28.2 & 9.4 & 15.6 \\
\bottomrule
\end{tabular}
\caption{Long-context performance of layer-wise hybrid models under the YOCO-like layout.\label{tab-ruler-yoco}}
\end{table}

\begin{table}[!ht]
\small
\tabcolsep=0.12cm
\centering
\begin{tabular}{l ccccc ccccccc ccccc}
\toprule
& \multicolumn{5}{c}{\textbf{RULER}} & \multicolumn{7}{c}{\textbf{BABILong}} & \multicolumn{5}{c}{\textbf{Average}} \\
\cmidrule(lr){2-6} \cmidrule(lr){7-13} \cmidrule(lr){14-18}
 & 4k & 8k & 16k & 32k & 64k & 0k & 2k & 4k & 8k & 16k & 32k & 64k & RU. & BA & $\leq$4k & >4k & All \\
\midrule
\multicolumn{18}{l}{\textbf{\textit{376M Short}}} \\[0.2ex]
SWA-NoPE-LH & 29.1 & 23.5 & 19.3 & \textbf{11.0} & \textbf{9.2} & 31.0 & 27.2 & 23.9 & 20.7 & 17.1 & 12.9 & 12.6 & 18.4 & 20.8 & 27.8 & 15.8 & 19.8 \\ 
+ base=5000 & 27.9 & 19.6 & 12.7 & 8.0 & 7.0 & 40.7 & 34.5 & 32.1 & 27.5 & 24.0 & 15.9 & 14.2 & 15.0 & 27.0 & 33.8 & 16.1 & 22.0 \\ 
+ base=2500 & 28.5 & 16.2 & 11.4 & 9.1 & 7.7 & \textbf{45.7} & 33.8 & 28.0 & 24.0 & 17.2 & 11.3 & 14.6 & 14.6 & 24.9 & 34.0 & 13.9 & 20.6 \\ 
+ base=1000 & 28.7 & 21.9 & 15.5 & 10.9 & 7.7 & 32.6 & 32.0 & 28.8 & 25.4 & 21.2 & 13.1 & 11.4 & 16.9 & 23.5 & 30.5 & 15.9 & 20.8 \\ 
+ base=500 & 25.6 & 21.1 & 17.5 & 10.7 & 7.3 & 32.4 & 35.0 & 30.6 & 26.3 & 20.8 & 14.8 & 14.4 & 16.4 & 24.9 & 30.9 & 16.6 & 21.4 \\
\bottomrule
\end{tabular}
\caption{Long-context performance of SWA-NoPE-LH with smaller rotary bases, from 10000 to 100, under direct or log-scaled NoPE extrapolation after short-context pretraining, and context extension after long-context pretraining.\label{tab-ruler-swa-base}}
\end{table}

\end{document}